\documentclass[twoside]{article}

\usepackage[accepted]{aistats2024}
\usepackage[utf8]{inputenc} %
\usepackage[T1]{fontenc}    %
\usepackage{url}            %
\usepackage{booktabs}       %
\usepackage{amsfonts}       %
\usepackage{nicefrac}       %
\usepackage{microtype}      %

\usepackage{xcolor}
\definecolor{darkblue}{rgb}{0,0,0.5}
\usepackage[breaklinks,colorlinks, linkcolor=darkblue, urlcolor=darkblue, citecolor=darkblue]{hyperref}
\usepackage{amsmath}
\usepackage{amssymb}
\usepackage{algorithm}
\usepackage{comment}
\usepackage{graphicx}
\usepackage{xspace}
\usepackage{subcaption}
\usepackage{tikz}
\usepackage{pgfplots}
\usetikzlibrary{hobby, fit, shapes.geometric, patterns, calc, positioning}
\usepackage{tabularx}
\usepackage{listings}
\usepackage{xcolor}         %
\definecolor{codegreen}{rgb}{0,0.6,0}
\definecolor{codegray}{rgb}{0.5,0.5,0.5}
\definecolor{codepurple}{rgb}{0.58,0,0.82}
\definecolor{backcolour}{rgb}{0.95,0.95,0.92}
\newcommand{\mybox}[1]{\protect\tikz[baseline=-.5ex]\protect\draw[color=#1, line width=1.3pt](0, 0)--(1em, 0);}

\lstdefinestyle{mystyle}{
    backgroundcolor=\color{backcolour},   
    commentstyle=\color{codegreen},
    keywordstyle=\color{magenta},
    numberstyle=\tiny\color{codegray},
    stringstyle=\color{codepurple},
    basicstyle=\ttfamily\footnotesize,
    breakatwhitespace=false,         
    breaklines=true,                 
    captionpos=b,                    
    keepspaces=true,                 
    numbers=left,                    
    numbersep=5pt,                  
    showspaces=false,                
    showstringspaces=false,
    showtabs=false,                  
    tabsize=2
}
\usepackage[round]{natbib}

\newcommand{\N}{\mathrm{N}}   %
\newcommand{\E}{\mathbb{E}}
\newcommand{\RR}{\mathbb{R}}

\newcommand{\eg}{\textit{e.g.}\xspace}
\newcommand{\ie}{\textit{i.e.}\xspace}

\newcommand{\KL}[2]{\mathrm{D_{KL}}\left[ {#1} \, \| \, {#2}\right]}
\renewcommand{\mid}{\,|\,}

\newlength{\figurewidth}
\newlength{\figureheight}

\usepackage[capitalize,nameinlink]{cleveref}
\crefname{section}{Sec.}{Secs.}
\crefname{proposition}{Prop.}{Props.}
\crefname{lemma}{Lem.}{Lems.}
\crefname{model}{Mod.}{Mods.}
\crefname{appendix}{App.}{Apps.}
\crefname{algorithm}{Alg.}{Algs.}

\definecolor{pythonred}{HTML}{d62728}
\definecolor{pythonblue}{HTML}{1f77b4}
\definecolor{pythonorange}{HTML}{ff7f0e}

\begin{document}

\runningtitle{PriorCVAE: Scalable MCMC Parameter Inference}

\runningauthor{Semenova, Verma, Cairney-Leeming, Solin, Bhatt, Flaxman}

\twocolumn[

	\aistatstitle{PriorCVAE: Scalable MCMC Parameter Inference \\ with Bayesian Deep Generative Modelling}
	
	\aistatsauthor{Elizaveta Semenova \And Prakhar Verma\thanks{} \And Max Cairney-Leeming\thanks{}}
	\aistatsaddress{ University of Oxford \And  Aalto University  \And ISTA (Institute of Science \\and Technology Austria)}
	
	\aistatsauthor{Arno Solin \And  Samir Bhatt \And Seth Flaxman }
	\aistatsaddress{Aalto University \And University of Copenhagen \\Imperial College London  \And University of Oxford }

]

\begin{abstract}
Recent advances have shown that GP priors, or their finite realisations, can be encoded using deep generative models such as variational autoencoders (VAEs). These learned generators can serve as drop-in replacements for the original priors during MCMC inference. While this approach enables efficient inference, it loses information about the hyperparameters of the original models, and consequently makes inference over hyperparameters impossible and the learned priors indistinct. To overcome this limitation, we condition the VAE on stochastic process hyperparameters. This allows the joint encoding of hyperparameters with GP realizations and their subsequent estimation during inference. Further, we demonstrate that our proposed method, PriorCVAE, is agnostic to the nature of the models which it approximates, and can be used, for instance, to encode solutions of ODEs. It provides a practical tool for approximate inference and shows potential in real-life spatial and spatiotemporal applications. \looseness-1
\end{abstract}

\section{INTRODUCTION}
\label{sec:introduction}
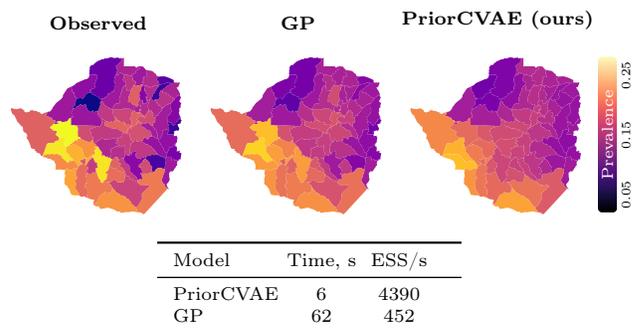
\begin{figure}[t!]
  \centering
  \begin{tikzpicture}[inner sep=0,node distance=0cm]

    \node (obs) at (0,0)
    {\includegraphics[width=.32\columnwidth,trim=40 25 25 25,clip]{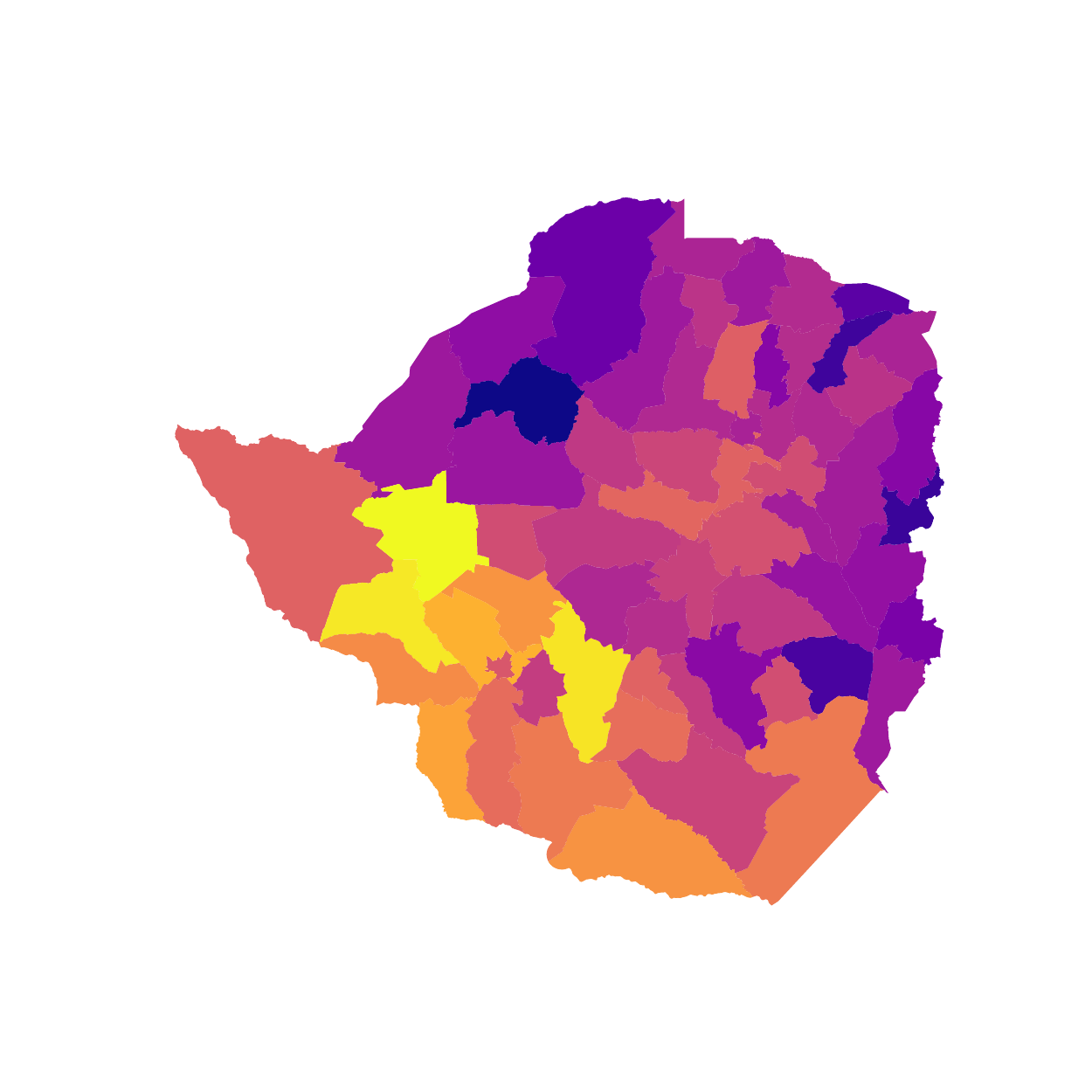}};
    \node[right=of obs] (gp)
    {\includegraphics[width=.32\columnwidth,trim=40 25 25 25,clip]{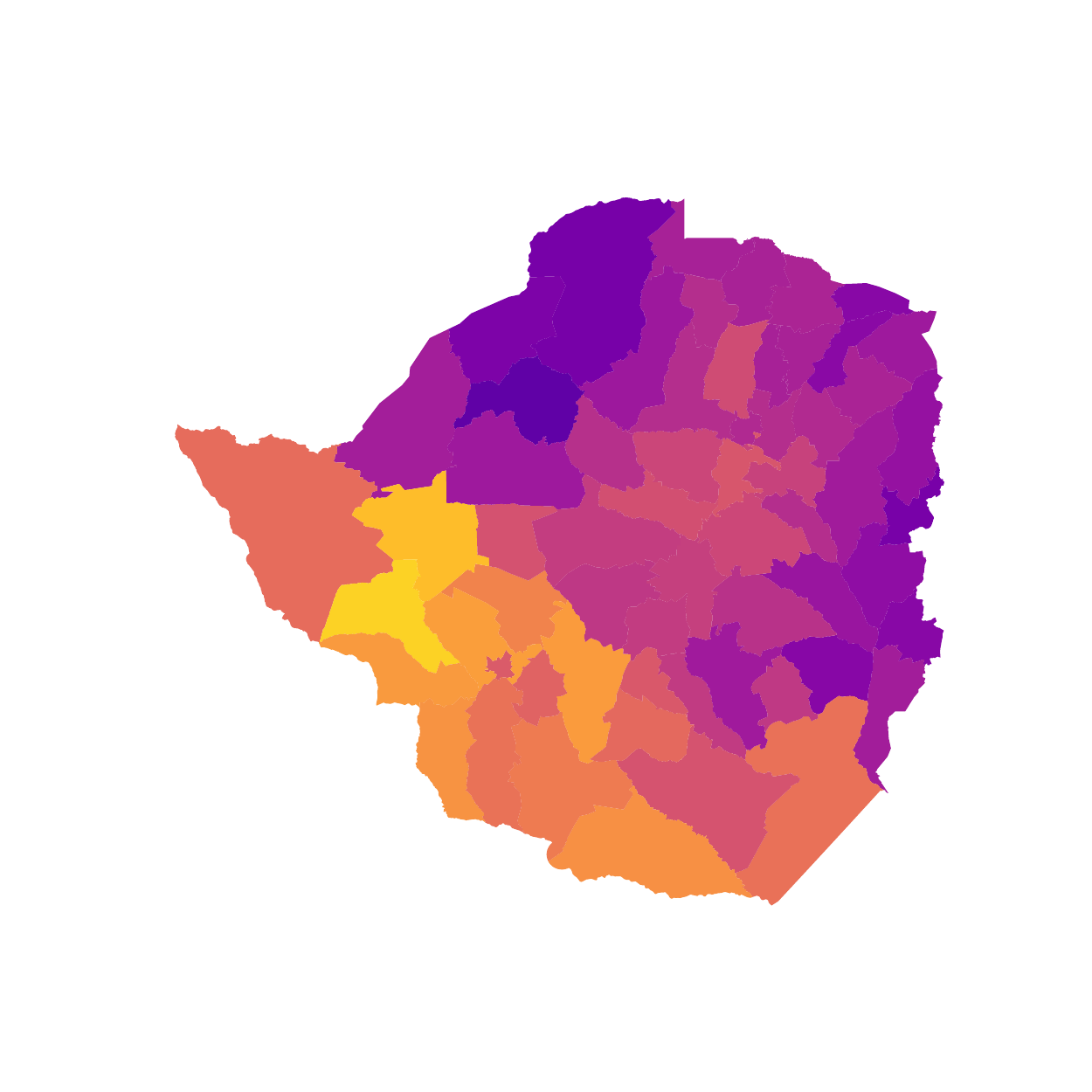}};
    \node[right=of gp] (pcvae)
    {\includegraphics[width=.32\columnwidth,trim=40 25 25 25,clip]{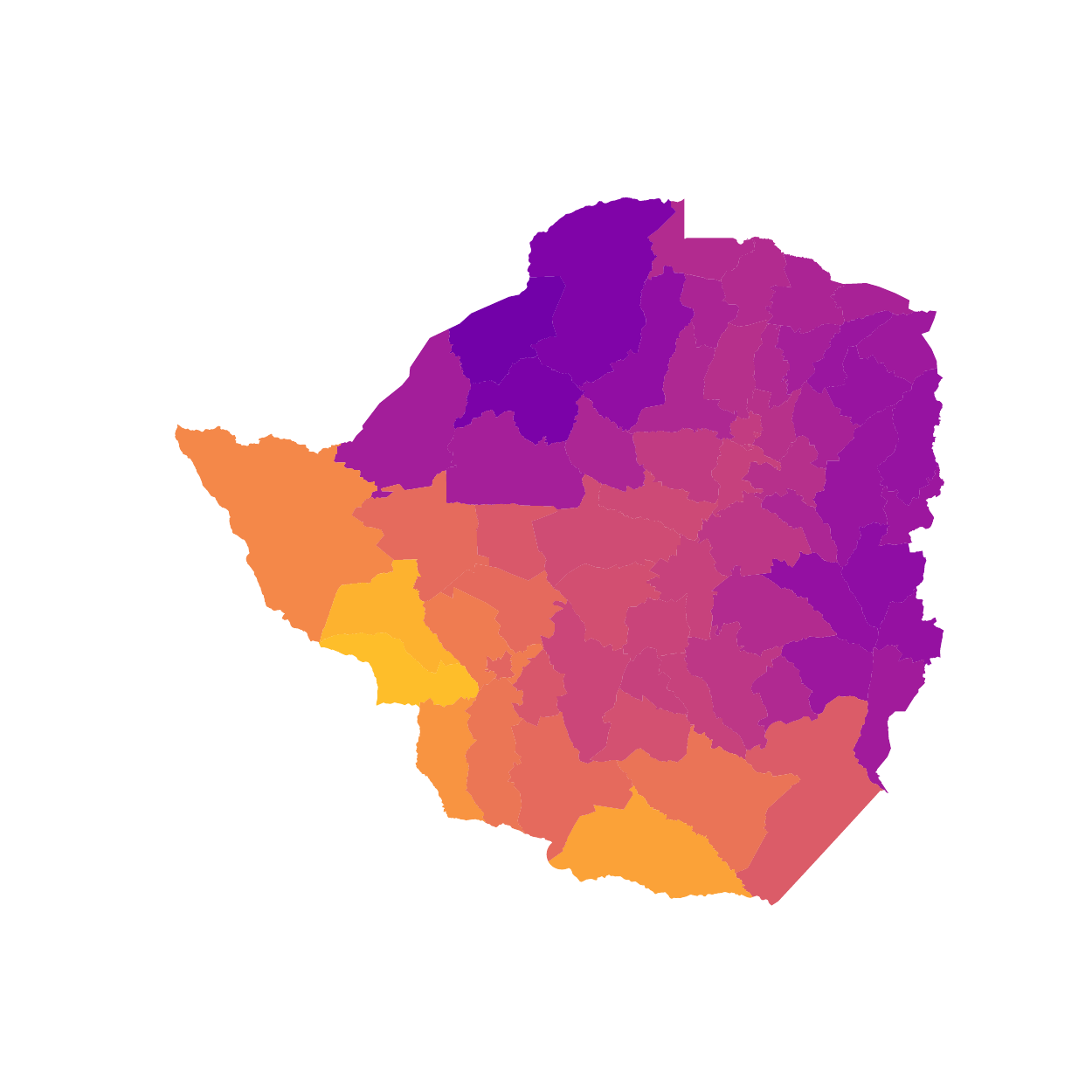}};

    \node[right=of pcvae,minimum width=.03\columnwidth,minimum height=.25\columnwidth] (cbar) {};
    \begin{scope}[shift={(cbar.south west)}, x={(cbar.south east)}, y={(cbar.north west)}]
      \clip[rounded corners=2pt] (0,0) rectangle (1,1);
      \node at (.5,.5) {\includegraphics[width=.03\columnwidth,height=.25\columnwidth]{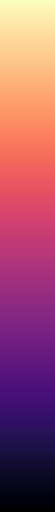}};
    \end{scope}

    \node[above=of obs,font=\scriptsize\bf] {Observed};
    \node[above=of gp,font=\scriptsize\bf] {GP};
    \node[above=of pcvae,font=\scriptsize\bf] {PriorCVAE (ours)};

    \node[rotate=90,font=\tiny\color{white}] at (cbar.center) {Prevalence};
    \node[anchor=north west,font=\tiny,scale=.8,inner sep=2pt,rotate=90] at ($(cbar.south east)$) {0.05};
    \node[anchor=north,font=\tiny,scale=.8,inner sep=2pt,rotate=90] at ($(cbar.east)$) {0.15};
    \node[anchor=north east,font=\tiny,scale=.8,inner sep=2pt,rotate=90] at ($(cbar.north east)$) {0.25};
   
  \end{tikzpicture}\\
  \begin{subfigure}[!t]{\columnwidth}
  	\centering
	\scriptsize
	\begin{tabularx}{.49\textwidth}{l@{\hspace{0.4em}}c@{\hspace{0.7em}}c}
		\toprule 
		Model & Time, s & ESS/s \\
		\midrule 
		PriorCVAE & 6 & 4390\\ 
		GP &  62& 452 \\ 
		\bottomrule
	\end{tabularx}
  \end{subfigure}
  \caption{PriorCVAE versus GP inference on HIV Prevalence in Zimbabwe. PriorCVAE runs $10\times$ faster than the GP model and gives competitive results.}
  \label{fig:teaser}
\end{figure}

In numerous applied domains, Gaussian processes \citep[GPs,][]{williams2006gaussian} have emerged as the preferred priors within Bayesian hierarchical frameworks. Their wide adoption in environmental sciences~\citep{dai2022gaussian}, geosciences~\citep{axen2022spatiotemporal}, agriculture~\citep{belda2021crop}, the pharmaceutical industry~\citep{obrezanova2007gaussian, schroeter2007predicting, semenova2021flexible, shapovalova2022non}, remote sensing~\citep{you2017deep}, robotics~\citep{deisenroth2013gaussian}, and active learning ~\citep{krause2007nonmyopic} can be attributed to their versatility as universal approximators, depending on the covariance function, and their analytical convenience when used with a Gaussian likelihood. Despite these advantages, limitations arise when applying GPs to real-world tasks: inference scales cubically with the number of observed data points, making characterising  uncertainty with \emph{fully Bayesian} inference (incl.\ hyperparameters, see \cref{fig:teaser}) prohibitively computationally expensive \citep[see][]{Doucet_undated-zx}.\looseness-1

Deep generative models offer a pathway to address the scaling challenge of GPs in fully Bayesian modelling. GP priors and a wider range of stochastic processes can be encoded using variational autoencoders \citep[VAEs,][]{kingma2013auto}. Recently, a group of two-stage methods has been proposed in the literature: $\pi$VAE~\citep{mishra2020pi} encoding stochastic processes, and PriorVAE~\citep{semenova2022priorvae} encoding realisations thereof. By training a VAE on draws from a stochastic process prior and using the trained decoder as a surrogate for the original GP during Bayesian inference, PriorVAE effectively sidesteps the computational overhead associated with GPs. This enables fast and efficient inference using methods like Markov chain Monte Carlo (MCMC). However, there is a serious drawback in this approach: while $\pi$- and PriorVAE encode function values, neither can explicitly encode and estimate hyperparameters. In this work, we propose to overcome this issue with a conditional variational autoencoder \citep[CVAE,][]{sohn2015learning} architecture.

This paper introduces the novel PriorCVAE approach, which employs the conditional VAE framework to shift learning emphasis towards priors, rather than solely on observed data.  The major advantage is the ability to condition on the hyperparameters of the priors, which retains both the benefits of reduced computation times and high effective sample sizes as seen with PriorVAE, but also endows the model with a capacity for clear prior identification. Consequently, downstream MCMC tasks can now explicitly estimate hyperparameters. This integration is especially promising for spatial and spatiotemporal inference, as shown in the experiments, offering a robust class of approximate GP models, and our new method is versatile and seamlessly integrates with well-known probabilistic programming frameworks such as Stan~\citep{carpenter2017stan}, NumPyro~\citep{phan2019composable}, PyMC~\citep{pymc2023}, and others~\citep{vstrumbelj2023past}. Even though this work was originally motivated by the problem of GP scalability, the proposed method is agnostic to the model that it approximates. For example, solutions of dynamical systems described by ordinary differential equations can also be encoded as we show in one of the experiments.

Our contributions are summarised as follows:
\begin{itemize}
    \item  We propose the PriorCVAE method that can `identify' the learned priors by introducing a conditional variational autoencoder architecture, allowing us to explicitly encode hyperparameters and estimate them at the inference stage;
    \item We show that PriorCVAE can extrapolate prior draws with respect to hyperparameters---\ie~we can draw priors for values of hyperparameters which the PriorCVAE was not trained on;
    \item We demonstrate for the first time the applicability of the method to GPs with non-stationary kernels, and show that properties of functions obtained via non-linear transformations of function values can be learned alongside function evaluations.
    \item We demonstrate for the first time that PriorCVAE is agnostic to the type of models that it can approximate by encoding a solution of a system of ordinary differential equations (ODEs).
\end{itemize}
We provide comparison to different approximate inference methods (Laplace, ADVI, PriorVAE), and showcase our approach on real-world epidemiology data sets.\looseness-1

\section{BACKGROUND}
\label{sec:background}
We provide a concise overview of MCMC as a tool for Bayesian inference in GP models, introduce deep generative models, and summarise VAE and CVAE architectures as required to set up the PriorVAE and $\pi$VAE methods.

\subsection{Markov Chain Monte Carlo and Probabilistic Programming Languages}
Under the Bayesian inference paradigm, we are interested in learning the probability distribution of a set of unknown parameters (or latent variables) $z$ given observed data $y$. This is defined by their joint distribution $p(y, z) = p(y \mid z)\,p(z)$, where $p(y \mid z)$ denotes the \emph{likelihood} of the observed data and $p(z)$ the \emph{prior} for the latent variable $z$. Consequently, the \emph{posterior} of the latent variable is expressed as $p(z \mid y) \propto p(y\mid z) \, p(z).$ 

Markov Chain Monte Carlo (MCMC) is a general sampling technique for sampling from a (possibly) unnormalised target density $\pi(z)$, \ie~the posterior~\citep{robert1999monte, gelman1995bayesian}. MCMC simulates an ergodic Markov chain $\{z^{(i)}\}_{i \in \mathbb{N}}$ which, under the right conditions, converges to the target distribution $\pi(z)$. In practice, the simulation is performed for a finite number of iterations $N$, which is sufficiently large to ensure convergence. Diagnotic tools like the R-hat statistic and effective sample size (ESS) metrics aid in assessing the performance and convergence of MCMC \citep{vehtari2021rank}. The main advantage of MCMC is the guarantee (under ergodicity conditions) of asymptotic convergence to the target density, which makes MCMC preferable to other methods when our modelling informs decision-making.

Probabilistic programming languages \citep[PPLs,][]{van2018introduction, gordon2014probabilistic, salvatier2016probabilistic} abstract MCMC inference algorithms from a user, and allow modellers to focus on model formulation. By automating inference, they significantly lower the cost of iterating model design, leading to a better overall model in a shorter period of time. PPLs also often provide in-built diagnostics and visualisation tools~\citep{arviz_2019} so users can quickly assess results.  

However, MCMC has its limitations. Especially in complex settings, \eg~with Gaussian processes or large spatial models, it can be computationally demanding and scales poorly. Simpler inference methods can be useful here, but because PPLs abstract away the visibility to the model internals, this can also pose risks to practical use. We seek to provide a plug\&play inference framework used in PPLs as part of large-scale GP modelling or models where a similar approach can provide a remedy.

\subsection{Auto-encoding Generative Models}
Our work hinges on the insight that deep generative models can be used as surrogates in Bayesian modelling. A common shared principle behind generative models~\citep{bond2021deep, tomczak2022deep} is to start with samples from the latent distribution, a simple probability distribution such as an i.i.d.~Gaussian normal $z \sim \mathcal{N}(0, I)$, and apply a generative model in the form of a neural network with trainable parameters to transform the latent distribution into samples from the target distribution. This procedure ensures that the model, during training with samples from the target distribution, learns both the underlying probability structure and the relationship between the generator's output and the target distribution.

\textbf{Variational autoencoders} \citep[VAEs,][]{kingma2013auto} are deep generative models with a dual structure: {\em (i)}~the encoder $E_\gamma(\cdot)$ maps input $y \in \mathcal{Y} \subset \RR^n$ into a latent space $\mathcal{Z} \subset \RR^d$, where $d$ is typically lower than $n$; {\em (ii)}~the decoder $D_\psi(\cdot)$ reconstructs this input from the latent representation $z \in \mathcal{Z}$. This setup introduces an `information bottleneck', ensuring the latent representation $z$ encapsulates a compressed version of the input $y$. 

Unlike traditional autoencoders, a VAE maps $y$ to a distribution in the latent space, rather than a fixed vector. To facilitate this, VAEs employ variational approximation \citep{wainwright2008introduction, blei2017variational, wainwright2008graphical} to estimate the posterior distribution: $p(z \mid y) \propto p(y \mid z) \, p(z)$ and $p(z \mid y) \approx q(z \mid y)$. Typically, a Gaussian is chosen as the variational family, leading to $q(z \mid y) = \N(\mu_z, \sigma^2_zI_d)$ where $(\mu_z, \log \sigma^2_z) = E_\gamma(y)$ and $z \sim \N(\mu_z, \sigma^2_z I_d)$. The prior over $z$, $p(z)$, remains a standard Gaussian $\N(0,I_d)$. As per \citet{kingma2013auto}, to achieve optimal encoder and decoder parameters, one maximizes the evidence lower bound or equivalently, minimizes the following loss function:
\begin{equation}
\mathcal{L}_\text{VAE} = \E_{q(z \mid y)} \left[-\log p (y \mid z)\right] + \KL{q(z \mid y)}{p(z)},    
\end{equation}
where the first component represents the reconstruction quality and the second regularises the difference between the latent variable distribution and its prior.

\begin{figure}[t!]
  \centering
  \includegraphics[width=\columnwidth]{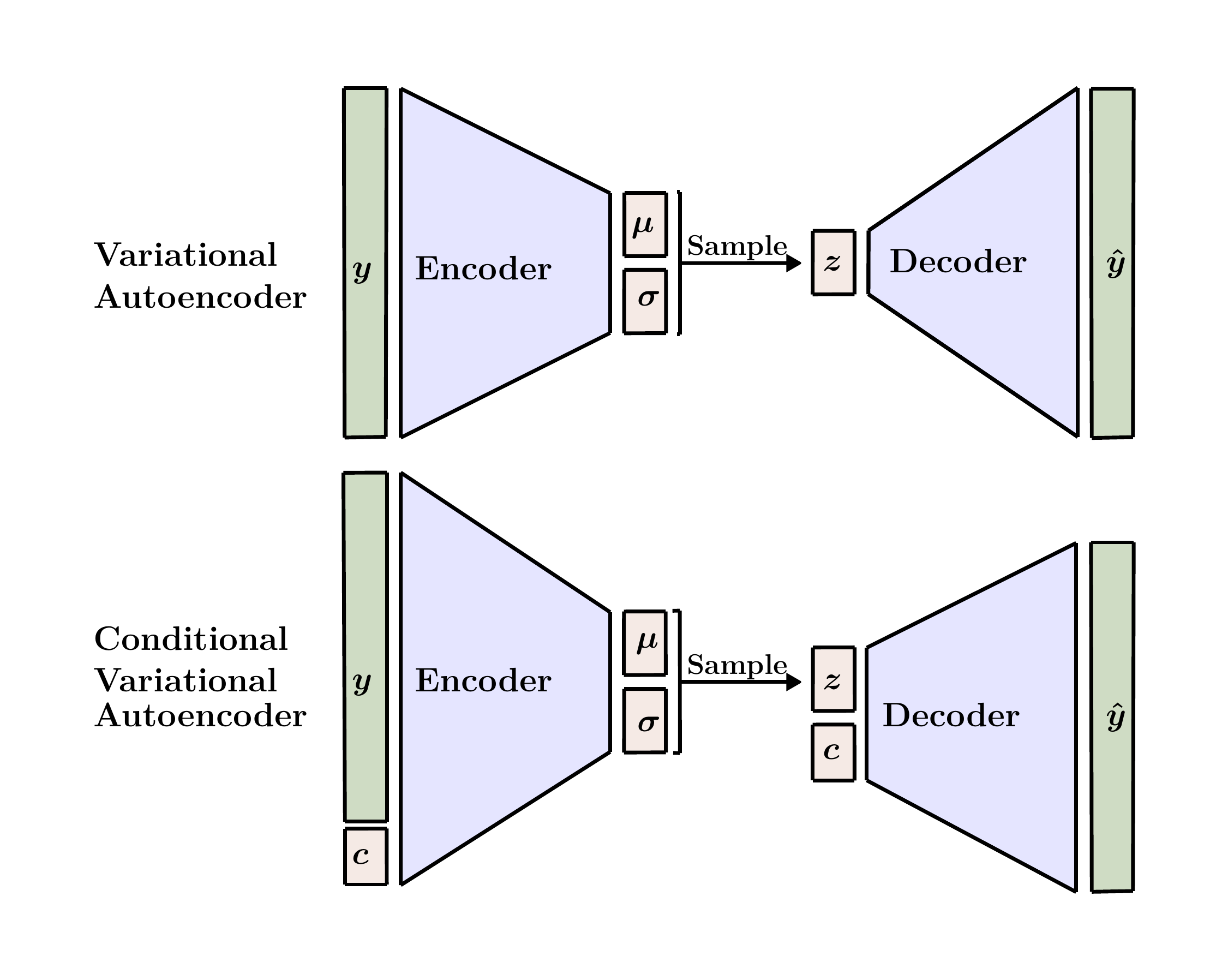}
  \caption{VAE versus CVAE architectures. In CVAE models, both the encoder and decoder get the label $c$ as input. Traditionally, $c$ is an observed class label. Here we interpret $c$ as a hyperparameter of the prior.}
  \label{fig:vae_cvae_architecture}
\end{figure}

\textbf{Conditional VAEs} \citep[CVAEs,][]{sohn2015learning} extend VAEs by conditioning the generative process on additional information $c$ (\eg, a class label)---see \cref{fig:vae_cvae_architecture}. This allows CVAEs to handle diverse input--output mappings, as the optimization objective now involves both the data and the conditioning input. This allows us to control the class which we want to generate samples from, and allows the CVAE to learn how to encode and decode separate classes of input differently. The objective for optimisation becomes:
\begin{multline}
    \mathcal{L}_\text{CVAE} =  \E_{q(z \mid y,c)} \left[-\log p (y \mid z,c)\right] + \\ \KL{q(z \mid y,c)}{p(z \mid c)}.    
\end{multline}

\begin{figure*}[t!]
  \centering\scriptsize
  \pgfplotsset{axis on top,scale only axis,width=\figurewidth,height=\figureheight, ylabel near ticks,ylabel style={yshift=6pt},y tick label style={rotate=90},yticklabels={},xticklabels={}} 
  \setlength{\figurewidth}{.14\textwidth}
  \setlength{\figureheight}{\figurewidth}
  \newcommand{\mytitle}[1]{\tikz\node[minimum width=\figurewidth,minimum height=2em,align=center]{#1};}
  \begin{subfigure}{.19\textwidth}  
    \raggedleft
    \mytitle{\textbf{GP} ($\ell{=}0.1$)}    
    \pgfplotsset{ylabel={Output, $y$}}
    \tikz[outer sep=0pt, inner sep=0pt]\node{\input{figures/Matern52/GP_samps_0_1}};
  \end{subfigure}
  \hfill
  \begin{subfigure}{.15\textwidth}
    \raggedleft
    \mytitle{\textbf{PriorCVAE}}
    \tikz[outer sep=0pt, inner sep=0pt]\node{\input{figures/Matern52/PriorCVAE_samps_0_1}};
  \end{subfigure}
  \hfill
  \begin{subfigure}{.15\textwidth}
    \raggedleft
    \mytitle{\textbf{PriorVAE}}  
    \tikz[outer sep=0pt, inner sep=0pt]\node{\input{figures/Matern52/PriorVAE_samps_0_1}};
  \end{subfigure}
  \hfill
  \begin{subfigure}{.15\textwidth} 
    \raggedleft
    \mytitle{\textbf{GP} ($\ell{=}1.0$)}
    \tikz[outer sep=0pt, inner sep=0pt]\node{\input{figures/Matern52/GP_samps_1_0}};
  \end{subfigure}
  \hfill
  \begin{subfigure}{.15\textwidth} 
    \raggedleft
    \mytitle{\textbf{PriorCVAE}}  
    \tikz[outer sep=0pt, inner sep=0pt]\node{\input{figures/Matern52/PriorCVAE_samps_1_0}};
  \end{subfigure}
  \hfill
  \begin{subfigure}{.15\textwidth} 
    \raggedleft
    \mytitle{\textbf{PriorVAE}}    
    \tikz[outer sep=0pt, inner sep=0pt]\node{\input{figures/Matern52/PriorVAE_samps_1_0}};
  \end{subfigure}\\[1em]
  \newcommand{\addxy}[1]{\tikz[node distance=10pt,inner sep=0]{\node(a){\includegraphics[width=\figurewidth]{figures/Matern52/#1}};\node[left=of a,rotate=90,anchor=center]{Input, $x'$};\node[below=of a,anchor=center]{Input, $x$};\node[draw,minimum width=\figurewidth,minimum height=\figurewidth] at (a) {};}}
  \newcommand{\addx}[1]{\tikz[node distance=10pt,inner sep=0]{\node(a){\includegraphics[width=\figurewidth]{figures/Matern52/#1}};\node[below=of a,anchor=center]{Input, $x$};\node[draw,minimum width=\figurewidth,minimum height=\figurewidth] at (a) {};}}
  \begin{subfigure}{.19\textwidth}  
    \raggedleft  
    \addxy{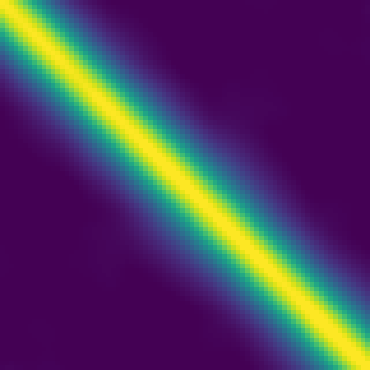}
  \end{subfigure}
  \hfill
  \begin{subfigure}{.15\textwidth}
    \raggedleft
    \addx{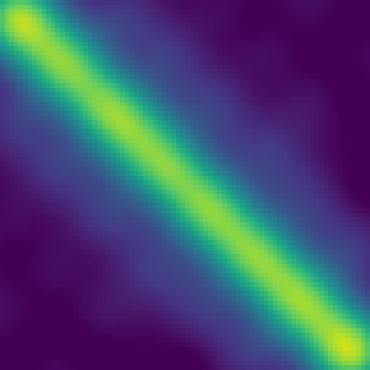}
  \end{subfigure}
  \hfill
  \begin{subfigure}{.15\textwidth}
    \raggedleft
    \addx{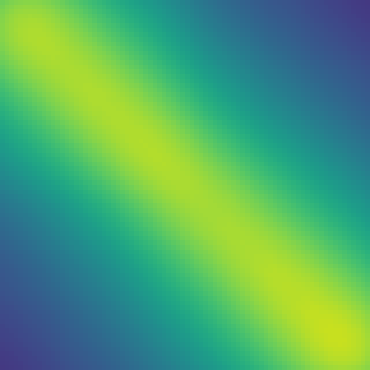}
  \end{subfigure}
  \hfill
  \begin{subfigure}{.15\textwidth}
    \raggedleft
    \addx{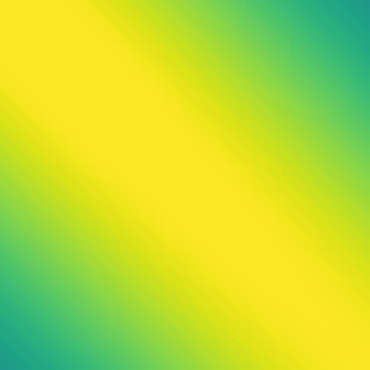}
  \end{subfigure}
  \hfill
  \begin{subfigure}{.15\textwidth} 
    \raggedleft
    \addx{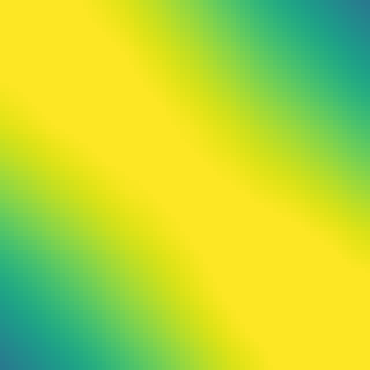}
  \end{subfigure}
  \hfill
  \begin{subfigure}{.15\textwidth}
    \raggedleft 
    \addx{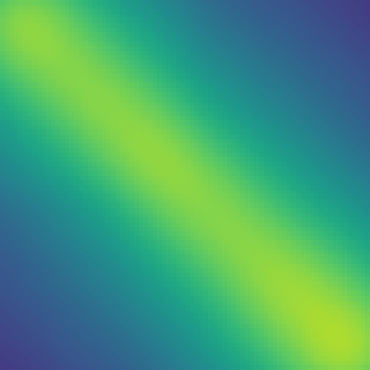}
  \end{subfigure}\\
  \newcommand{\cbar}{\protect\includegraphics[width=2.5em,height=.7em]{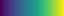}}
  \caption{\textbf{Our PriorCVAE samples and covariance resemble the original GP.} Example draws (top) and computed Gram (covariance) matrices (bottom, 0~\cbar~1) from a Mat\'ern-$\nicefrac{5}{2}$ 
  prior with lengthscale $\ell=0.1$ (three left-hand columns) and $\ell=1.0$ (three right-hand columns).}
  \label{fig:exp_matern52_encode_prior}
\end{figure*}

\subsection{PriorVAE/$\pi$VAE: Encoding Priors}
$\pi$VAE~\citep{mishra2020pi} and PriorVAE~\citep{semenova2022priorvae} are two related VAE-based methods that can encode continuous stochastic processes and their finite realisations, respectively. They use the trained decoder as an approximation of computationally complex structures for Bayesian inference with MCMC. In this way, the rigour of MCMC is preserved, enabling inference of complex and expressive models, while these models are scalable due to the simple structure of the VAE's latent space. %
$\pi$VAE learns low-dimensional embeddings of function classes inspired by the Karhunen--Lo\'eve expansion: any stochastic process $f(s)$ can be represented as an infinite sum $f(s) = \sum_{j=1}^\infty \beta_j \phi_j(s)$, where $\{\phi_j(\cdot)\}_{j=1}^\infty$ is a set of orthonormal basis functions and $\{\beta_j\}_{j=1}^\infty$ are uncorrelated random variables. To train $\pi$VAE, the basis functions are learned with a feed-forward neural network over $s$, and the random coefficients $\beta_j$ are encoded using a VAE. 

The PriorVAE method was originally proposed as a scalable solution to the small area estimation (SAE) problem in spatial statistics. It encodes finite realisations of a Gaussian process, \ie~multivariate normal distributions, which are widely used in spatial statistics. Such distributions are often defined by precision matrices based on the adjacency structure of the modelled areas and include the conditional auto-regressive (CAR), intrinsic conditional auto-regressive (ICAR) and Besag--York--Molli\'e (BYM) models~\citep{besag1974spatial, besag1991bayesian, riebler2016intuitive}. PriorVAE must be trained on a pre-defined spatial structure, which is a disadvantage compared to $\pi$VAE as PriorVAE cannot make predictions at off-grid locations, but is much simpler in common scenarios where the spatial structure is known in advance, as only the VAE needs to be trained, not the feature map. The PriorVAE method is as follows:\looseness-1
\begin{enumerate}
    \item Fix the spatial structure of interest $\{x_1, \dots, x_n\}$, \eg~a set of administrative units, or an artificial computational grid.
    \item Draw evaluations of a GP prior $f \sim \mathcal{GP}(\cdot)$ over the spatial structure and train the VAE to approximate these evaluations $f_\text{GP}=\left(f(x_1), \dots, f(x_n)\right)^\top$.
    \item Perform Bayesian inference of the overarching model using MCMC, where $f_\text{GP}$ is approximated using the trained decoder $D_\psi(\cdot)$: $f_\text{GP} \approx f_\text{PriorVAE} = D_\psi(z_d), z_d \sim \N(0, I_d).$
\end{enumerate}

\section{METHODS}
\label{sec:methods}
In this section, we introduce the PriorCVAE method which aims to enable parameter inference by explicitly conditioning the VAE on stochastic process hyperparameters. This enhancement builds upon the strengths of the PriorVAE while addressing its inability to encode hyperparameters.

\subsection{Enabling Parameter Inference}
While the PriorVAE is effective at encoding priors, it has an inherent limitation: the omission of hyperparameter encoding, which prevents the estimation of hyperparameters during Bayesian inference. This limitation arises because the encoding network, $E_\gamma(f_\text{GP})$, does not distinguish between samples generated using different values of hyperparameters. Similarly, the decoder $D_\psi(z)$ only models $f_\text{GP}$ directly using the variables $z$, and no additional condition.

In restricted simple cases this issue can be bypassed. Consider a GP $f \sim \mathcal{GP}(0, \kappa)$ with a kernel $\kappa(\cdot,\cdot)$, where the magnitude (variance) is separable from other model hyperparameters $\theta$: $\kappa(s_i, s_j) = \sigma^2 R_\theta(s_i, s_j)$. Here $R_\theta$ is, for example, the radial basis function kernel \citep[RBF,][]{broomhead1988radial}, $R_\theta(s_i, s_j) = \exp\left( -{\|s_i - s_j\|^2}/{2 \ell^2} \right)$, which has one parameter $\theta = \{\ell\}$, the lengthscale $\ell$.
Due to this separability, while training a VAE, it is sufficient to encode `standardised' draws, \ie with $\sigma=1$, so the priors to be encoded are drawn as $f_\text{std} \sim \mathcal{GP}(0, \kappa_\text{std})$, where $\kappa_\text{std}$ is simply $\kappa_\text{std}(s_i, s_j) = R_\theta(s_i, s_j)$. At the inference stage, the magnitude $\sigma^2$ can be estimated explicitly since $f = \sigma f_\text{std}$. This trick is not possible in general and even in this specific example, information about the lengthscale $\ell$ is lost during the VAE training. The issue is amplified further when working with even more complicated cases, such as non-stationary kernels. 

Our solution is a new method called PriorCVAE, pairing the PriorVAE workflow with a CVAE architecture. The encoder and decoder condition on the hyperparameters of the GPs, \ie~$c=\theta$, allowing us to generate approximate evaluations of the prior for specific hyperparameters, and ultimately to perform inference on hyperparameters. The hyperparameters may be categorical or real-valued, allowing freedom of GP kernel choice.\looseness-1

\subsection{PriorCVAE Objective}
We introduce an additional tuning parameter and use the original probabilistic formulation in the VAE rather than traditional mean squared error (MSE) as the reconstruction loss. The log-likelihood, assuming that the reconstructed sample $f_\text{PriorCVAE}$ is given by $f_\text{PriorCVAE} \sim \N(f_\text{GP}, \sigma^2_\text{vae})$, is
\begin{multline}
  -\log \N(f_\text{GP}, \sigma^2_\text{vae})=\frac{1}{2 \sigma^2_\text{vae}}\text{MSE}(f_\text{GP}, f_\text{PriorCVAE}) \\ + D \log (\sigma_\text{vae} \sqrt{2\pi}),
\end{multline} 
where $D$ is the dimensionality of $x$. Previously \citet{rybkin2021simple} have shown that $\mathcal{L} {=} D \log \sigma_\text{vae} {+} \frac{1}{2 \sigma^2_\text{vae}}\text{MSE}(f_\text{GP},  f_\text{PriorCVAE}) + \KL{q(z \mid y)}{p(z)}$ leads to improved results when optimising for $\sigma_\text{vae}$, but we did not find this approach beneficial.

Instead, we use $\sigma_\text{vae}$ as a hyperparameter, which affects the quantity of uncertainty learned by PriorCVAE, leading to better uncertainty calibration. The log-likelihood can be written as $ -\frac{1}{2 \sigma^2_\text{vae}}\text{MSE}(f_\text{GP}, f_\text{PriorCVAE})+c$, and the full objective becomes:\looseness-1
\begin{multline}
    \mathcal{L}_\text{PriorCVAE} = \frac{1}{2 \sigma^2_\text{vae}}\text{MSE}(f_\text{GP}\mid \theta, f_\text{PriorCVAE}\mid \theta) \\
    + \KL{\N(\mu_z, \sigma^2_z I_d \mid \theta)}{\N(0, I_d)}.
\label{eq:loss_priorcvae}
\end{multline}
Thus, $\sigma^2_\text{vae}$ varies the weighting of the two terms in the loss. This approach is closely related to $\beta$-VAE~\citep{higgins2017betavae} trained using the objective $\mathcal{L}_{\beta\text{-VAE}}=\frac{1}{2}\text{MSE}(y, \hat{y}) + \beta\,\KL{q(z \mid y)}{p(z)}$, but in our formulation the weighting parameter is interpretable because it is linked to the amplitude of generated samples. The resulting workflow of our PriorCVAE method is presented in \cref{alg:PriorCVAE}.

\begin{algorithm}[t!]
\caption{PriorCVAE workflow}
\begin{enumerate}
    \item Fix the spatial structure of interest $\{x_1, \dots, x_n\}$ to, \eg, a set of administrative units $B=\{B_1, \dots, B_n\}$, or an artificial computational grid $G=\{g_1, \dots, g_n\}$.
    \item Draw evaluations of a prior $f$ over the spatial structure governed by hyperparameters $\theta$: $f \sim \mathcal{GP}_\theta(\cdot)$ over $G,$ or $f \sim \mathcal{MVN}_\theta(\cdot)$ over $B$.
    \item Use the vector of realisations $f_\text{GP}=\left(f(x_1), \dots, f(x_n)\right)^\top$ as data for a CVAE to encode, conditional on hyperparameters value $c{=}\theta$. Train PriorCVAE using the loss from \cref{eq:loss_priorcvae}.\looseness-1
    \item Perform Bayesian inference with MCMC of the overarching model, including latent variables and hyperparameters $\theta$, by approximating $f_\text{GP}\mid\theta$ with $f_\text{PriorVAE}\mid\theta$ in a drop-in manner using the trained decoder $D_\psi(\cdot):$ $f_\text{GP} \mid \theta \approx f_\text{PriorVAE}\mid \theta = D_\psi(z_d,\theta), z_d \sim \N(0, I_d)$.
\end{enumerate}
\label{alg:PriorCVAE}
\end{algorithm}

\subsection{Outlook}
The last step of the PriorCVAE workflow in \cref{alg:PriorCVAE} highlights why the method is computationally efficient. GP evaluations can be computed as $f_\text{GP} \mid \theta = L_\theta z_n, \quad z_n \sim \mathcal{N}(0, I_n),$ where $L_\theta$ is a Cholesky decomposition of the covariance matrix, $K_\theta {=} L_\theta L_\theta^\top$. %
The Cholesky decomposition has $\mathcal{O}(n^3)$ cost, and must be re-calculated for different values of the parameters $\theta$ we want to infer. 

PriorCVAE, instead, performs a non-linear transformation $f_\text{PriorCVAE} \mid \theta = D_\psi(z_d, \theta)$, s.t.\ $z_d \sim \N(0, I_d)$, of a smaller vector $z_d$, with $d<n$. This is efficient to evaluate since, once trained, the parameters $\psi$ remain static. The primary computational overhead relates to matrix operations within the decoder $D_\psi(\cdot)$. For instance, for a multilayer perceptron the complexity is $\mathcal{O}(dhn)$, where $h$ is the number of hidden dimensions. In the examples shown in \cref{sec:experiments}, we use small $h$, \eg\ 1 or 2. Hence, the compexity is $\mathcal{O}(dn) \le \mathcal{O}(n^2)$, providing clear computational advantages. %

\section{EXPERIMENTS}
\label{sec:experiments}
We showcase the capability of the proposed method PriorCVAE in a series of experiments, both in terms of encoding priors efficiently as well as inference and parameter inference. In \cref{sec:exp_gp_prior}, we experiment with a Gaussian process prior and showcase the capability of PriorCVAE to encode the prior and speed up MCMC inference. In \cref{sec:exp_non_stationary_kernel}, we move to the challenging setup of non-stationary kernels and show the prior encoding capability of PriorCVAE. In \cref{sec:dw_sde}, we demonstrate the capability of PriorCVAE to encoder diffusion process priors. In \cref{sec:hiv} we demonstrate the method on real-life spatial data of HIV prevalence in Zimbabwe, and, finally, in \cref{sec:sir} we demonstrate that applications of PriorCVAE are not limited by spatial modelling by using the method to encode a solution of the SIR model---an ODE used to model infectious disease dynamics.\looseness-1
\begin{figure*}
	\centering\scriptsize
	\pgfplotsset{axis on top,scale only axis,width=\figurewidth,height=\figureheight, ylabel near ticks,ylabel style={yshift=-2pt},y tick label style={rotate=90},legend style={nodes={scale=0.8, transform shape}},tick label style={font=\tiny,scale=.8}}
	\begin{minipage}{0.4\textwidth}
	    \begin{subfigure}{\linewidth}
	        \setlength{\figurewidth}{\textwidth}
	    	\setlength{\figureheight}{.4\figurewidth}
			\input{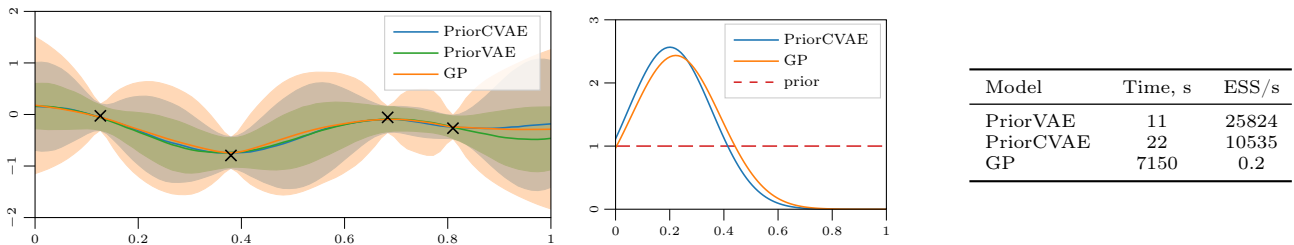}\\[-1em]
		\end{subfigure}
	\end{minipage}
	\hfill
	 \begin{minipage}{0.30\textwidth}
	 	\begin{subfigure}{\linewidth}
		 	\centering
	        \setlength{\figurewidth}{.7\textwidth}
		    \setlength{\figureheight}{.7\figurewidth}
\begin{tikzpicture}

\definecolor{color0}{rgb}{0.83921568627451,0.152941176470588,0.156862745098039}
\definecolor{color1}{rgb}{0.12156862745098,0.466666666666667,0.705882352941177}
\definecolor{color2}{rgb}{1,0.498039215686275,0.0549019607843137}

\begin{axis}[
height=\figureheight,
legend cell align={left},
legend style={fill opacity=0.8, draw opacity=1, text opacity=1, draw=white!80!black},
tick align=outside,
tick pos=left,
width=\figurewidth,
x grid style={white!69.0196078431373!black},
xmin=0, xmax=1,
xtick style={color=black},
y grid style={white!69.0196078431373!black},
ymin=0, ymax=3,
ytick style={color=black}
]
\path [draw=color0, semithick, dash pattern=on 5.55pt off 2.4pt]
(axis cs:-0.001,1)
--(axis cs:1,1);

\addplot [semithick, color1]
table {%
-0.001 1.1058778288607
0.00911111111111111 1.20070350263242
0.0192222222222222 1.29815776098686
0.0293333333333333 1.39759790478799
0.0394444444444444 1.49830447952791
0.0495555555555556 1.59948799648502
0.0596666666666667 1.70029768199726
0.0697777777777778 1.7998321717902
0.0798888888888889 1.89715200226058
0.09 1.99129368634437
0.100111111111111 2.0812851006532
0.110222222222222 2.16616185551973
0.120333333333333 2.24498427292077
0.130444444444444 2.31685456119051
0.140555555555556 2.38093375189072
0.150666666666667 2.43645795461712
0.160777777777778 2.48275349079588
0.170888888888889 2.51925048796677
0.181 2.54549455132484
0.191111111111111 2.56115617842879
0.201222222222222 2.56603764439361
0.211333333333333 2.56007715643459
0.221444444444444 2.54335015572277
0.231555555555556 2.5160677282068
0.241666666666667 2.47857217120243
0.251777777777778 2.43132984592622
0.261888888888889 2.37492152461896
0.272 2.31003051155264
0.282111111111111 2.23742887748969
0.292222222222222 2.15796219497925
0.302333333333333 2.0725331957065
0.312444444444444 1.98208479004679
0.322555555555556 1.88758289274645
0.332666666666667 1.7899994876237
0.342777777777778 1.69029633931858
0.352888888888889 1.58940972291444
0.363 1.48823649464325
0.373111111111111 1.38762177114165
0.383222222222222 1.28834842333213
0.393333333333333 1.19112852654225
0.403444444444444 1.09659684349511
0.413555555555556 1.00530635371589
0.423666666666667 0.917725783872963
0.433777777777778 0.834239040459557
0.443888888888889 0.75514640049623
0.454 0.680667278655063
0.464111111111111 0.610944361004871
0.474222222222222 0.54604887666546
0.484333333333333 0.485986768858884
0.494444444444444 0.430705525630821
0.504555555555556 0.380101437068508
0.514666666666667 0.334027059120522
0.524777777777778 0.292298682928624
0.534888888888889 0.254703631625788
0.545 0.221007232528726
0.555111111111111 0.19095934028641
0.565222222222222 0.16430031465533
0.575333333333333 0.14076638410314
0.585444444444444 0.120094352498358
0.595555555555556 0.102025630004331
0.605666666666667 0.0863095904236595
0.615777777777778 0.0727062752787654
0.625888888888889 0.0609884796805036
0.636 0.050943266498871
0.646111111111111 0.0423729636099828
0.656222222222222 0.0350957042616254
0.666333333333333 0.0289455731684804
0.676444444444444 0.0237724211670714
0.686555555555555 0.0194414095120976
0.696666666666667 0.0158323415738925
0.706777777777778 0.0128388351878895
0.716888888888889 0.0103673835756279
0.727 0.00833634693384232
0.737111111111111 0.00667491076303425
0.747222222222222 0.00532204102320355
0.757333333333333 0.00422546045823324
0.767444444444444 0.00334066507125949
0.777555555555556 0.00262999486691878
0.787666666666667 0.0020617686687789
0.797777777777778 0.00160948910330337
0.807888888888889 0.00125112071846044
0.818 0.000968441655787848
0.828111111111111 0.000746467282340635
0.838222222222222 0.000572942664411684
0.848333333333333 0.000437899671736204
0.858444444444444 0.000333273779105753
0.868555555555556 0.000252575221702265
0.878666666666667 0.000190609003049592
0.888777777777778 0.000143238296389555
0.898888888888889 0.000107185972960667
0.909 7.98692915549335e-05
0.919111111111111 5.92631566136566e-05
0.929222222222222 4.37877669869727e-05
0.939333333333333 3.22169102156595e-05
0.949444444444444 2.36035890872377e-05
0.959555555555555 1.7220084345715e-05
0.969666666666667 1.25099499879799e-05
0.979777777777778 9.04979919638003e-06
0.989888888888889 6.5190660685038e-06
1 4.67621954050669e-06
};
\addlegendentry{PriorCVAE}
\addplot [semithick, color2]
table {%
-0.001 0.96160365999595
0.00911111111111111 1.04401528260619
0.0192222222222222 1.12918054957273
0.0293333333333333 1.21665012653956
0.0394444444444444 1.30591168183352
0.0495555555555556 1.39639309384985
0.0596666666666667 1.48746707684352
0.0697777777777778 1.57845721989346
0.0798888888888889 1.66864539461528
0.09 1.75728044669017
0.100111111111111 1.84358804579204
0.110222222222222 1.92678152945524
0.120333333333333 2.00607354028524
0.130444444444444 2.08068822410038
0.140555555555556 2.14987373043415
0.150666666666667 2.21291473750386
0.160777777777778 2.26914471222881
0.170888888888889 2.31795761286427
0.181 2.35881874770956
0.191111111111111 2.39127451822453
0.201222222222222 2.41496079847867
0.211333333333333 2.42960973455236
0.221444444444444 2.43505478637824
0.231555555555556 2.43123387932902
0.241666666666667 2.41819058216103
0.251777777777778 2.39607328005897
0.261888888888889 2.36513236472843
0.272 2.32571551593598
0.282111111111111 2.27826119882014
0.292222222222222 2.22329054701406
0.302333333333333 2.16139784162877
0.312444444444444 2.0932398291759
0.322555555555556 2.0195241465746
0.332666666666667 1.94099713782064
0.342777777777778 1.85843135436896
0.352888888888889 1.77261302980636
0.363 1.68432980931805
0.373111111111111 1.59435899643346
0.383222222222222 1.5034565544975
0.393333333333333 1.41234706939849
0.403444444444444 1.32171484460185
0.413555555555556 1.23219626090247
0.423666666666667 1.14437349297206
0.433777777777778 1.058769634174
0.443888888888889 0.975845241604223
0.454 0.895996276122118
0.464111111111111 0.819553378323088
0.474222222222222 0.746782391836668
0.484333333333333 0.677886020656472
0.494444444444444 0.613006487836219
0.504555555555556 0.552229049010134
0.514666666666667 0.495586205790771
0.524777777777778 0.443062460942251
0.534888888888889 0.394599458932218
0.545 0.350101361504099
0.555111111111111 0.30944031765136
0.565222222222222 0.272461900118173
0.575333333333333 0.23899039556318
0.585444444444444 0.208833852071023
0.595555555555556 0.181788805073382
0.605666666666667 0.157644620293644
0.615777777777778 0.136187409474486
0.625888888888889 0.117203490889105
0.636 0.100482381572368
0.646111111111111 0.0858193215338901
0.656222222222222 0.0730173417237979
0.666333333333333 0.0618888970989056
0.676444444444444 0.0522570937511092
0.686555555555555 0.0439565447531401
0.696666666666667 0.0368338932528067
0.706777777777778 0.0307480435568798
0.716888888888889 0.0255701416764336
0.727 0.0211833462658505
0.737111111111111 0.0174824292979611
0.747222222222222 0.0143732433989551
0.757333333333333 0.0117720897320385
0.767444444444444 0.00960501686712405
0.777555555555556 0.00780707738427087
0.787666666666667 0.00632156518727713
0.797777777777778 0.00509925278203732
0.807888888888889 0.00409764420787572
0.818 0.0032802559801788
0.828111111111111 0.00261593536716393
0.838222222222222 0.0020782226194015
0.848333333333333 0.001644761416225
0.858444444444444 0.00129675979138257
0.868555555555556 0.00101850214160971
0.878666666666667 0.000796911586956615
0.888777777777778 0.000621160914409121
0.898888888888889 0.000482329565745354
0.909 0.000373103593342924
0.919111111111111 0.000287515169749043
0.929222222222222 0.000220718064900059
0.939333333333333 0.000168795467351711
0.949444444444444 0.000128596593724516
0.959555555555555 9.75986777858385e-05
0.969666666666667 7.37911344891822e-05
0.979777777777778 5.5578935551487e-05
0.989888888888889 4.17024957424169e-05
1 3.11716400570758e-05
};
\addlegendentry{GP}
\addplot [semithick, color0, dashed]
table {%
-1 -1
};
\addlegendentry{prior}
\end{axis}

\end{tikzpicture}
		\end{subfigure}
	\end{minipage}
	\hfill
	\noindent
	 \begin{minipage}[c]{0.25\textwidth}
		\centering
			 \begin{tabularx}{\linewidth}{lcc}
			     \toprule 
			     Model & Time, s & ESS/s \\
			     \midrule 
			     PriorVAE & 11 & 25824\\ 
			     PriorCVAE & 22 & 10535\\ 
			     GP &  7150& 0.2 \\ 
			     \bottomrule
			 \end{tabularx}
    \end{minipage}
    \caption{MCMC inference comparison between PriorCVAE, PriorVAE, and GP. Left: Estimated means and uncertainty bounds obtained by the three models; PriorCVAE is much closer to the baseline(GP) than PriorVAE. Middle: Lengthscale posteriors of the GP and PriorCVAE model. Right: Comparison of inference statistics.}
    \label{fig:gp_mcmc_inference}
    \vspace*{-1em}
\end{figure*}

\subsection{Gaussian Process Prior}
\label{sec:exp_gp_prior}
We experiment with Gaussian process prior kernels and encode it using both PriorVAE \citep{semenova2022priorvae} and the proposed model, PriorCVAE. We compare against PriorVAE and show the superior prior encoding capability of PriorCVAE. Then, we move to the inference and parameter inference task and compare against the baseline, GP using MCMC inference.
\paragraph{Encoding prior} We encode the priors evaluated over a regular one-dimensional grid of $n {=} 80$ points over the interval $[0, 1]$. Training samples are drawn from a GP with zero mean and standardised ($\sigma^2{=}1$) kernel $\kappa^{\text{Mat\'ern-\nicefrac{5}{2}}}_{\ell}$. 
The model has the following hierarchical structure with a hyperprior $\mathcal{U}$ on the lengthscale $\ell$ of the kernel, which also acts as the condition $c$ of the CVAE model
\begin{align}
	 \begin{cases}
		    \ell &\sim \mathcal{U}(0.01,0.99),\\
		    f \mid \ell &\sim \mathcal{GP}(0,\kappa^{\text{Mat\'ern-\nicefrac{5}{2}}}_{\ell}(\cdot, \cdot)).
	\end{cases}    
	\label{eq:gp_encoding_prior_model}
\end{align}
\cref{exp:gp1d_contin_details} provides neural network training details. \cref{fig:exp_matern52_encode_prior} showcases three sets of priors: GP, PriorCVAE, and PriorVAE for two lengthscales $\ell{=}\{0.1, 1.0\}$. From the plot, we can observe that the samples from PriorCVAE closely resemble to the GP samples in comparison with PriorVAE whose samples are not dependent on the lengthscale and are thus inflexible. We also present the empirical covariance in \cref{fig:exp_matern52_encode_prior}.  

\paragraph{MCMC inference} A ground truth curve was generated using the RBF kernel with the lengthscale $l_\text{true}=0.2$. To generate observed data, we selected four locations and added a random amount of noise at each of the locations, distributed as $s \sim \mathcal{N}^+(0.1)$. We fit three models to this data using the NUTS sampler using the three different priors: $f_\text{GP-RBF}$, $f_\text{PriorVAE-RBF}$ and $f_\text{PriorCVAE-RBF}$. \cref{fig:gp_mcmc_inference} presents estimated means and Bayesian credible intervals (BCIs) obtained by the three models. All the estimated means are close to each other, but PriorCVAE's BCIs are much closer to the GP's BCIs than the BCIs from PriorVAE. All three models were run using 5000 warm-up steps, 50000 post warm-up iterations\footnote{PriorVAE and PriorCVAE models require much fewer iterations for convergence. These high values are required for the GP model to converge and, hence, have been set equal for all three models.} and three chains. Evaluation of run times, number of effective samples and number of effective samples per second are also presented in \cref{fig:gp_mcmc_inference}. The number of effective steps of PriorCVAE per second is order of 10K times higher than the one of the original GP model, while the lengthscale parameter can also be recovered. \cref{fig:gp_mcmc_inference} presents posterior estimates of the lengthscale parameter produced by PriorCVAE and GP models, which are closely matched. 

\paragraph{Comparison with Laplace approximation and ADVI} Using the same implementation software NumPyro~\citep{phan2019composable, bingham2019pyro} as for the above models, we performed inference using popular approximation techniques - Laplace approximation and automatic differentiation variational inference \citep[ADVI,][]{kucukelbir2017automatic}. Results presented in \cref{fig:NUTS_Laplace_ADVI_gp} show that ADVI failed at estimating both the mean and the lengthscale; Laplace approximation is able to estimate the mean well, but produced no characterisation of uncertainty, as well as only a point estimate for the lengthscale.%

\paragraph{Encoding discrete priors and function properties}
It is also possible to encode and infer priors over a discrete set of values. Example with a binary prior on the lengthscale is presented in \cref{exp:gp1d_n100_binary_details}.

Similarly to other prior-encoding VAEs, PriorCVAE is able to encode function properties alongside function realisations. The integral $\mathcal{I} = \int_D \exp(f(s))ds, s\in D$ is a crucial quantity required to evaluate log-Gaussian Cox process (LGCP) likelihood~\citep{moller1998log}. In \cref{sec:lgcp_details} we demonstrate how this function property can be learnt with PriorCVAE.

\subsection{Encoding Non-stationary Kernels}
\label{sec:exp_non_stationary_kernel}
We show for the first time that prior-encoding VAEs are also able to learn non-stationary GPs. Consider a kernel which is a product of linear and RBF kernels:
\begin{align}
&\kappa(s_i, s_j) = \kappa_\text{lin}(s_i, s_j) \, \kappa_\text{rbf}(s_i, s_j) 
 \nonumber \\
 &\quad=(s_i - c_\text{lin})^{\top}(s_j - c_\text{lin})  \exp \left( - \frac{\|s_i- s_j\|^2}{ 2\ell^2} \right).
\end{align}
We fix $c_\text{lin}{=}0.4$ and aim to encode the lengthscale as the condition. We also set a narrower prior on the lengthscale:  $\ell {\sim} \mathcal{U}(0.01, 0.4)$, instead of $\ell {\sim} \mathcal{U}(0.01, 0.99)$, in order to test PriorCVAE's extrapolation abilities. Training details are reported in \cref{sec:lin_rbf_details}. Visual assessment of the quality of the learned priors is presented in \cref{fig:exp_non_stationary_gp}, which corresponds to lengthscales $\ell{=}0.05$ and $\ell{=}0.2$ contained in the $(0.01, 0.4)$ interval along with the empirical covariance matrices.
\paragraph{Extrapolation of hyperparameters} 
Even though we only trained the decoder with lengthscales drawn from $(0.01, 0.4)$ interval, it is possible to condition PriorCVAE draws on lengthscales $\ell$ which lie outside of this interval. \cref{fig:lin_rbf_priors_05_09} compares the quality of extrapolation for $\ell=0.5$ and $\ell=0.9$, showing that the prior quality is high close to the interval, and deteriorates for $\ell$ further away.

\subsection{Encoding diffusion processes}
\label{sec:dw_sde}
We experiment with encoding priors of a diffusion process (DP) double-well, which is defined by an SDE:
\begin{equation}
	\mathrm{d} x_t = \theta_1 x_t (\theta_2 - x_t^2) + \mathrm{d} \beta_t \, ,
\end{equation}
with $\theta_1, \theta_2$ being the parameters of the DP and $Q$ is the spectral density. The marginal state distributions have two modes that sample state trajectories keep visiting through time. In \cref{app:dw_experiment}, we discuss the setup and show how the proposed PriorCVAE model can also be used to encode the DP priors, which can then be used to speed-up inference in a model with such DP prior.

\subsection{HIV Prevalence in Zimbabwe}
\label{sec:hiv}
We consider household survey data from the 2015 to 2016 Population-based HIV Impact Assessment Survey in Zimbabwe \citep{sachathep2021population}. The observed positive cases $y_i$ among all observed cases $n_i$ in each administrative unit $B_i, i=\{1,\ldots, 63\}$ are modelled as follows:
$y_i \sim \text{Binomial}(n_i, \theta_i)$, $\text{logit}^{-1}(\theta_i) {=} b_0 + f_i$,
${f \sim \mathcal{MVN}(0, K)}$,
${\kappa_{ij} {=} \sigma^2 \exp \left( - \frac{||c_i - c_j||^2}{2\ell^2} \right)}$,
${b_0 {\sim} \mathcal{N}(0,1)}$,
${\ell {\sim} \text{Gamma}(2,4)}$, and
${\sigma \sim \text{Gamma}(1.5, 1.5)}$.
Here $\theta_i$ is the estimate of HIV prevalence in unit $i$, and $\phi=(f_1, \ldots, f_{63})$ is the spatial random effect, a GP with RBF kernel evaluated at the centroids of $B_i$. Neural network training details are available in \cref{sec:hiv_details}.%

\textbf{MCMC inference} We performed inference %
using both the original GP prior $f_\text{GP}$ and the trained $f_\text{PriorCVAE}$ for $\phi$. MCMC inference was performed using $n_\text{warmup}=2000$ burn-in steps, and $n_\text{samples}=10000$ post-warmup steps with the NUTS sampler. Run times and sampling efficiency are presented in Table on \cref{fig:teaser}. Estimates of parameters $l$ and $\sigma$ are shown on \cref{fig:zimbabwe_estimates_params}, comparison of estimated prevalence obtained using the original $f_\text{GP}$ prior, and the model with the $f_\text{PriorCVAE}$ prior are presented on \cref{fig:zimbabwe_estimates_prev} and resulting maps of estimated prevalence are presented on \cref{fig:teaser}.

\textbf{Comparison with R-INLA} R-INLA \citep{lindgren2015bayesian} is a popular software for Bayesian spatial modelling and is viewed as state-of-the-art in many applied fields. While being very fast, R-INLA has limitations, such as, for example, it does not permit the RBF kernel, or Mat\'ern kernels with smoothness higher than $\nu {=} \alpha - m/2$, where $\alpha \in (0,2]$ and $m$ is space dimensionality. We discuss the comparison of the proprosed PriorCVAE tool with R-INLA in \cref{sec:hiv_details}.

\subsection{Encoding ODE Solutions: SIR Example}
\label{sec:sir}
All of the previous examples were concerned with encoding GP priors. In this section we demonstrate that also ODE solutions can be encoded using the same technique on the example of the SIR model.
The SIR model is a classical model from mathematical epidemiology \citep{kermack1927contribution} that divides a population into three compartments (Susceptible, Infected, and Recovered) to study the spread of infectious diseases by modelling time-dependent volumes of compartments as solutions of a system of differential equations:\looseness-1
\begin{equation*}
\dot{S} = -\beta \cdot \frac{S \cdot I}{N} \quad {,} \quad  \dot{I} = \beta \cdot \frac{S \cdot I}{N} - \gamma \cdot I \quad {,} \quad \dot{R} = \gamma \cdot I.
\end{equation*}
Here $S$ is the number of susceptible individuals,
$I$ is the number of infected individuals,
$R$ is number of recovered individuals, 
$N$ is the total population size,
$\beta$ is the transmission rate and
$\gamma$ is the recovery rate. 

\begin{figure*}[t!]
	\begin{minipage}[c]{.33\textwidth}
		\centering\scriptsize
		\pgfplotsset{ylabel near ticks, legend style={nodes={scale=0.8, transform shape}},tick label style={font=\tiny,scale=.8}}
		\setlength{\figurewidth}{\textwidth}
		\setlength{\figureheight}{\figurewidth}
\begin{tikzpicture}

\definecolor{color0}{rgb}{1,0.647058823529412,0}
\definecolor{color1}{rgb}{0.12156862745098,0.466666666666667,0.705882352941177}

\begin{axis}[
height=\figureheight,
legend cell align={left},
legend style={fill opacity=0.8, draw opacity=1, text opacity=1, draw=white!80!black},
tick align=outside,
tick pos=left,
width=\figurewidth,
x grid style={white!69.0196078431373!black},
xlabel={Day},
xmin=-0.65, xmax=13.65,
xtick style={color=black},
y grid style={white!69.0196078431373!black},
ylabel={\#Infected},
ymin=-24.955, ymax=524.055,
ytick style={color=black}
]
\path [draw=color0, fill=color0, opacity=0.4]
(axis cs:0,0)
--(axis cs:0,0)
--(axis cs:1,1)
--(axis cs:2,9)
--(axis cs:3,38)
--(axis cs:4,104)
--(axis cs:5,153.95)
--(axis cs:6,128.95)
--(axis cs:7,94)
--(axis cs:8,62)
--(axis cs:9,40)
--(axis cs:10,24)
--(axis cs:11,15)
--(axis cs:12,8)
--(axis cs:13,4)
--(axis cs:13,4)
--(axis cs:13,25)
--(axis cs:12,39)
--(axis cs:11,60.05)
--(axis cs:10,94)
--(axis cs:9,144.05)
--(axis cs:8,215)
--(axis cs:7,317.05)
--(axis cs:6,413.05)
--(axis cs:5,499.1)
--(axis cs:4,397.05)
--(axis cs:3,160)
--(axis cs:2,39.05)
--(axis cs:1,10)
--(axis cs:0,3)
--cycle;

\path [draw=color1, fill=color1, opacity=0.4]
(axis cs:0,0)
--(axis cs:0,0)
--(axis cs:1,1)
--(axis cs:2,8)
--(axis cs:3,27)
--(axis cs:4,66)
--(axis cs:5,97)
--(axis cs:6,92)
--(axis cs:7,72.95)
--(axis cs:8,47)
--(axis cs:9,31)
--(axis cs:10,17)
--(axis cs:11,13)
--(axis cs:12,9)
--(axis cs:13,5)
--(axis cs:13,5)
--(axis cs:13,38)
--(axis cs:12,57)
--(axis cs:11,76)
--(axis cs:10,104)
--(axis cs:9,154)
--(axis cs:8,245)
--(axis cs:7,340.15)
--(axis cs:6,434.05)
--(axis cs:5,462.05)
--(axis cs:4,351.1)
--(axis cs:3,185)
--(axis cs:2,65)
--(axis cs:1,13)
--(axis cs:0,3)
--cycle;

\addplot [semithick, color0]
table {%
0 0.978000044822693
1 4.91700029373169
2 22.113000869751
3 86.5540008544922
4 227.397003173828
5 309.524017333984
6 266.646026611328
7 195.504013061523
8 133.563003540039
9 85.0970077514648
10 54.003002166748
11 34.8030014038086
12 21.8730010986328
13 13.5900011062622
};
\addlegendentry{ODE}
\addplot [semithick, color1]
table {%
0 1.14300000667572
1 6.03200006484985
2 29.6740016937256
3 88.818000793457
4 189.230010986328
5 255.436004638672
6 242.743011474609
7 189.279006958008
8 130.833999633789
9 83.3720016479492
10 55.7820014953613
11 40.0920028686523
12 28.7460021972656
13 19.3190002441406
};
\addlegendentry{PriorCVAE}
\addplot [semithick, black, mark=x, mark size=3, mark options={solid}, only marks]
table {%
0 3
1 8
2 26
3 76
4 225
5 298
6 258
7 233
8 189
9 128
10 68
11 29
12 14
13 4
};
\addlegendentry{data}
\end{axis}

\end{tikzpicture}
		\vspace*{-1em}
		\caption{Influenza outbreak SIR model: PriorCVAE model gives inference similar to the exact ODE solution.}
		\label{fig:sir_fit}
	\end{minipage}
	\hfill
	\begin{minipage}[c]{.65\textwidth}
		\centering\scriptsize
		\pgfplotsset{axis on top,scale only axis,width=\figurewidth,height=\figureheight, ylabel near ticks,ylabel style={yshift=6pt},y tick label style={rotate=90},yticklabels={},xticklabels={}} 
		\setlength{\figurewidth}{.14\textwidth}
		\setlength{\figureheight}{\figurewidth}
		\newcommand{\mytitle}[1]{\tikz\node[minimum width=\figurewidth,minimum height=2em,align=center]{#1};}
		\begin{subfigure}{.25\textwidth}  
			\raggedleft
			\mytitle{\textbf{GP} ($\ell{=}0.05$)}    
			\pgfplotsset{ylabel={Output, $y$}}
			\tikz[outer sep=0pt, inner sep=0pt]\node{\input{figures/non_stationary/GP_samples_0_05.tex}};
		\end{subfigure}
		\hfill
		\begin{subfigure}{.24\textwidth}
			\raggedleft
			\mytitle{\textbf{PriorCVAE}}
			\tikz[outer sep=0pt, inner sep=0pt]\node{\input{figures/non_stationary/VAE_samples_0_05.tex}};
		\end{subfigure}
		\hfill
		\begin{subfigure}{.24\textwidth}
			\raggedleft
			\mytitle{\textbf{GP} ($\ell{=}0.2$)}
			\tikz[outer sep=0pt, inner sep=0pt]\node{\input{figures/non_stationary/GP_samples_0_2.tex}};
		\end{subfigure}
		\hfill
		\begin{subfigure}{.24\textwidth}
			\raggedleft
			\mytitle{\textbf{PriorCVAE}}
			\tikz[outer sep=0pt, inner sep=0pt]\node{\input{figures/non_stationary/VAE_samples_0_2.tex}};
		\end{subfigure}\\[1em]

		\newcommand{\addxy}[1]{\tikz[node distance=5pt,inner sep=0]{\node(a){\includegraphics[width=\figurewidth]{figures/non_stationary/#1}};\node[left=of a,rotate=90,anchor=center]{Input, $x'$};\node[below=of a,anchor=center]{Input, $x$};\node[draw,minimum width=\figurewidth,minimum height=\figurewidth] at (a) {};}}
		\newcommand{\addx}[1]{\tikz[node distance=5pt,inner sep=0]{\node(a){\includegraphics[width=\figurewidth]{figures/non_stationary/#1}};\node[below=of a,anchor=center]{Input, $x$};\node[draw,minimum width=\figurewidth,minimum height=\figurewidth] at (a) {};}}
		\begin{subfigure}{.25\textwidth}  
			\raggedleft  
			\addxy{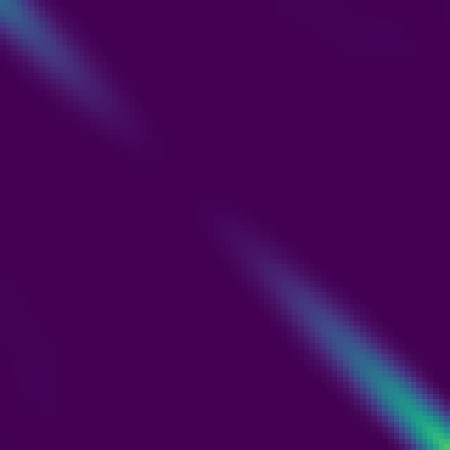}
		\end{subfigure}
		\hfill
		\begin{subfigure}{.24\textwidth}
			\raggedleft
			\addx{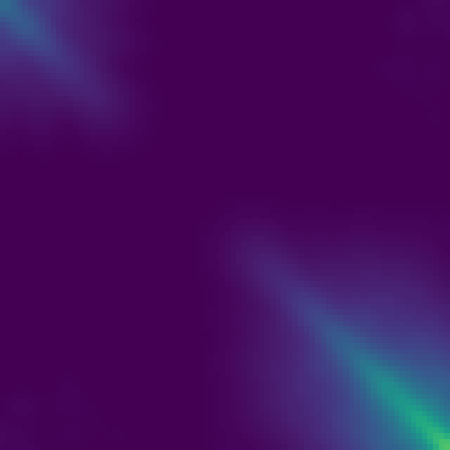}
		\end{subfigure}
		\hfill
		\begin{subfigure}{.24\textwidth}
			\raggedleft
			\addx{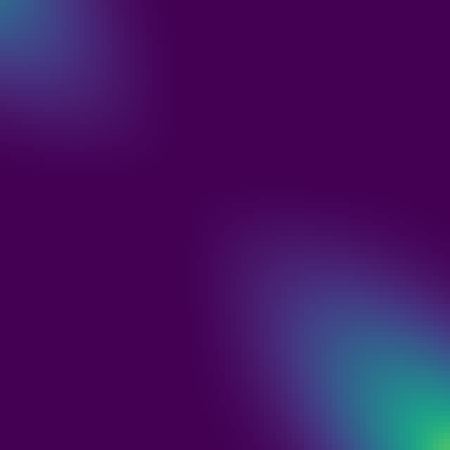}
		\end{subfigure}
		\hfill
		\begin{subfigure}{.24\textwidth}
			\raggedleft
			\addx{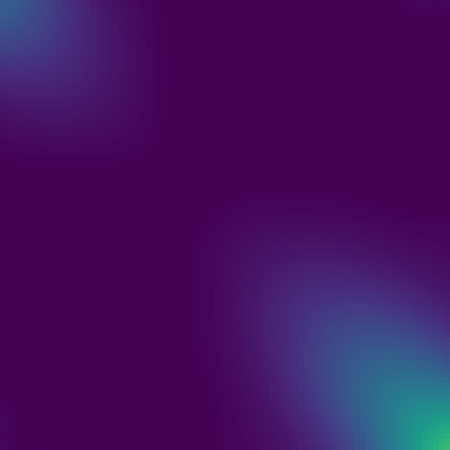}
		\end{subfigure}
		\newcommand{\cbar}{\protect\includegraphics[width=2.5em,height=.7em]{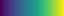}}
		\caption{Non-stationary kernel: Example draws (top) from the GP and the PriorCVAE model, and empirical covariance matrices (bottom 0~\cbar~1) with lengthscale $\ell=0.05$ (two left-hand columns) and $\ell=0.2$ (two right-hand columns).}
		\label{fig:exp_non_stationary_gp}
\end{minipage}
\vspace*{-1em}
\end{figure*}
\textbf{Encoding priors}
Initial datum $(S_0, I_0, R_0)$ and parameters $\beta, \gamma$ define a unique solution of the ODE. Inference is then performed by matching observed number of infected individuals $y(t)$ and the expected number of infected individuals $I(t)$ via, for example, a count distribution, such as Poisson or Negative Binomial: $y(t) \sim \text{NegBin}(I(t), \phi),$ where $\phi$ is the overdispersion parameter. We rely in this example on the data of influenza A (H1N1) outbreak in 1978 at a British boarding school. The data consists of the daily number of students in bed, spanning over a time interval of 14 days. There were $N{=}763$ male students and it is reported that one infected boy started the epidemic $I_0=1$. The data are freely available in the R package \textit{outbreaks}, maintained as part of the R Epidemics Consortium.\looseness-1

\textbf{PriorCVAE training} Since ODE solution is unique given the initial conditions, and only one compartment ($I$) is used for inference, it is sufficient to train the neural network on the draws of $I(t)$. Since $I(t)$ needs to lie between 0 and $N$, at the last layer we apply the sigmoid activation function and multiply the result by $N$. For training details see \cref{sec:sir_details}. \cref{fig:sir_priors} demonstrates the family of learnt trajectories of the infected compartment parameterised by the condition $c {=} (\beta, \gamma)$.

\textbf{MCMC inference} For inference, we used the same priors as in \cite{grinsztajn2021bayesian}: $\beta \sim \mathcal{N}^+(2, 1), \gamma \sim \mathcal{N}^+(0.4, 0.5), \phi^{-1} \sim \mathcal{\text{Exp}}(5)$ and initial datum $(S_0, I_0, R_0) = (N-1, 1, 0).$ \cref{fig:sir_fit} shows inference results under the original and PriorCVAE-based models and \cref{fig:sir_params} shows parameter estimates.\looseness-1
\section{DISCUSSION AND CONCLUSION}
\label{sec:discussion}
In this paper, we introduce PriorCVAE, a method based on deep generative modelling that enables explicit parameter estimation while performing fast and efficient MCMC inference. It covers a gap in the literature on encoding finite evaluations of stochastic processes, where previous methods lost information about underlying parameters. %
We solve the problem by conditioning the generative process on the parameters of interest and demonstrate that the conditioning can be performed both by categorical and continuous parameters. PriorCVAE does this while retaining the advantages of PriorVAE at the inference stage, \ie much shorter computation times and higher effective samples size than an equivalent model fully inferred with MCMC. %

Side-by-side comparison of the GP, PriorVAE, and PriorCVAE models has shown that although PriorCVAE is less efficient than PriorVAE, it can better reconstruct the underlying model; furthermore, the number of effective steps per second achieved by PriorCVAE is still about 10K higher than for the GP. Using non-stationary GP example, we demonstrated that it is possible to perform extrapolation with respect to hyperparameters; that the quality of reconstructed priors remains high if the out-of-sample hyperparameter values are close to those in-sample and the quality of reconstruction deteriorates with distance. We quantified the quality of the learnt PriorVAE and PriorCVAE priors by measuring distance of their empirical covariance matrices to the one of the original GP prior.  \cref{tab:frob_norms} showed that this form of quantification agrees with the results obtained at inference, \ie that PriorCVAE priors match GP priors better than PriorVAE. Future work includes variations in the reconstruction loss, using, for example, Frobenius norm or maximum mean discrepancy \citep{gretton2012kernel} to match higher moments by design. PriorCVAE is kernel-agnostic, which gives it a serious competitive advantage over state-of-the-art methods such as R-INLA \citep{rue2017bayesian}, which only allows Mat\'ern kernels of smoothness up to $\nu{=}1$ in two dimensions, not RBF kernels or products of non-Mat\'ern kernels.\looseness-1 %

In the final example, we demonstrate that PriorCVAE is agnostic not only to the kernel of a GP, but to the prior model as a whole. This model can be of mechanistic nature, such as a system of differential equations.\looseness-1

The key features of PriorCVAE --- speed, efficiency, and ability to estimate parameters within MCMC --- %
have great potential to make a major impact in spatial statistics and related domains requiring inference over correlated structures. Immediate potential applications include time series, as well as a wider array of kernels.

\clearpage
\phantomsection%
\addcontentsline{toc}{section}{References}
\begingroup
\small
\bibliographystyle{abbrvnat}

\endgroup

\clearpage

\newcommand{\apptitle}[1]{%
  \hsize\textwidth
  \linewidth\hsize \toptitlebar {\centering
  {\Large\bfseries #1 \par}}
 \bottomtitlebar \vskip 0.2in
}

\onecolumn
\appendix
\apptitle{PriorCVAE: Scalable MCMC Parameter Inference with Bayesian Deep Generative Modelling:
Supplementary Materials}
\thispagestyle{empty}
\renewcommand{\thetable}{A\arabic{table}}
\renewcommand{\thefigure}{A\arabic{figure}}
\renewcommand{\theequation}{A\arabic{equation}}

This Supplementary document is organised as follows. \ref{exp:gp1d_contin_details} provides details of continuous condition PriorVAE and PriorCVAE models. \cref{exp:gp1d_n100_binary_details} demonstrates example of a binary prior on the GP lengthcsale.
\cref{sec:lgcp_details} demonstrates how to encode integral of the intensity function of the log-Gaussian Cox process. \cref{sec:lin_rbf_details} provides details of the example with non-stationary kernels. \cref{sec:hiv_details} covers details of the real-life data example of HIV Prevalence in Zimbabwe Estimation, including comparison of MCMC, PriorCVAE and R-INLA results. \cref{sec:sir_details} and \cref{app:dw_experiment} describe details of models encoding SIR and double-well trajectories, correspondingly. \cref{sec:software_hardware} explains software, hardware and reproducibility details. \cref{sec:numpyro_example} shows a sample NumPyro code. \cref{sec:suppl_figures} contains supplementary figures.\looseness-1

\section{Continuous Condition PriorVAE and PriorCVAE: Experiment Details}\label{exp:gp1d_contin_details}

\textbf{Model for priors}
\begin{align}
 \begin{cases}
    \ell &\sim \mathcal{U}(0.01,0.99),\\
    f \mid \ell &\sim \mathcal{GP}(0, \kappa^\text{RBF}_\ell(\cdot, \cdot)).
\end{cases}    
\label{eq:continuous_priors}
\end{align}

\begin{figure}[h!]
\begin{center}    
    \includegraphics[width=0.5\textwidth]{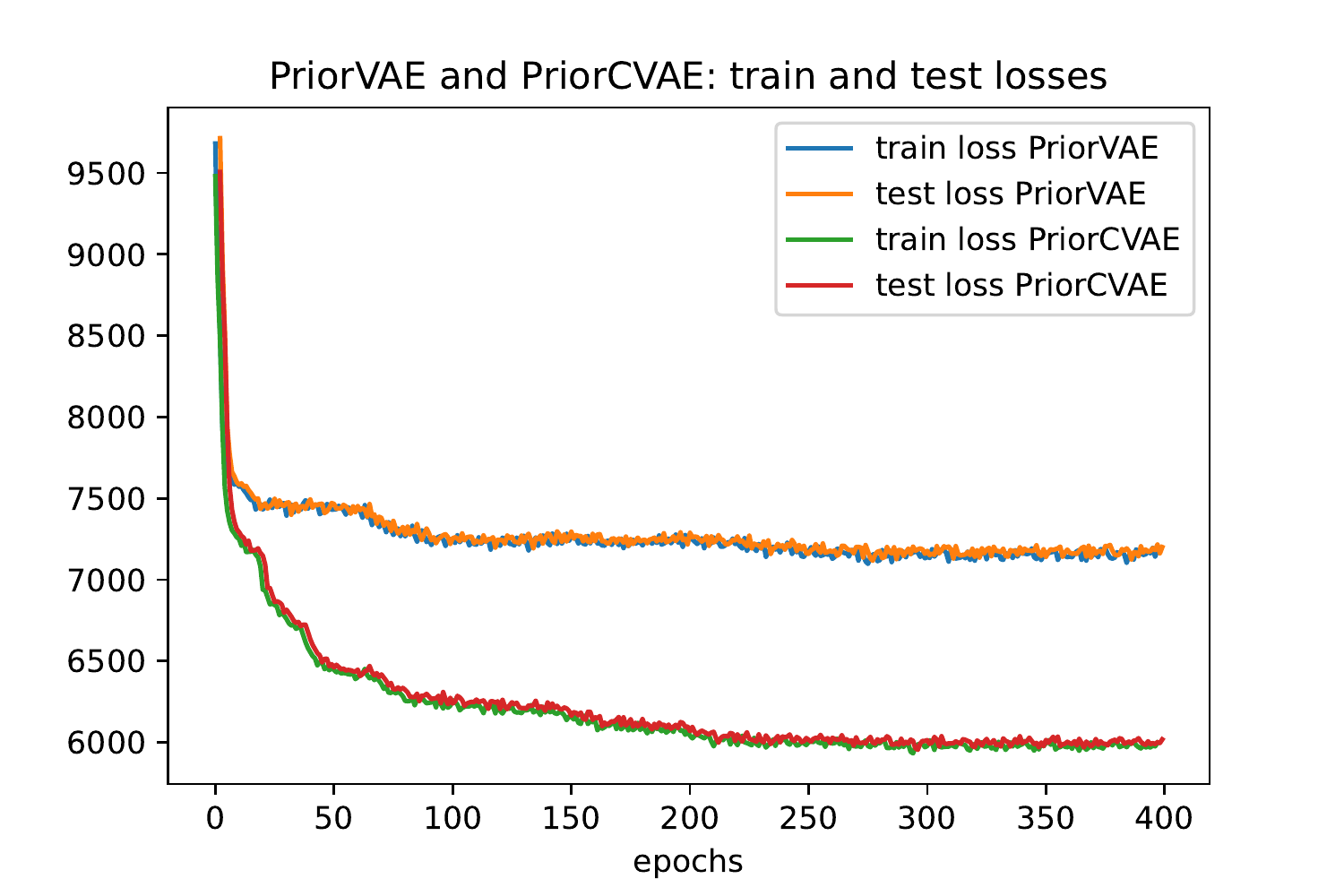}
\end{center}    
\caption{Training and test losses of PriorVAE and PriorCVAE models.}
\label{fig:cvae_train_test_losses}
\end{figure}

\textbf{Training PriorVAE and PriorCVAE}. We create training and test datasets, each of 100000 randomly sampled GP prior evaluations according to Equations \ref{eq:continuous_priors}. In order to perform one-to-one comparison of PriorVAE and PriorCVAE we opt to keep the train and test sets fixed (as opposed to generating data on the fly during the training process). The architecture of the two models is also identical: both the encoder and the decoder are MLPs with one hidden layer and $n_\text{input}=80$ input (output for the decoder) nodes, $n_\text{h}=60$ hidden nodes, and the dimension of the latent space is $n_\text{z}=40.$ Both models were trained for $n_\text{epochs}=500$ epochs, with $n_\text{batch}=2000$ batch size, and $1e^{-3}$ learning rate using the Adam optimiser. Comparing the training and test losses already provides insights into which model can describe the data better (\cref{fig:cvae_train_test_losses}): both the test and training losses of PriorCVAE are lower than those for PriorVAE throughout the training process.

\begin{table}[h]
\caption{Quantification of the quality of the learned priors: Frobenius norms measure the distance between empirical covariance matrices obtained from the original GP prior, and learned PriorVAE and PriorCVAE priors $F_\text{PriorVAE} = \|\mathbf{K}_\text{GP} - \mathbf{K}_\text{PriorVAE}\|_F,$ $F_\text{PriorCVAE} = \|\mathbf{K}_\text{GP} - \mathbf{K}_\text{PriorCVAE}\|_F$.}
\begin{center}
	\begin{tabular}{lcccc}
		\hline 
		lengthscale $c=\ell$ &0.05 & 0.1& 0.3  & 0.9  \\
		\hline 
		$F_\text{PriorVAE}$ & 45.47 &37.48& 8.29 &  22.94 \\ 
		$F_\text{PriorCVAE}$ & 14.45 & 7.63 &  6.42 & 1.99 \\ 
	\hline 
	\end{tabular}
\end{center}
\label{tab:frob_norms}
\end{table}

\section{Encoding Lengthscale of a GP as a Binary Condition: experiment details}
\label{exp:gp1d_n100_binary_details}

\textbf{Model for priors}
\begin{align}
 \begin{cases}
    c  &\sim \text{Bernoulli}(0.5),\\
    \ell \mid c & = (1-c) \ell_1 + c \, \ell_2,\\
    f \mid \ell &\sim \mathcal{GP}(0, \kappa^\text{RBF}_\ell(\cdot, \cdot))
\end{cases}    
\label{eq:binary_priors}
\end{align}

\textbf{Neural network training details}. We create training and test datasets, each of 100000 randomly sampled GP prior evaluations according to Equations \ref{eq:binary_priors}. We follow the architecture used in \cite{semenova2022priorvae}, i.e. a multilayer perceptron (MLP). However, our experiments showed that even one hidden layer is enough to achieve performance similar to the one with two or more hidden layers under this architecture. Both the encoder and the decoder are MLPs and contain one hidden layer of $n_\text{h}=70$ nodes, and the dimension of the latent space is $n_\text{z}=50$. Note, of course, that the input layer of the encoder and the output layer of the encoder both have $n_\text{input}=100$ nodes, the same size as $x$. The model was trained for $n_\text{epochs}=250$ epochs, with $n_\text{batch}=1000$ prior evaluations in each batch, and a learning rate of $1e^{-3}$, using the Adam optimiser~\citep{kingma2014adam}. The $\sigma^2_\text{vae}$ hyperparameter was set to 0.9. Training of the neural network took 9 minutes\footnote{See Section \ref{sec:software_hardware} for hardware details.}. 

\begin{figure}[h!]
\begin{center}
    \includegraphics[width=0.95\textwidth]{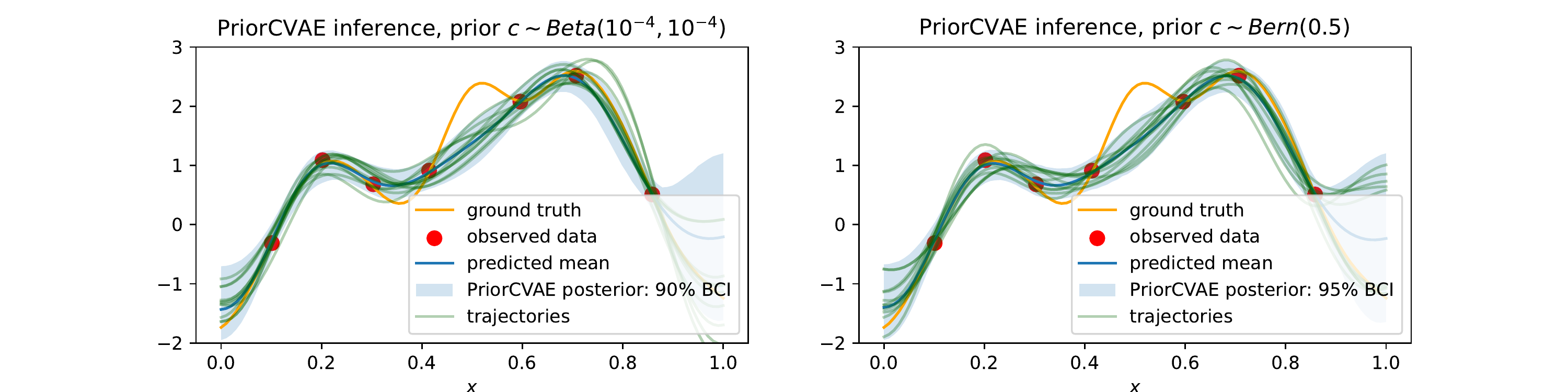}
    \includegraphics[width=0.95\textwidth]{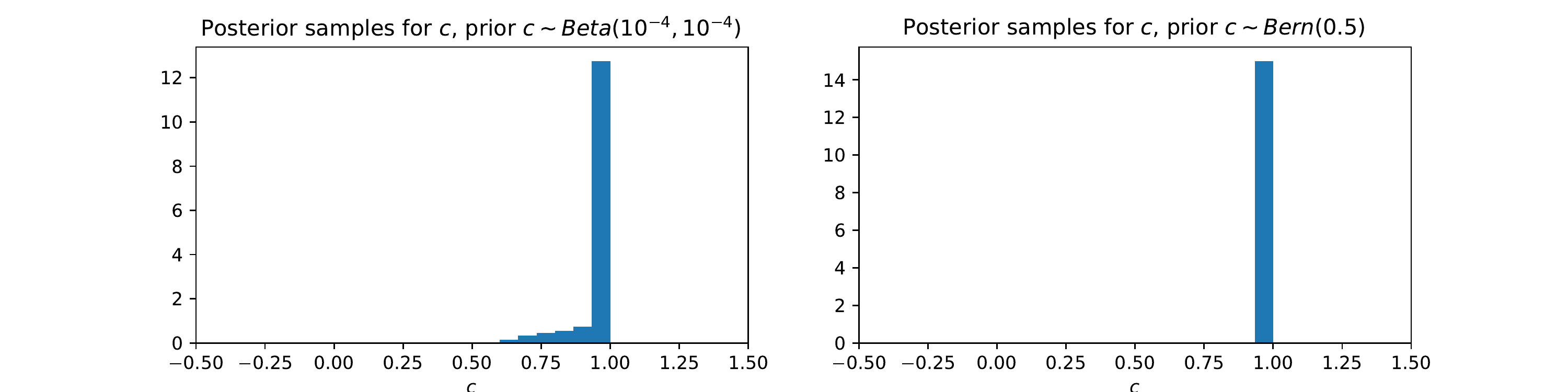}
\end{center}    
\caption{Left: inference performed using NUTS and PriorCVAE with a binary condition, approximating the true Bernoulli prior with a continuous Beta prior. Right: inference performed using MixedHMC and PriorCVAE with a binary condition, inferred as a discrete variable with a Bernoulli prior.}
\label{fig:binary_inference_Beta_Bern}
\end{figure}

\begin{figure}[h!]
\begin{center}
    \includegraphics[width=0.95\textwidth]{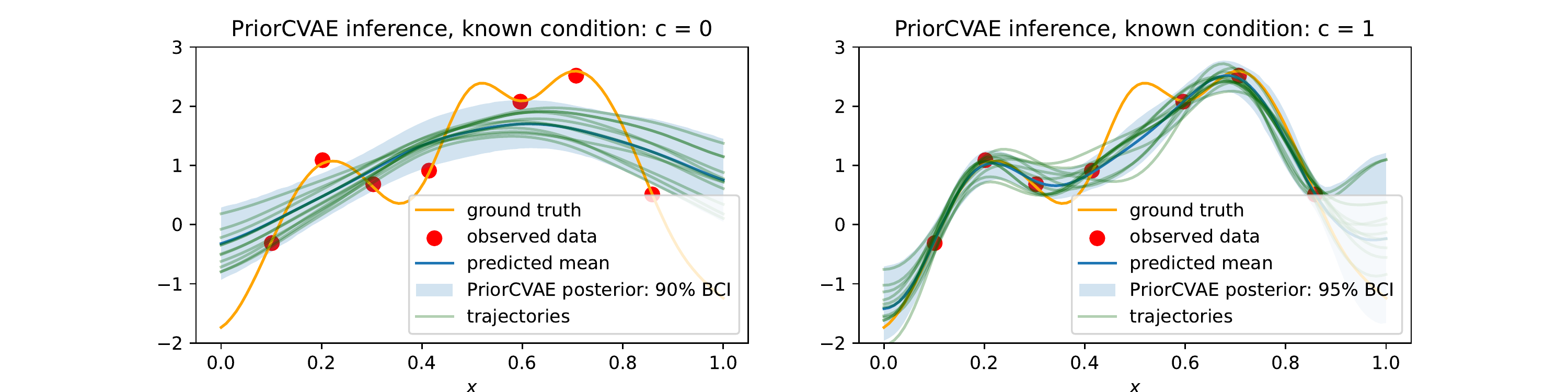}
\end{center}    
\caption{Inference performed using NUTS and PriorCVAE, assuming the binary condition $c$ is known. Left: $c=0$ (misspecified model), right: $c=1$ (correctly specified model).}
\label{fig:binary_inference_true_false}
\end{figure}

\textbf{MCMC Inference.} To perform inference, we generate one GP realisation according to the true value of the condition $c=1$ and use it as the ground truth $f_\text{true}$. Seven observed data points are sampled by adding i.i.d.~noise to the ground truth: $f_\text{obs} \sim \mathcal{N}(f_\text{true}, s)$, where the standard deviation of the noise is modelled with a half-Normal prior $s \sim \mathcal{N}^+(0.1)$. Using this observed data we aim to reconstruct the underlying true curve according to three different scenarios: (a) assuming an incorrect condition is known, (b) assuming a known and true condition, and (c) inferring the condition alongside reconstruction of the true trajectory. For all scenarios, we ran 4 MCMC chains with $n_\text{warmup}=1000$ burn-in steps, and $n_\text{samples}=100000$ post-warmup steps. Results of scenarios (a) and (b) are presented in \cref{fig:binary_inference_true_false}; inference in both cases was performed using a NUTS sampler~\citep{hoffman14a} and took 8 seconds and 30 seconds, correspondingly. Scenario (c) needs a separate treatment as direct inference of discrete parameters, such as $c \in \{0,1\}$, is impossible with NUTS. Instead, we use two alternative approaches. The first approach is to approximate the discrete distribution of the condition with a Beta-prior $c \sim \text{Beta}(10^{-4}, 10^{-4})$ which is concentrated around values 0 and 1 and assigns low probability away from 0 and 1. This model is inferred using NUTS (see \cref{fig:binary_inference_Beta_Bern}, left) and took 24 seconds to run. The second approach is to assign $c$ its original discrete prior $c \sim \text{Bernoulli}(0.5)$ and use the MixedHMC sampler~\citep{zhou2020mixed} (see \cref{fig:binary_inference_Beta_Bern}, right); it took 9 minutes to run. The results are very close to those when the true condition value is known (\cref{fig:binary_inference_Beta_Bern}, left), and the  posterior samples of $c$ obtained using the two approaches are shown on the bottom of \cref{fig:binary_inference_Beta_Bern}. 
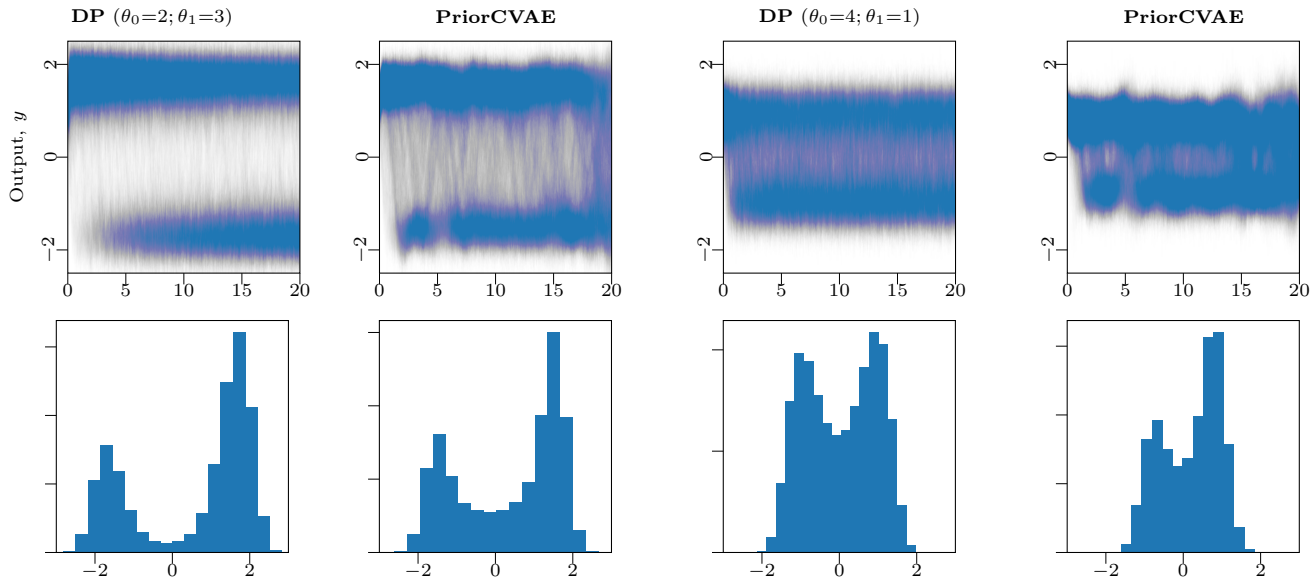
\begin{figure*}[t!]
	\centering\scriptsize
	\pgfplotsset{axis on top,scale only axis,width=\figurewidth,height=\figureheight, ylabel near ticks,ylabel style={yshift=6pt},y tick label style={rotate=90}} 
	\setlength{\figurewidth}{.18\textwidth}
	\setlength{\figureheight}{\figurewidth}
	\newcommand{\mytitle}[1]{\tikz\node[minimum width=\figurewidth,minimum height=2em,align=center]{#1};}
	\begin{subfigure}{.2\textwidth}  
		\raggedleft
		\mytitle{\textbf{DP} ($\theta_0{=}2; \theta_1{=}3$)}    
		\pgfplotsset{ylabel={Output, $y$}}
		\tikz[outer sep=0pt, inner sep=0pt]\node{
\begin{tikzpicture}

\begin{axis}[
height=\figureheight,
tick align=outside,
tick pos=left,
width=\figurewidth,
x grid style={white!69.0196078431373!black},
xmin=0, xmax=20,
xtick style={color=black},
y grid style={white!69.0196078431373!black},
ymin=-2.5, ymax=2.5,
ytick style={color=black}
]
\addplot graphics [includegraphics cmd=\pgfimage,xmin=-1.60283548291818, xmax=21.0810278625578, ymin=-2.95419001694253, ymax=2.67594161344976] {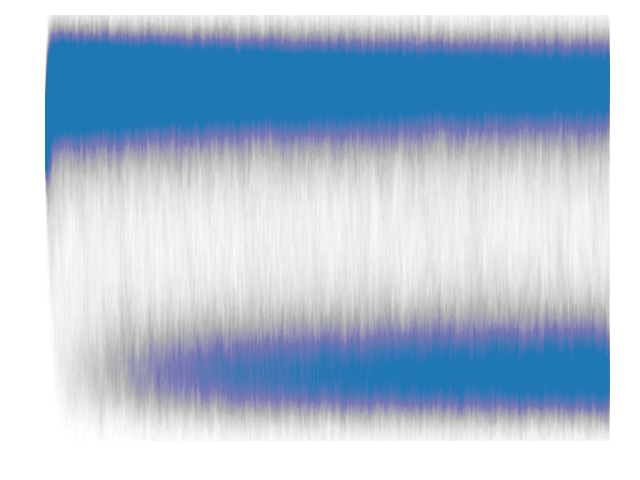};
\end{axis}

\end{tikzpicture}};
	\end{subfigure}
	\hfill
	\begin{subfigure}{.2\textwidth}
		\raggedleft
		\mytitle{\textbf{PriorCVAE}}
		\tikz[outer sep=0pt, inner sep=0pt]\node{
\begin{tikzpicture}

\begin{axis}[
height=\figureheight,
tick align=outside,
tick pos=left,
width=\figurewidth,
x grid style={white!69.0196078431373!black},
xmin=0, xmax=20,
xtick style={color=black},
y grid style={white!69.0196078431373!black},
ymin=-2.5, ymax=2.5,
ytick style={color=black}
]
\addplot graphics [includegraphics cmd=\pgfimage,xmin=-1.60283548291818, xmax=21.0810278625578, ymin=-2.95419001694253, ymax=2.67594161344976] {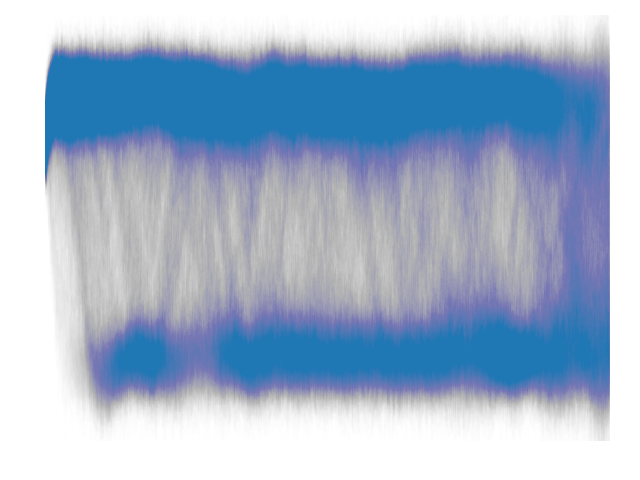};
\end{axis}

\end{tikzpicture}};
	\end{subfigure}
	\hfill
	\begin{subfigure}{.2\textwidth}
		\raggedleft
		\mytitle{\textbf{DP} ($\theta_0{=}4; \theta_1{=}1$)}    
		\tikz[outer sep=0pt, inner sep=0pt]\node{
\begin{tikzpicture}

\begin{axis}[
height=\figureheight,
tick align=outside,
tick pos=left,
width=\figurewidth,
x grid style={white!69.0196078431373!black},
xmin=0, xmax=20,
xtick style={color=black},
y grid style={white!69.0196078431373!black},
ymin=-2.5, ymax=2.5,
ytick style={color=black}
]
\addplot graphics [includegraphics cmd=\pgfimage,xmin=-1.60283548291818, xmax=21.0810278625578, ymin=-2.95419001694253, ymax=2.67594161344976] {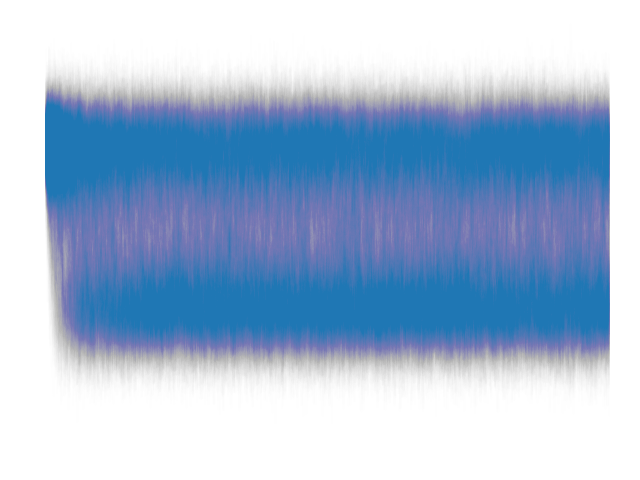};
\end{axis}

\end{tikzpicture}};
	\end{subfigure}
	\hfill
	\begin{subfigure}{.2\textwidth}
		\raggedleft
		\mytitle{\textbf{PriorCVAE}}
		\tikz[outer sep=0pt, inner sep=0pt]\node{
\begin{tikzpicture}

\begin{axis}[
height=\figureheight,
tick align=outside,
tick pos=left,
width=\figurewidth,
x grid style={white!69.0196078431373!black},
xmin=0, xmax=20,
xtick style={color=black},
y grid style={white!69.0196078431373!black},
ymin=-2.5, ymax=2.5,
ytick style={color=black}
]
\addplot graphics [includegraphics cmd=\pgfimage,xmin=-1.60283548291818, xmax=21.0810278625578, ymin=-2.95419001694253, ymax=2.67594161344976] {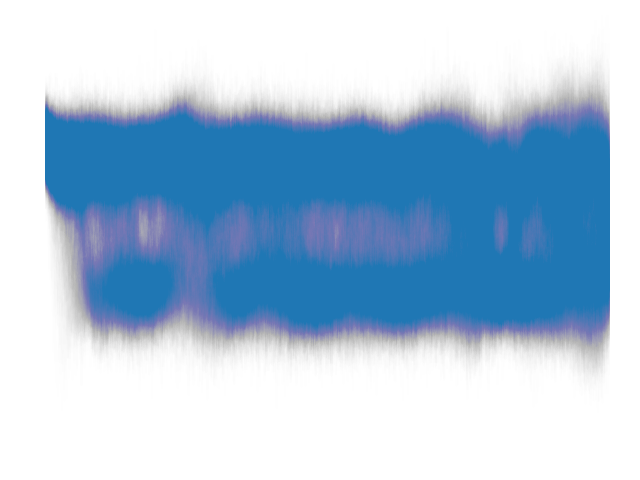};
\end{axis}

\end{tikzpicture}};
	\end{subfigure}
	\\[1em]
	\begin{subfigure}{.2\textwidth}  
		\raggedleft
		\pgfplotsset{ylabel={\phantom{Output, $y$}}}
		\tikz[outer sep=0pt, inner sep=0pt]\node{
\begin{tikzpicture}

\definecolor{color0}{rgb}{0.12156862745098,0.466666666666667,0.705882352941177}

\begin{axis}[
height=\figureheight,
tick align=outside,
tick pos=left,
width=\figurewidth,
x grid style={white!69.0196078431373!black},
xmin=-3, xmax=3,
xtick style={color=black},
y grid style={white!69.0196078431373!black},
ymin=0, ymax=0.677001944215776,
ytick style={color=black},
ytick={},
yticklabels={}
]
\draw[draw=none,fill=color0] (axis cs:-3.11875198382567,0) rectangle (axis cs:-2.8062769136912,6.46451571042352e-05);
\draw[draw=none,fill=color0] (axis cs:-2.8062769136912,0) rectangle (axis cs:-2.49380184355672,0.00404448265090755);
\draw[draw=none,fill=color0] (axis cs:-2.49380184355672,0) rectangle (axis cs:-2.18132677342225,0.0546363586466364);
\draw[draw=none,fill=color0] (axis cs:-2.18132677342225,0) rectangle (axis cs:-1.86885170328778,0.211208529280745);
\draw[draw=none,fill=color0] (axis cs:-1.86885170328778,0) rectangle (axis cs:-1.5563766331533,0.313036812690064);
\draw[draw=none,fill=color0] (axis cs:-1.5563766331533,0) rectangle (axis cs:-1.24390156301883,0.237101474905233);
\draw[draw=none,fill=color0] (axis cs:-1.24390156301883,0) rectangle (axis cs:-0.931426492884353,0.122605300907726);
\draw[draw=none,fill=color0] (axis cs:-0.931426492884353,0) rectangle (axis cs:-0.618951422749879,0.0582728087465147);
\draw[draw=none,fill=color0] (axis cs:-0.618951422749879,0) rectangle (axis cs:-0.306476352615405,0.0336007605198122);
\draw[draw=none,fill=color0] (axis cs:-0.306476352615405,0) rectangle (axis cs:0.005998717519069,0.0273788241612947);
\draw[draw=none,fill=color0] (axis cs:0.005998717519069,0) rectangle (axis cs:0.318473787653543,0.0326765267885398);
\draw[draw=none,fill=color0] (axis cs:0.318473787653543,0) rectangle (axis cs:0.630948857788018,0.0544171411584364);
\draw[draw=none,fill=color0] (axis cs:0.630948857788018,0) rectangle (axis cs:0.943423927922492,0.114886925169904);
\draw[draw=none,fill=color0] (axis cs:0.943423927922492,0) rectangle (axis cs:1.25589899805697,0.258639833140032);
\draw[draw=none,fill=color0] (axis cs:1.25589899805697,0) rectangle (axis cs:1.56837406819144,0.496646340244724);
\draw[draw=none,fill=color0] (axis cs:1.56837406819144,0) rectangle (axis cs:1.88084913832591,0.644763756395977);
\draw[draw=none,fill=color0] (axis cs:1.88084913832591,0) rectangle (axis cs:2.19332420846039,0.423543228402331);
\draw[draw=none,fill=color0] (axis cs:2.19332420846039,0) rectangle (axis cs:2.50579927859486,0.10513286703436);
\draw[draw=none,fill=color0] (axis cs:2.50579927859486,0) rectangle (axis cs:2.81827434872934,0.00749403784113802);
\draw[draw=none,fill=color0] (axis cs:2.81827434872934,0) rectangle (axis cs:3.13074941886381,0.000104648348381608);
\end{axis}

\end{tikzpicture}};
	\end{subfigure}
	\hfill
	\begin{subfigure}{.2\textwidth}
		\raggedleft
		\tikz[outer sep=0pt, inner sep=0pt]\node{
\begin{tikzpicture}

\definecolor{color0}{rgb}{0.12156862745098,0.466666666666667,0.705882352941177}

\begin{axis}[
height=\figureheight,
tick align=outside,
tick pos=left,
width=\figurewidth,
x grid style={white!69.0196078431373!black},
xmin=-3, xmax=3,
xtick style={color=black},
y grid style={white!69.0196078431373!black},
ymin=0, ymax=0.632527691869715,
ytick style={color=black},
ytick={},
yticklabels={}
]
\draw[draw=none,fill=color0] (axis cs:-3.26669394098378,0) rectangle (axis cs:-2.93723264201044,3.33878365509947e-06);
\draw[draw=none,fill=color0] (axis cs:-2.93723264201044,0) rectangle (axis cs:-2.60777134303711,0.000115643324781172);
\draw[draw=none,fill=color0] (axis cs:-2.60777134303711,0) rectangle (axis cs:-2.27831004406378,0.00342953786536081);
\draw[draw=none,fill=color0] (axis cs:-2.27831004406378,0) rectangle (axis cs:-1.94884874509045,0.0506699878013632);
\draw[draw=none,fill=color0] (axis cs:-1.94884874509045,0) rectangle (axis cs:-1.61938744611712,0.230904207070943);
\draw[draw=none,fill=color0] (axis cs:-1.61938744611712,0) rectangle (axis cs:-1.28992614714378,0.323621925647268);
\draw[draw=none,fill=color0] (axis cs:-1.28992614714378,0) rectangle (axis cs:-0.960464848170452,0.207784647888313);
\draw[draw=none,fill=color0] (axis cs:-0.960464848170452,0) rectangle (axis cs:-0.63100354919712,0.135289638385138);
\draw[draw=none,fill=color0] (axis cs:-0.63100354919712,0) rectangle (axis cs:-0.301542250223787,0.115849418790428);
\draw[draw=none,fill=color0] (axis cs:-0.301542250223787,0) rectangle (axis cs:0.0279190487495447,0.111414603519096);
\draw[draw=none,fill=color0] (axis cs:0.0279190487495447,0) rectangle (axis cs:0.357380347722877,0.117157614931653);
\draw[draw=none,fill=color0] (axis cs:0.357380347722877,0) rectangle (axis cs:0.686841646696209,0.136522863657017);
\draw[draw=none,fill=color0] (axis cs:0.686841646696209,0) rectangle (axis cs:1.01630294566954,0.192775601255494);
\draw[draw=none,fill=color0] (axis cs:1.01630294566954,0) rectangle (axis cs:1.34576424464287,0.374300715696449);
\draw[draw=none,fill=color0] (axis cs:1.34576424464287,0) rectangle (axis cs:1.67522554361621,0.602407325590204);
\draw[draw=none,fill=color0] (axis cs:1.67522554361621,0) rectangle (axis cs:2.00468684258954,0.369375402753604);
\draw[draw=none,fill=color0] (axis cs:2.00468684258954,0) rectangle (axis cs:2.33414814156287,0.0607880199052488);
\draw[draw=none,fill=color0] (axis cs:2.33414814156287,0) rectangle (axis cs:2.6636094405362,0.00272080515311923);
\draw[draw=none,fill=color0] (axis cs:2.6636094405362,0) rectangle (axis cs:2.99307073950953,0.000114125695847036);
\draw[draw=none,fill=color0] (axis cs:2.99307073950953,0) rectangle (axis cs:3.32253203848287,1.24445572599162e-05);
\end{axis}

\end{tikzpicture}};
	\end{subfigure}
	\hfill
	\begin{subfigure}{.2\textwidth}
		\raggedleft
		\tikz[outer sep=0pt, inner sep=0pt]\node{
\begin{tikzpicture}

\definecolor{color0}{rgb}{0.12156862745098,0.466666666666667,0.705882352941177}

\begin{axis}[
height=\figureheight,
tick align=outside,
tick pos=left,
width=\figurewidth,
x grid style={white!69.0196078431373!black},
xmin=-3, xmax=3,
xtick style={color=black},
y grid style={white!69.0196078431373!black},
ymin=0, ymax=0.457896299006015,
ytick style={color=black},
ytick={},
yticklabels={}
]
\draw[draw=none,fill=color0] (axis cs:-2.35696120140005,0) rectangle (axis cs:-2.11660533864486,0.000111085287015423);
\draw[draw=none,fill=color0] (axis cs:-2.11660533864486,0) rectangle (axis cs:-1.87624947588966,0.00299181385366256);
\draw[draw=none,fill=color0] (axis cs:-1.87624947588966,0) rectangle (axis cs:-1.63589361313447,0.0311438211416722);
\draw[draw=none,fill=color0] (axis cs:-1.63589361313447,0) rectangle (axis cs:-1.39553775037928,0.13646349052616);
\draw[draw=none,fill=color0] (axis cs:-1.39553775037928,0) rectangle (axis cs:-1.15518188762409,0.299797971116657);
\draw[draw=none,fill=color0] (axis cs:-1.15518188762409,0) rectangle (axis cs:-0.914826024868898,0.394663142028772);
\draw[draw=none,fill=color0] (axis cs:-0.914826024868898,0) rectangle (axis cs:-0.674470162113707,0.377838505618222);
\draw[draw=none,fill=color0] (axis cs:-0.674470162113707,0) rectangle (axis cs:-0.434114299358515,0.311006351761584);
\draw[draw=none,fill=color0] (axis cs:-0.434114299358515,0) rectangle (axis cs:-0.193758436603324,0.255459131706467);
\draw[draw=none,fill=color0] (axis cs:-0.193758436603324,0) rectangle (axis cs:0.0465974261518678,0.231855796489392);
\draw[draw=none,fill=color0] (axis cs:0.0465974261518678,0) rectangle (axis cs:0.286953288907059,0.242319032006811);
\draw[draw=none,fill=color0] (axis cs:0.286953288907059,0) rectangle (axis cs:0.527309151662251,0.28879345485614);
\draw[draw=none,fill=color0] (axis cs:0.527309151662251,0) rectangle (axis cs:0.767665014417442,0.365452704140885);
\draw[draw=none,fill=color0] (axis cs:0.767665014417442,0) rectangle (axis cs:1.00802087717263,0.436091713339062);
\draw[draw=none,fill=color0] (axis cs:1.00802087717263,0) rectangle (axis cs:1.24837673992783,0.412149713614008);
\draw[draw=none,fill=color0] (axis cs:1.24837673992783,0) rectangle (axis cs:1.48873260268302,0.263596649042552);
\draw[draw=none,fill=color0] (axis cs:1.48873260268302,0) rectangle (axis cs:1.72908846543821,0.0941761925027606);
\draw[draw=none,fill=color0] (axis cs:1.72908846543821,0) rectangle (axis cs:1.9694443281934,0.0155506920328669);
\draw[draw=none,fill=color0] (axis cs:1.9694443281934,0) rectangle (axis cs:2.20980019094859,0.00102015402157984);
\draw[draw=none,fill=color0] (axis cs:2.20980019094859,0) rectangle (axis cs:2.45015605370378,1.62259408000056e-05);
\end{axis}

\end{tikzpicture}};
	\end{subfigure}
	\hfill
	\begin{subfigure}{.2\textwidth}
		\raggedleft
		\tikz[outer sep=0pt, inner sep=0pt]\node{
\begin{tikzpicture}

\definecolor{color0}{rgb}{0.12156862745098,0.466666666666667,0.705882352941177}

\begin{axis}[
height=\figureheight,
tick align=outside,
tick pos=left,
width=\figurewidth,
x grid style={white!69.0196078431373!black},
xmin=-3, xmax=3,
xtick style={color=black},
y grid style={white!69.0196078431373!black},
ymin=0, ymax=0.674575710139545,
ytick style={color=black},
ytick={},
yticklabels={}
]
\draw[draw=none,fill=color0] (axis cs:-2.39400942439209,0) rectangle (axis cs:-2.12929140636108,5.28864642616058e-06);
\draw[draw=none,fill=color0] (axis cs:-2.12929140636108,0) rectangle (axis cs:-1.86457338833008,0.000153748506817668);
\draw[draw=none,fill=color0] (axis cs:-1.86457338833008,0) rectangle (axis cs:-1.59985537029907,0.00255063861924544);
\draw[draw=none,fill=color0] (axis cs:-1.59985537029907,0) rectangle (axis cs:-1.33513735226806,0.0259963415077923);
\draw[draw=none,fill=color0] (axis cs:-1.33513735226806,0) rectangle (axis cs:-1.07041933423705,0.137357858261619);
\draw[draw=none,fill=color0] (axis cs:-1.07041933423705,0) rectangle (axis cs:-0.805701316206045,0.33064201919851);
\draw[draw=none,fill=color0] (axis cs:-0.805701316206045,0) rectangle (axis cs:-0.540983298175038,0.386077988797984);
\draw[draw=none,fill=color0] (axis cs:-0.540983298175038,0) rectangle (axis cs:-0.27626528014403,0.300185837711628);
\draw[draw=none,fill=color0] (axis cs:-0.27626528014403,0) rectangle (axis cs:-0.0115472621130221,0.252041400491994);
\draw[draw=none,fill=color0] (axis cs:-0.0115472621130221,0) rectangle (axis cs:0.253170755917986,0.274412374874652);
\draw[draw=none,fill=color0] (axis cs:0.253170755917986,0) rectangle (axis cs:0.517888773948993,0.395928092766709);
\draw[draw=none,fill=color0] (axis cs:0.517888773948993,0) rectangle (axis cs:0.782606791980001,0.625411905209291);
\draw[draw=none,fill=color0] (axis cs:0.782606791980001,0) rectangle (axis cs:1.04732481001101,0.642453057275757);
\draw[draw=none,fill=color0] (axis cs:1.04732481001101,0) rectangle (axis cs:1.31204282804202,0.315983402347029);
\draw[draw=none,fill=color0] (axis cs:1.31204282804202,0) rectangle (axis cs:1.57676084607302,0.0748490795880735);
\draw[draw=none,fill=color0] (axis cs:1.57676084607302,0) rectangle (axis cs:1.84147886410403,0.0114884510794568);
\draw[draw=none,fill=color0] (axis cs:1.84147886410403,0) rectangle (axis cs:2.10619688213504,0.00173883139282979);
\draw[draw=none,fill=color0] (axis cs:2.10619688213504,0) rectangle (axis cs:2.37091490016605,0.000274254093242328);
\draw[draw=none,fill=color0] (axis cs:2.37091490016605,0) rectangle (axis cs:2.63563291819706,4.72200573764337e-05);
\draw[draw=none,fill=color0] (axis cs:2.63563291819706,0) rectangle (axis cs:2.90035093622806,6.79968826220646e-06);
\end{axis}

\end{tikzpicture}};
	\end{subfigure}	
	\caption{\textbf{Encoding diffusion process prior:} (Top) Samples of the Double-Well diffusion process of the true diffusion process and the trained PriorCVAE model for different values of $\theta$. (Bottom) The histogram of the samples showing the two wells of the process.}
	\label{fig:exp_dp_dw}
\end{figure*}
\section{Encoding Integral of the Intensity Function of the Log-Gaussian Cox Process} %
\label{sec:lgcp_details}
Similarly to other prior-encoding VAEs, PriorCVAE is able to encode function properties alongside function realisations. The integral
\begin{equation}
	\mathcal{I} = \int_D \exp(f(s)) \mathrm{d} s, s\in D
\end{equation}
 is a crucial quantity required to evaluate log-Gaussian Cox process (LGCP) likelihood~\citep{moller1998log}. %
\begin{figure*}[t!]
	\begin{center}
		\hspace*{-1cm}
		\includegraphics[width=1.1\textwidth]{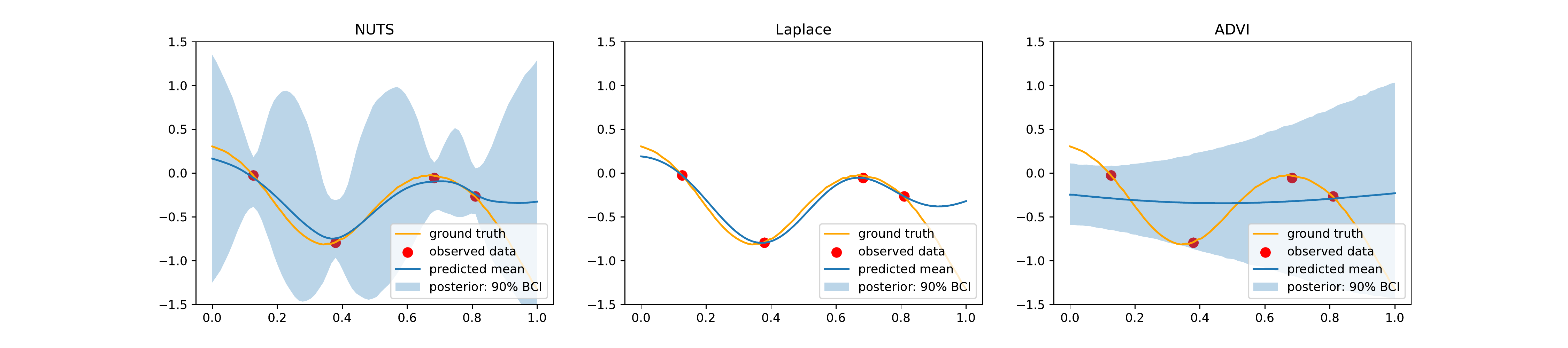}\\
		\hspace*{-1cm}
		\includegraphics[width=1.1\textwidth]{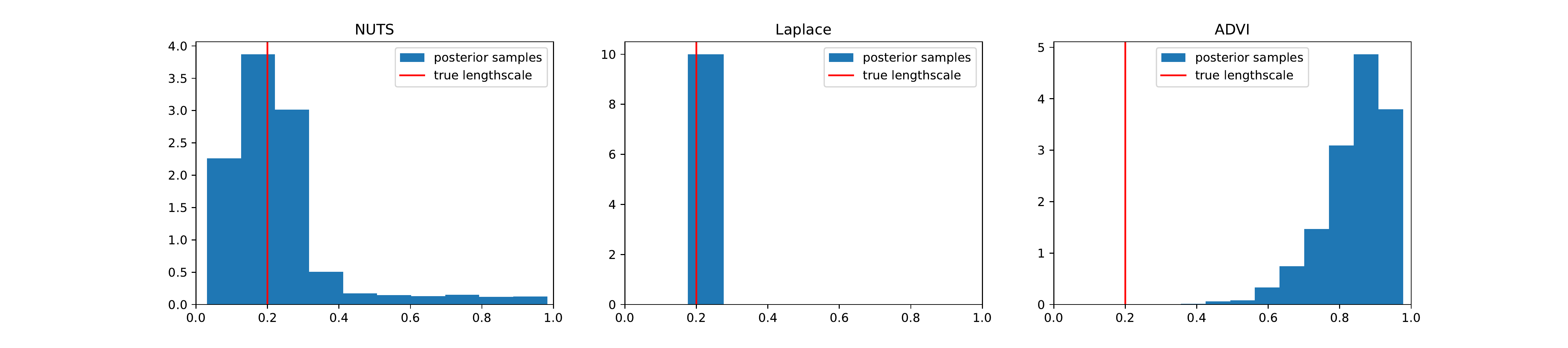}
	\end{center}
	\caption{Comparing NUTS, Laplace and ADVI inference results on a one-dimensional Gaussian Process. Top: inferred mean and 90\% BCI, bottom: inferred lengthscale. Note, that results of the Laplace approximation and ADVI differ between hardware. See \cref{fig:laplace_colab} and \cref{fig:advi_colab} for results obtained on Google Colab.}
	\label{fig:NUTS_Laplace_ADVI_gp}
\end{figure*}

LGCP is defined by an intensity function 
\begin{equation}
  \lambda(s) = \exp(f(s)), \quad f(s) \sim \mathcal{GP}(\cdot,\cdot)  
\end{equation}
which is Gaussian on the log-scale. 
The log-likelihood of an observed point pattern $S$, \ie ~a collection of observed event locations $S=\{s_1, s_2, \ldots, s_N\}, s_j \in D$, can then be written as 
$L(\lambda | s_1, s_2, \ldots, s_N) =\exp\left( -\int_D \lambda(s) ds \right) \prod_{j=1}^N \lambda(s_j).$
Therefore, the log-likelihood involves the integral of the intensity function over the whole observation domain $\mathcal{I} := \int_D \lambda(s)\mathrm{d}s = \int_D \exp(f(s))\mathrm{d}s.$ The presence of this term makes LGCP inference challenging. Here we demonstrate in one-dimensional case $D=(0,1)$ that the integral $\mathcal{I}$ can be encoded jointly with the GP realisations and recovered at inference. For computational stability, we used the following scheme
\begin{align}
    \log(I) &\approx -\log n +\log\left( \sum_{i=1}^n \exp(f(x_i)) \right) , \nonumber \\
            &= -\log n + \text{log-sum-exp}(f(x_i)).
\end{align}
Here $\text{log-sum-exp}(v)=\log(\sum_{i=1}^n \exp(v_i))=\max(v) + \log \left[\sum_{i=1}^n \exp(v_i-\max(v)) \right]$.
The objective now includes an additional term - reconstruction loss of the integral value:
\begin{multline}
\mathcal{L}_\text{PriorCVAE} =
    \frac{1}{2 \sigma^2_{f,\text{VAE}}}\text{MSE}(f_\text{GP}, f_\text{PriorCVAE}) + \frac{1}{2 \sigma^2_{\mathcal{I},\text{VAE}}}\text{MSE}(\mathcal{I}_\text{GP}, \mathcal{I}_\text{PriorCVAE}) \\- \KL{\mathcal{N}(\mu_z, \sigma^2_z I_d)}{\mathcal{N}(0, I_d)}, 
\end{multline}    
 where $\mathcal{I}_\text{GP} \approx \frac{1}{n} \sum_{i=1}^n f_\text{GP}^i, \quad f_\text{GP}^i = f(g_i), \quad \{g_i\}_{i=1}^n$ is the computational grid, 
$\mathcal{I}_\text{PriorCVAE} \approx \frac{1}{n} \sum_{i=1}^n f_\text{PriorCVAE}^i$, and $\sigma_{f,\text{VAE}}$ and $\sigma_{\mathcal{I},\text{VAE}}$ are two different tuning parameters.

\begin{figure}[h!]
\begin{center}
   \includegraphics[width=0.5\columnwidth]{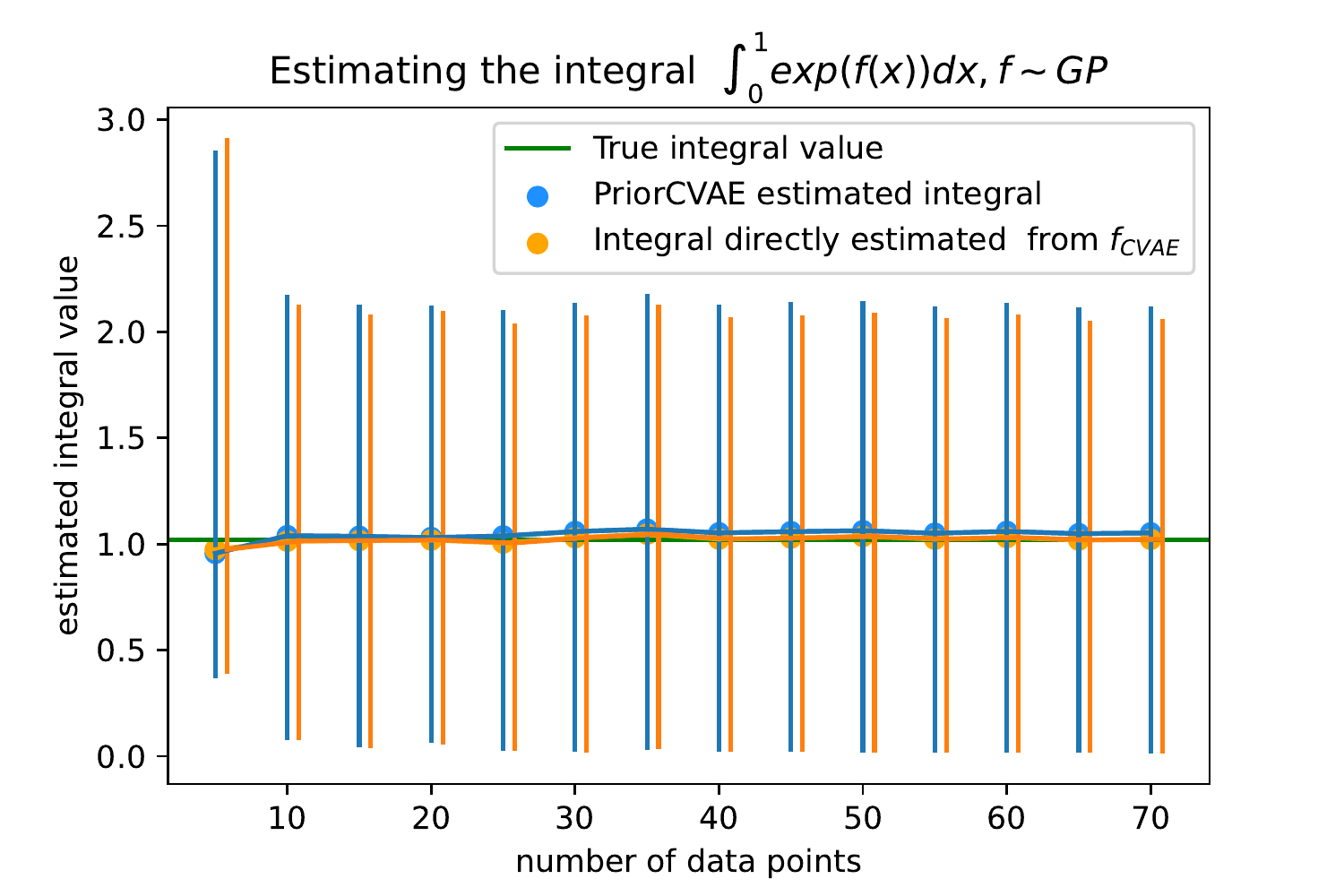}
   \vspace{-0.2cm}
\end{center}
\caption{Encoding $\int_0^1\exp(f(x))dx$. The graph shows the mean estimate and uncertainty over five runs of the two methods to compute the integral: the first method uses encoded integral values directly, and the second uses $f_\text{PriorCVAE}$ draws and then computes the integral.}
\label{fig:int_exp_f}
\end{figure}

In this experiment we have focused on GP samples with RBF kernel and $\ell=0.2$. To assess the quality of the integral estimation by the decoder, we compare $\mathcal{I}_\text{PriorCVAE}$ with the value of the integral computed from the recovered $f_\text{PriorCVAE}$ samples. The benefit of such encoding is that if a more complex quadrature is required, it would be performed prior to inference and not during it, even if the evaluations $f_\text{GP}$ are no longer costly. We fix one curve for which we know the true value of the integral computed via the same quadrature rule. Then we iterate through the number of observed points from 5 to 70 in increments of 5 points, and fit the PriorCVAE model to the data five times. Results are presented on \cref{fig:int_exp_f}. For any assessed number of observed points, the mean estimates produced by the two approaches differ by at most $0.1$. The same statement holds for uncertainty intervals. %
Here, we used the simplest quadrature scheme, however, other quadratures could be used, without adding extra computational cost to the inference step.
\begin{figure}[t!]
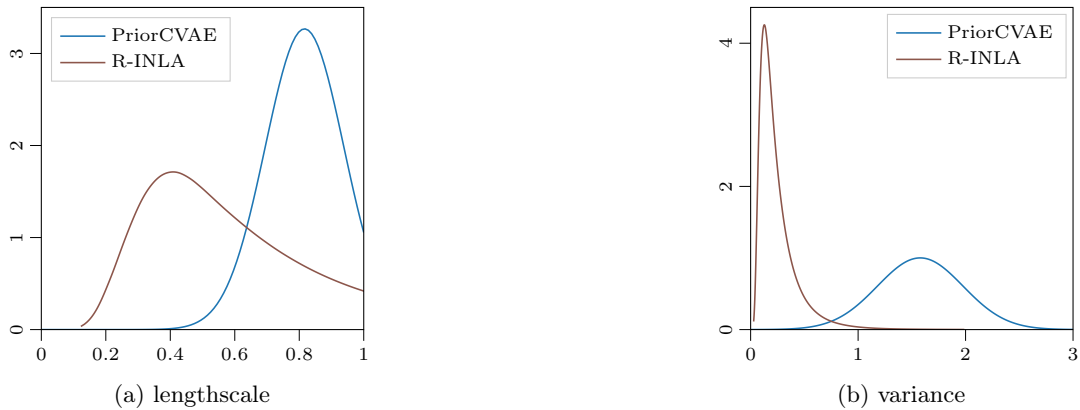

	\scriptsize
	\pgfplotsset{axis on top,scale only axis,width=\figurewidth,height=\figureheight, ylabel near ticks,ylabel style={yshift=6pt},y tick label style={rotate=90}}  
	\setlength{\figurewidth}{.25\textwidth}
	\setlength{\figureheight}{\figurewidth}
	\begin{subfigure}{.45\textwidth}
		\centering
		\input{figures/zimbabwe/lengthscale_inla.tex}
		\caption{lengthscale}
	\end{subfigure}
	\hfill
	\begin{subfigure}{.45\textwidth}
		\centering
		\input{figures/zimbabwe/variance_inla.tex}
		\caption{variance}
	\end{subfigure}
	\caption{\textbf{Zimbabwe experiment:} Parameter estimates produced by R-INLA and PriorCVAE.}
	\label{fig:inla_zimbabwe_comparison}
\end{figure}
\section{Encoding Non-stationary Kernels: Experiment Details}\label{sec:lin_rbf_details}
\textbf{Neural network training details}. CVAE architecture is identical to the two experimenst above. The neural network was trained for $n_\text{epochs}=5000$ epochs, with $n_\text{batch}=2000$ batch size, and $1e^{-3}$ learning rate. Value of the $\sigma^2_\text{vae}$ hyperparameter was set to $0.01$. 

\section{HIV Prevalence in Zimbabwe Estimation}
\label{sec:hiv_details}

\textbf{Inference model}
\begin{align}
\begin{split}
    y_i &\sim \text{Binomial}(n_i, \theta_i), \\
    \text{logit}^{-1}(\theta_i) &= f_i,\\
    f &\sim \mathcal{MVN}(0, \mathbf{K}),\\
    \kappa_{ij} &= \text{Mat\'ern}^{1/2}(x_i, x_j),\\
    \ell \sim &\text{Uniform}(0,1), \sigma \sim \text{Gamma}(1.5, 1.5).
\end{split}
\label{eq:Zimbabwe}
\end{align}

\textbf{Data pre-processing}
In order to train the priors, we transformed the coordiantes into the ENU system and normalized them. ENU coordinates, short for East-North-Up coordinates, are a type of Cartesian coordinate system commonly used in geodesy, navigation, and earth sciences to represent positions on or near the Earth's surface. The ENU coordinate system is a local, tangential coordinate system that is centered at a specific reference point on the Earth's surface.

\textbf{Neural network training details}. Data for neural network training were drawn from the uniform prior on the lengthscale. The neural network was trained for $n_\text{epochs}=10000$ epochs, with $n_\text{batch}=2000$ batch size, and $1e^{-3}$ learning rate. Value of the $\sigma^2_\text{vae}$ hyperparameter was set to 0.8. 
\begin{figure*}[t!]
	\centering\scriptsize
	\pgfplotsset{axis on top,scale only axis,width=\figurewidth,height=\figureheight, ylabel near ticks,ylabel style={yshift=6pt},y tick label style={rotate=90},yticklabels={},xticklabels={}} 
	\setlength{\figurewidth}{.14\textwidth}
	\setlength{\figureheight}{\figurewidth}
	\newcommand{\mytitle}[1]{\tikz\node[minimum width=\figurewidth,minimum height=2em,align=center]{#1};}
	\begin{subfigure}{.25\textwidth}  
		\raggedleft
		\mytitle{\textbf{GP} ($\ell{=}0.1$)}    
		\pgfplotsset{ylabel={Output, $y$}}
		\tikz[outer sep=0pt, inner sep=0pt]\node{\input{figures/SquaredExponential/GP_samps_0_1.tex}};
	\end{subfigure}
	\hfill
	\begin{subfigure}{.24\textwidth}
		\raggedleft
		\mytitle{\textbf{PriorCVAE}}
		\tikz[outer sep=0pt, inner sep=0pt]\node{\input{figures/SquaredExponential/PriorCVAE_samps_0_1.tex}};
	\end{subfigure}
	\hfill
	\begin{subfigure}{.24\textwidth}
		\raggedleft
		\mytitle{\textbf{GP} ($\ell{=}1.0$)}
		\tikz[outer sep=0pt, inner sep=0pt]\node{\input{figures/SquaredExponential/GP_samps_1_0.tex}};
	\end{subfigure}
	\hfill
	\begin{subfigure}{.24\textwidth}
		\raggedleft
		\mytitle{\textbf{PriorCVAE}}
		\tikz[outer sep=0pt, inner sep=0pt]\node{\input{figures/SquaredExponential/PriorCVAE_samps_1_0.tex}};
	\end{subfigure}\\[1em]
	\newcommand{\addxy}[1]{\tikz[node distance=5pt,inner sep=0]{\node(a){\includegraphics[width=\figurewidth]{figures/SquaredExponential/#1}};\node[left=of a,rotate=90,anchor=center]{Input, $x'$};\node[below=of a,anchor=center]{Input, $x$};\node[draw,minimum width=\figurewidth,minimum height=\figurewidth] at (a) {};}}
	\newcommand{\addx}[1]{\tikz[node distance=5pt,inner sep=0]{\node(a){\includegraphics[width=\figurewidth]{figures/SquaredExponential/#1}};\node[below=of a,anchor=center]{Input, $x$};\node[draw,minimum width=\figurewidth,minimum height=\figurewidth] at (a) {};}}
	\begin{subfigure}{.25\textwidth}  
		\raggedleft  
		\addxy{GP_cov_0_1.png}
	\end{subfigure}
	\hfill
	\begin{subfigure}{.24\textwidth}
		\raggedleft
		\addx{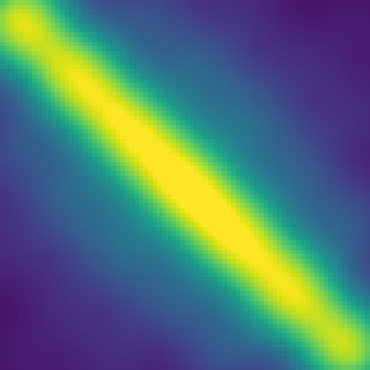}
	\end{subfigure}
	\hfill
	\begin{subfigure}{.24\textwidth}
		\raggedleft
		\addx{GP_cov_1_0.png}
	\end{subfigure}
	\hfill
	\begin{subfigure}{.24\textwidth}
		\raggedleft
		\addx{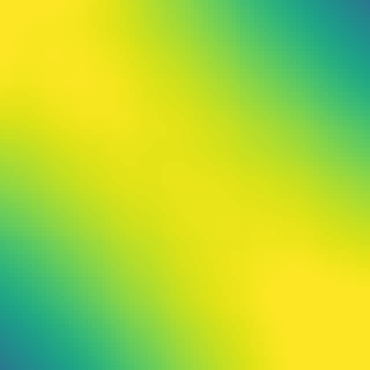}
	\end{subfigure}
	\newcommand{\cbar}{\protect\includegraphics[width=2.5em,height=.7em]{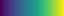}}
	\caption{\textbf{Our PriorCVAE samples and covariance resemble the original GP.} Example draws (top) and computed Gram (covariance) matrices (bottom, 0~\cbar~1) from a Squared Exponential (RBF) 
		prior with lengthscale $\ell=0.1$ (two left-hand columns) and $\ell=1.0$ (two right-hand columns).}
	\label{fig:exp_rbf_encode_prior}
\end{figure*}

\textbf{Comparison with R-INLA}. R-INLA \citep{lindgren2015bayesian} is a popular software for Bayesian spatial modelling and is viewed as state-of-the-art in many applied fields. While being very fast, R-INLA has limitations, such as, for example, it does not permit RBF kernel, or Mat\'ern kernels with smoothness higher than $\nu = \alpha - m/2$, where $\alpha \in (0,2]$ and $m$ is space dimensionality. In our case, $m=2$, hence, the highest available smoothness is $\nu=1.$ In order to compare hyperparameter estimates, we need to align R-INLA parameterisation with Mat\'ern kernel parametrisation that we use. Denote the distance between a pair of points as $ \|s \| = \| c_i-c_j \|$. Then our parametrisation is
\begin{equation}
	C_\nu(\|s \|) = \sigma^2 \frac{2^{1-\nu}}{\Gamma(\nu)} \left(\sqrt{2 \nu}\frac{\|s\|}{l} \right)^\nu K_\nu\left(\sqrt{2 \nu}\frac{\|s\|}{l}\right),
\end{equation}
and R-INLA parametrisation is
\begin{align}
	&C_\nu(\|s\|) = \sigma^2 \frac{2^{1-\nu}}{\Gamma(\nu)} \left(\kappa \|s\| \right)^\nu K_\nu\left(\kappa \|s\|\right), \\
	& \sigma^2 = \frac{\Gamma(\nu)}{\Gamma(\alpha) (4\pi)^{m/2} \kappa^{2\nu} \tau^2}.
\end{align}
Priors in R-INLA are provided with respect to parameters
\begin{align}
    \theta_1 &= \log(\tau), \\
    \theta_2 &= \log(\kappa).
\end{align}

Hence, lengthscale can be recovered from R-INLA parametrisation as 
\begin{equation}
	 l = \frac{\sqrt{2 \nu}}{\kappa} = \frac{\sqrt{2 \nu}}{\exp(\theta_2)},
\end{equation}
and for $m=2, \nu=1, \alpha=2$ we get (since $\Gamma(1) =\Gamma(2) = 1$)
\begin{equation}
	\sigma^2 = \frac{1}{4 \pi k^2 \tau^2} = \frac{1}{4 \pi \exp^2(\theta_2) \exp^2(\theta_1)}.
\end{equation}

While we attempted a head-to-head comparison, R-INLA's interface for setting priors was not flexible enough to easily allow this. Hence we conducted a sensitivity analysis, which showed that the R-INLA posteriors were not sensitive to the prior choice. Comparison of parameter estimates produced by PriorCVAE and R-ILNA are shown on \cref{fig:inla_zimbabwe_comparison}. However, they should be takem with caution since, as explained, they use the same kernel (Mat\'ern-$\nicefrac{5}{2}$), the priors used by the two models were different.

\section{SIR: Training Details}
\label{sec:sir_details}
In this experiment, the values of $ (\beta, \gamma)$ act as a conditional vector for the PriorCVAE model, which is defined with the hidden dimensions as $10$, latent dimension as $6$ and leaky-ReLU activation function. We train the PriorCVAE model foo $10000$ iterations with $2000$ batch size and Adam optimizer with $10^{-3}$ learning rate. We put a uniform prior on both the $\beta$ and the $\gamma$ parameter, $\mathcal{U}(0, 1)$. The data for training was scaled by $N$, so that the values of the learnt curve $I(t)$ were scaled to the $(0,1)$ range: $I(t)/N.$
\section{Double-Well}
\label{app:dw_experiment}
We experiment to encode the non-linear Double-Well diffusion process (DP) using the PriorCVAE model. The Double-Well DP is defined by an SDE
\begin{equation}
		\mathrm{d} x_t = \theta_1 x_t (\theta_2 - x_t^2) + \mathrm{d} \beta_t \, 
\end{equation}
with $\theta_1, \theta_2$ being the parameters of the DP, and $Q$ is the spectral density. 

We simulate trajectories from the DP with two sets of parameters $(\theta_0{=}2; \theta_1{=}3)$, and $(\theta_0{=}4; \theta_1{=}1)$ using Euler--Maruayama for $T=[0,  20]$ with $\Delta t=0.01$. The values of $\theta = (\theta_0, \theta_1)$ act as a conditional vector for the PriorCVAE model, which is defined with the hidden dimensions as $[1000, 500, 100]$, latent dimension as $50$ and sigmoid activation function. We train the PriorCVAE model for $5000$ iterations with $2000$ batch size and Adam optimizer with $10^{-3}$ learning rate. 

\cref{fig:exp_dp_dw} shows the learnt samples from PriorCVAE along with the samples from the true Double-Well DP. The plot also shows the histogram of states which clearly shows two wells depending on the $\theta$ values. 

\section{Software, and Reproducibility}
\label{sec:software_hardware}
We provide a sample code as a proof-of-concept with this supplementary.
In terms of the implementation, the neural network models were implemented with JAX~\citep{jax2018github} in Python. Bayesian inference, including NUTS, Laplace approximation and ADVI, were implemented in NumPyro~\citep{phan2019composable, bingham2019pyro} using JAX as the backend. 

\section{Probabilistic Programming Example with NumPyro}
\label{sec:numpyro_example}
\begin{lstlisting}[language=Python, caption=Probabilistic programming example]
# PriorCVAE decoder as a function of neural network parameters expressed 
# in terms of `jax.numpy` (jnp) tensors
def c_decoder_numpy(z, W1, B1, W2, B2, c):
   
    def linear(z, W, B):
        lin_out = jnp.matmul(z, W) + B
        return lin_out
        c = jnp.array(c).reshape(1)
        z = jnp.concatenate([z, c], axis=0)

    hidden = jax.nn.relu(linear(z, W1, B1))
    out = linear(hidden, W2, B2)
    return out

# Numpyro model for inference
def numpyro_model(z_dim, conditional=False,  y=None, obs_idx=None):    
    
    # priors
    c = numpyro.sample("c", npdist.Uniform(0.01,0.99))
    z = numpyro.sample("z", npdist.Normal(jnp.zeros(z_dim), jnp.ones(z_dim))) 
    sigma = numpyro.sample("sigma", npdist.HalfNormal(0.1))
   
    # deterministic transformation
    f = numpyro.deterministic("f", c_decoder_numpy(z, W1, B1, W2, B2, c))

    # likelihood
    y = numpyro.sample("y", npdist.Normal(f[obs_idx], sigma), obs=y)

# Setup and run MCMC
kernel = NUTS(numpyro_model)
mcmc = MCMC(kernel,
        num_warmup=args["num_warmup"],   # number of warmup steps
        num_samples=args["num_samples"], # number of posterior steps
        num_chains=args["num_chains"]    # number of chains
  )
\end{lstlisting}

\clearpage
\section{Supplementary Figures.}
\label{sec:suppl_figures}
This section contains supplementary figures of the experiments.
\begin{figure}[h!]
\begin{center}
    \hspace*{-1.5cm} 
    \includegraphics[width=0.3\textwidth]{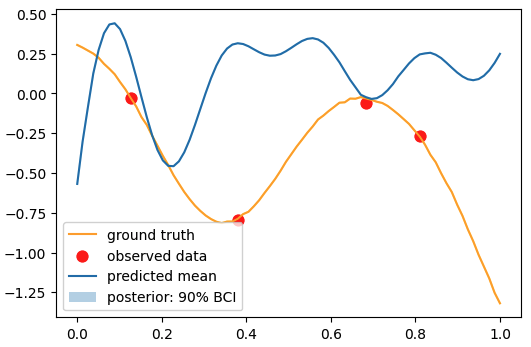}
    \includegraphics[width=0.26\textwidth]{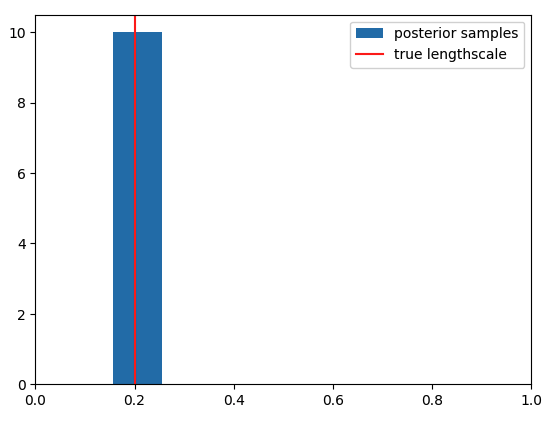}
\end{center}
\caption{Laplace approximation inference results produced by Google Colab: left - mean and uncertainty, right - hyperparameter estimate.}
\label{fig:laplace_colab}
\end{figure}

\begin{figure}[h!]
\begin{center}
    \hspace*{-1.5cm}
    \includegraphics[width=0.3\textwidth]{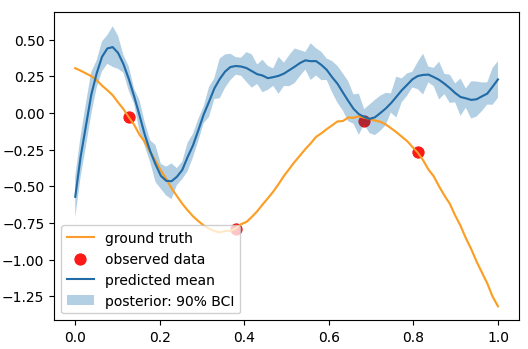}
    \includegraphics[width=0.26\textwidth]{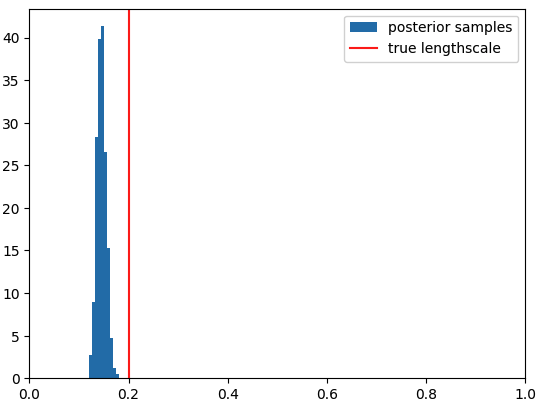}
\end{center}
\caption{ADVI inference results produced by Google Colab: left - mean and uncertainty,  right - hyperparameter estimate.}
\label{fig:advi_colab}
\end{figure}

\begin{figure*}[h!]
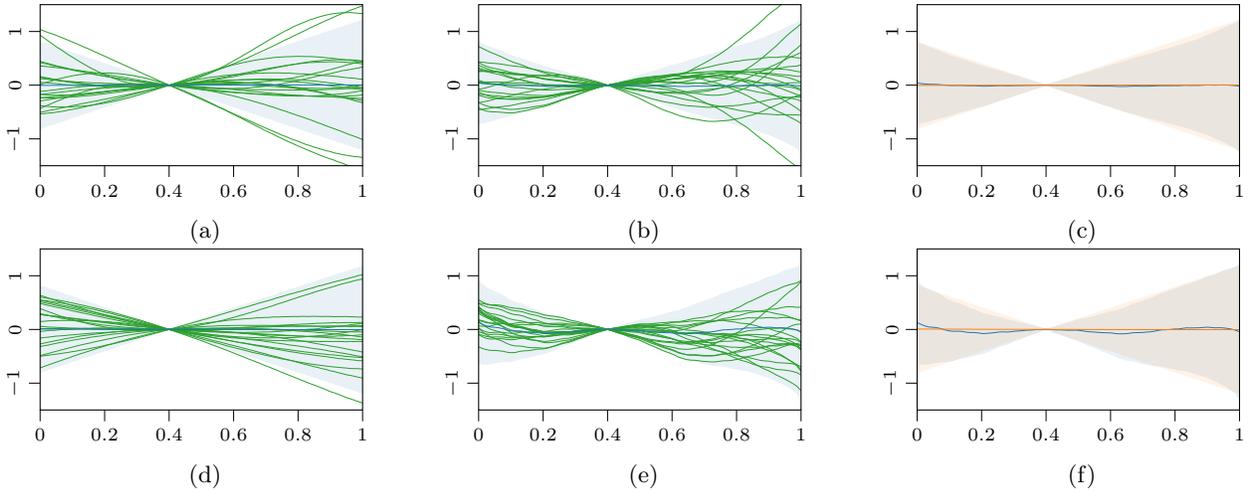

	\centering
	\scriptsize
	\pgfplotsset{axis on top,scale only axis,width=\figurewidth,height=\figureheight, ylabel near ticks,ylabel style={yshift=6pt},y tick label style={rotate=90}}  
	\setlength{\figurewidth}{.25\textwidth}
	\setlength{\figureheight}{.5\figurewidth}
	\begin{subfigure}{.32\textwidth}
		\input{figures/non_stationary/GP_samples_0_5.tex}
		\caption{}
	\end{subfigure}
	\hfill
	\begin{subfigure}{.32\textwidth}
		\input{figures/non_stationary/VAE_samples_0_5.tex}
		\caption{}
	\end{subfigure}
	\hfill
	\begin{subfigure}{.32\textwidth}
		\input{figures/non_stationary/GP_VAE_comparison_0_5.tex}
		\caption{}
	\end{subfigure}
	\begin{subfigure}{.32\textwidth}
		\input{figures/non_stationary/GP_samples_0_9.tex}
		\caption{}
	\end{subfigure}
	\hfill
	\begin{subfigure}{.32\textwidth}
		\input{figures/non_stationary/VAE_samples_0_9.tex}
		\caption{}
	\end{subfigure}
	\hfill
	\begin{subfigure}{.32\textwidth}
		\input{figures/non_stationary/GP_VAE_comparison_0_9.tex}
		\caption{}
	\end{subfigure}
	\caption{\textbf{Non-stationary kernel:} (a)~Priors drawn from a GP kernel ($\ell=0.5$) . (b)~ Learnt priors from the PriorCVAE model ($\ell=0.5$) . (c)~Comparison of the second moments of the GP and PriorCVAE model ($\ell=0.5$). (d)~Priors drawn from a GP kernel ($\ell=0.9$). (e)~ Learnt priors from the PriorCVAE model ($\ell=0.9$). (f)~Comparison of the second moments of the GP and PriorCVAE model ($\ell=0.9$).}
	\label{fig:lin_rbf_priors_05_09}
\end{figure*}

\begin{figure*}[h!]
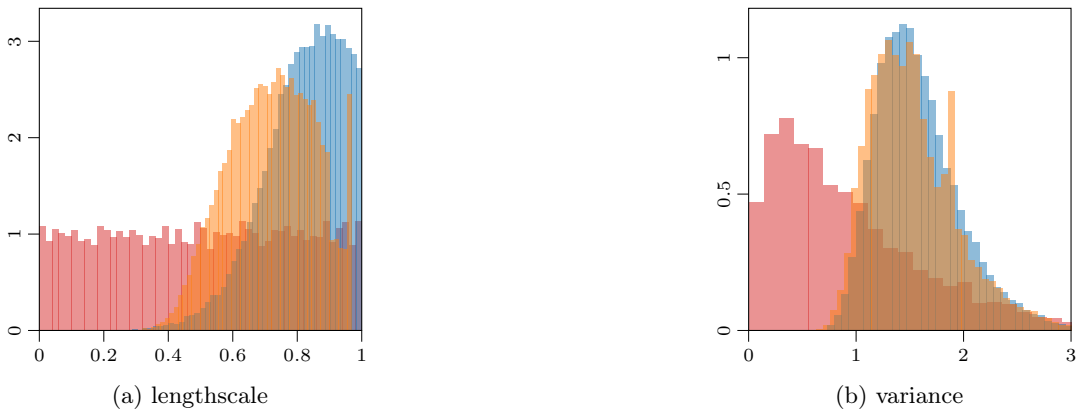

	\scriptsize
	\pgfplotsset{axis on top,scale only axis,width=\figurewidth,height=\figureheight, ylabel near ticks,ylabel style={yshift=6pt},y tick label style={rotate=90}}  
	\setlength{\figurewidth}{.25\textwidth}
	\setlength{\figureheight}{\figurewidth}
    \begin{subfigure}{.45\textwidth}
    	\centering
    	\input{figures/zimbabwe/lengthscale.tex}
    	\caption{lengthscale}
    \end{subfigure}
	\hfill
	\begin{subfigure}{.45\textwidth}
		\centering
		\input{figures/zimbabwe/variance.tex}
		\caption{variance}
	\end{subfigure}
	\caption{Parameter estimates produced by models with $f_\text{GP}$ and $f_\text{PriorCVAE}$ priors.}
	\label{fig:zimbabwe_estimates_params}
\end{figure*}

\begin{figure*}
	\centering
	\scriptsize
	\setlength{\figurewidth}{.5\textwidth}
	\setlength{\figureheight}{\figurewidth}
\begin{tikzpicture}

\definecolor{color0}{rgb}{0.12156862745098,0.466666666666667,0.705882352941177}

\begin{axis}[
height=\figureheight,
tick align=outside,
tick pos=left,
width=\figurewidth,
x grid style={white!69.0196078431373!black},
xlabel={GP},
xmin=0.04, xmax=0.26,
xtick style={color=black},
y grid style={white!69.0196078431373!black},
ylabel={PriorCVAE},
ymin=0.04, ymax=0.26,
ytick style={color=black},
ytick={0.05, 0.10, 0.15, 0.20, 0.25},
yticklabels={0.05, 0.10, 0.15, 0.20, 0.25},
xtick={0.05, 0.10, 0.15, 0.20, 0.25},
xticklabels={0.05, 0.10, 0.15, 0.20, 0.25}
]
\addplot [semithick, color0, mark=*, mark size=3, mark options={solid}, only marks]
table {%
0.179678992858686 0.182559068160779
0.139783522777545 0.140265356300353
0.137782920903114 0.142141964826237
0.11062400939496 0.110347317001041
0.0937027356183916 0.105842689955641
0.115456115892026 0.1212642248858
0.110989759554902 0.105941541565732
0.0970238851004918 0.0953646165995506
0.0938542294743727 0.102112574441211
0.0977946742812534 0.106814841409204
0.128165017143148 0.125924944405803
0.121043039562936 0.122903821932506
0.120610155216735 0.113783359989278
0.131184316540137 0.128671146037079
0.111043119730729 0.114522461828314
0.117348981638575 0.120132353738248
0.106893955738376 0.118531568729788
0.121106082526616 0.117458888071139
0.139495625199147 0.135344831697113
0.13301601632722 0.131400966129564
0.126448731440503 0.120314966480275
0.132774275995938 0.127362057042591
0.101942430014063 0.111263750614891
0.122113306666884 0.111691916859752
0.109364090277652 0.104394593938722
0.143565122608891 0.145963904081573
0.112642587262428 0.110280204196605
0.14230869275787 0.147150000635406
0.0837688876325482 0.0908233014587182
0.0708422653121077 0.0844481791024801
0.110333631960779 0.104129456339801
0.145575233335022 0.15161228680256
0.122878710544985 0.124398889083763
0.13045961391254 0.128026478949
0.11237638921883 0.110262871790147
0.175130430926341 0.164713811850793
0.144302354106202 0.141651117802828
0.125555325566581 0.129926764705928
0.133026098236076 0.13296235016888
0.177580611779602 0.182873085123171
0.126536686818321 0.123711447573056
0.111637044321312 0.118405438473609
0.176175997261278 0.175648696148026
0.191426062077718 0.214518638689871
0.194943330823634 0.166427024380458
0.160583781514219 0.147703177882322
0.236596703982183 0.221242140490122
0.194622509418388 0.196476533346798
0.212047281721684 0.212791406878283
0.223664223789921 0.223152920346066
0.186945623993025 0.178503794569737
0.16390961099944 0.16246452562175
0.200335234639061 0.205021973899196
0.186910131332205 0.190936073756423
0.171028321082504 0.172247368006432
0.142887565190263 0.145011720042415
0.0904800482157024 0.0878877389415541
0.121975069331071 0.11060437165014
0.151015305200412 0.155766554463949
0.145228630920131 0.153355185714106
0.163114560736306 0.157808207774554
0.145274713945405 0.144165061196459
0.151726715378393 0.148580511091279
};
\addplot [semithick, white!30!black, dashed]
table {%
0.05 0.05
0.25 0.25
};
\end{axis}

\end{tikzpicture}
	\caption{Comparison of estimated prevalence of HIV in Zimbabwe obtained using the original $f_\text{GP}$ prior, and the model with the $f_\text{PriorCVAE}$ prior.}
	\label{fig:zimbabwe_estimates_prev}
\end{figure*}
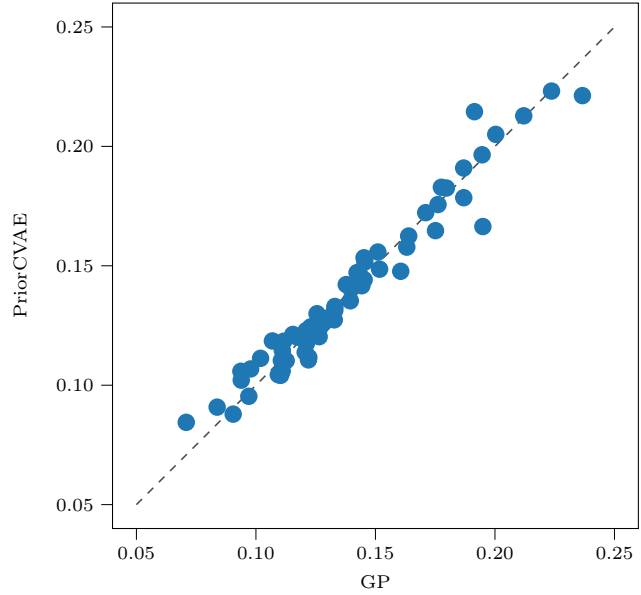

\begin{figure*}[h!]
	\centering
	\pgfplotsset{axis on top,scale only axis,width=\figurewidth,height=\figureheight, ylabel near ticks,y tick label style={rotate=90}, tick label style={font=\scriptsize}, ytick={0, 0.4, 0.8}, xtick={0, 6, 12}}
	 \begin{subfigure}{.24\textwidth}
		\setlength{\figurewidth}{.8\textwidth}
		\setlength{\figureheight}{\figurewidth}
\begin{tikzpicture}

\definecolor{color0}{rgb}{0.12156862745098,0.466666666666667,0.705882352941177}
\definecolor{color1}{rgb}{1,0.498039215686275,0.0549019607843137}

\begin{axis}[
height=\figureheight,
tick pos=left,
width=\figurewidth,
xmin=0, xmax=13,
ymin=0, ymax=1
]
\addplot [semithick, color0]
table {%
0 0.0015558996968434
1 0.00470708420362768
2 0.0165023395516048
3 0.0324712769602155
4 0.0950631788206707
5 0.236328655394256
6 0.364680269420971
7 0.410980034033055
8 0.378979917443193
9 0.311461538790132
10 0.245971056764577
11 0.184176090712745
12 0.136660708091767
13 0.0988027046792852
};
\addplot [semithick, color0]
table {%
0 0.001556423091087
1 0.00467833843970219
2 0.0164353270446765
3 0.0461045851893362
4 0.154213651116052
5 0.315047131287984
6 0.406099182285806
7 0.4070298361163
8 0.35114700309518
9 0.277636598050334
10 0.214240014225656
11 0.160388229215378
12 0.120190947195139
13 0.0885332678924173
};
\addplot [semithick, color0]
table {%
0 0.00120891219954625
1 0.00470361319888544
2 0.019478636083698
3 0.0772007821344945
4 0.259159589849108
5 0.485221177421801
6 0.6197253902618
7 0.677471816095445
8 0.687830766299052
9 0.665191159709212
10 0.610188211032455
11 0.542339148207099
12 0.472621563090802
13 0.403345667337824
};
\addplot [semithick, color0]
table {%
0 0.00141056294426299
1 0.00472070174416552
2 0.0175969248541977
3 0.0286244680042202
4 0.075009205116264
5 0.220149087201352
6 0.405562758816927
7 0.524418407247799
8 0.546608318955995
9 0.506962114663927
10 0.438450246814475
11 0.355889525731437
12 0.27906258059673
13 0.211938218966385
};
\addplot [semithick, color0]
table {%
0 0.00142059099535421
1 0.00483628374143791
2 0.0183026521043397
3 0.0970161880488117
4 0.357561861790127
5 0.537960177507674
6 0.575495679041689
7 0.537503453740294
8 0.470419078638428
9 0.39682647417943
10 0.322958421936188
11 0.26147448354991
12 0.214579221826462
13 0.172380569458039
};
\addplot [semithick, color0]
table {%
0 0.00132495436268001
1 0.00466343266886287
2 0.0181446038895084
3 0.0869401276112384
4 0.313365584792387
5 0.510331413592774
6 0.579220185037801
7 0.572469960640934
8 0.52969360385817
9 0.469199199314978
10 0.397595469185177
11 0.32870089056459
12 0.270150067052243
13 0.217887969988717
};
\addplot [semithick, color0]
table {%
0 0.00155275982728864
1 0.00465534175797905
2 0.0163326483847307
3 0.041457520244439
4 0.13426743824097
5 0.288362886076216
6 0.387416030172663
7 0.396161037494546
8 0.343376907981218
9 0.270122890683831
10 0.207890790834635
11 0.15445418300318
12 0.114525137443689
13 0.083304273913687
};
\addplot [semithick, color0]
table {%
0 0.00144845275156906
1 0.00467557250247662
2 0.0171904045488664
3 0.0577534624819932
4 0.199237311412075
5 0.381412671700528
6 0.474739945678142
7 0.480228962978482
8 0.432716176371512
9 0.363096379648485
10 0.293227413928303
11 0.22925307662546
12 0.178442527969484
13 0.13615189594956
};
\addplot [semithick, color0]
table {%
0 0.00134497661285191
1 0.00468984267901153
2 0.0181591007745176
3 0.0572171474726989
4 0.188203783082305
5 0.38477170420668
6 0.516196592146457
7 0.561960944350893
8 0.549213914009485
9 0.499264871564387
10 0.429293862062432
11 0.354411103242558
12 0.286873353089751
13 0.227070014776166
};
\addplot [semithick, color0]
table {%
0 0.00157527614550643
1 0.00468283897808664
2 0.0162023231173716
3 0.0233816631971879
4 0.0606563848085925
5 0.175224647390641
6 0.315648855272895
7 0.389737122588769
8 0.373331068594465
9 0.309028409782402
10 0.243442153774898
11 0.180196013819398
12 0.130598552529798
13 0.0922044466811708
};
\addplot [semithick, color0]
table {%
0 0.0015561156009557
1 0.00483275346015198
2 0.0172220094728909
3 0.0706763349018655
4 0.259476381020244
5 0.433631503282285
6 0.482011609185618
7 0.442785306310122
8 0.365642185236826
9 0.286745768865592
10 0.22125540331545
11 0.16935816818039
12 0.132651893456267
13 0.101519969897508
};
\addplot [semithick, color0]
table {%
0 0.00137841773688783
1 0.00474315985796409
2 0.0178991780590547
3 0.0222463747334311
4 0.0512901271150707
5 0.177488282233576
6 0.38873911835617
7 0.552522423124324
8 0.601735264265988
9 0.577139202443284
10 0.514661934844167
11 0.427868817902395
12 0.341099806922615
13 0.261276357119688
};
\addplot [semithick, color0]
table {%
0 0.00223343813359845
1 0.00567600469240507
2 0.0170155788602478
3 0.0575700791690895
4 0.212942526517151
5 0.326084475595403
6 0.307294387217021
7 0.218700246494653
8 0.128316853948487
9 0.0751535468607569
10 0.0480418488091801
11 0.0334708064655934
12 0.0261758054869705
13 0.0193413343837443
};
\addplot [semithick, color0]
table {%
0 0.00166014782482716
1 0.00470063741224623
2 0.0158615267455026
3 0.0469728228900607
4 0.163623278501743
5 0.313447680756886
6 0.373072915235042
7 0.339659570853355
8 0.262690700082572
9 0.188426271610198
10 0.137537837434393
11 0.0991006264131778
12 0.0729408450215439
13 0.0526462231169285
};
\addplot [semithick, color0]
table {%
0 0.00213197809077986
1 0.00620870147681174
2 0.0199034740169124
3 0.113897637205847
4 0.416064402018492
5 0.506454691184265
6 0.427989639765709
7 0.299992410625464
8 0.185619578574508
9 0.118796713469563
10 0.0782216117030749
11 0.0582536703549301
12 0.0498077131787036
13 0.0398975324646423
};
\addplot [semithick, color0]
table {%
0 0.00147887954390603
1 0.00468584558702209
2 0.0169366957823984
3 0.0335308002030787
4 0.097126965325537
5 0.246627625125162
6 0.391593516947501
7 0.456311794227285
8 0.439391764095104
9 0.377862784189442
10 0.308274074615318
11 0.237919314095692
12 0.180017103107972
13 0.132586809558918
};
\addplot [semithick, color0]
table {%
0 0.00164413798155619
1 0.00547237556152594
2 0.0197771934473701
3 0.135026405996048
4 0.494039931564303
5 0.633077621194843
6 0.61058699140628
7 0.517044313965069
8 0.407062013681164
9 0.317839931143526
10 0.24401382439559
11 0.195851729911662
12 0.169009454044368
13 0.13941148892789
};
\addplot [semithick, color0]
table {%
0 0.00151994142403783
1 0.00466554566003392
2 0.0166132271253523
3 0.0527355808939546
4 0.182218480515483
5 0.351797416804527
6 0.433898429770176
7 0.425620830048076
8 0.36543770513228
9 0.290744894935116
10 0.226099325496812
11 0.171031853073478
12 0.129724428882289
13 0.0966141219221873
};
\addplot [semithick, color0]
table {%
0 0.00192128020326548
1 0.00494963524905845
2 0.0154904486281379
3 0.0405843746331992
4 0.141614864857799
5 0.269860061429241
6 0.308923338311176
7 0.258007708276626
8 0.17553736715242
9 0.111884722557065
10 0.0760495441679437
11 0.0526945460496253
12 0.0385479863932791
13 0.0273443874288877
};
\addplot [semithick, color0]
table {%
0 0.00116464040717806
1 0.00469111824014955
2 0.0198525693207469
3 0.088564532152573
4 0.300235779046561
5 0.530305098833054
6 0.653658571465073
7 0.704934495526345
8 0.715155392361626
9 0.696656102845979
10 0.646572632586404
11 0.58323243182685
12 0.517133358226144
13 0.449220872567787
};
\addplot [semithick, color1]
table {%
0 0.00131061598951507
1 0.00461482175243029
2 0.0160222547759104
3 0.0530519746651178
4 0.152262417531184
5 0.316170795865019
6 0.423542277890164
7 0.407759051808234
8 0.332186014543812
9 0.251156824945641
10 0.1834162039436
11 0.131611853990954
12 0.0935303873838828
13 0.0660908420791797
};
\end{axis}

\end{tikzpicture}\\[-1em]
		\caption{$\beta{=}1.64, \gamma{=}0.38$}
	\end{subfigure}
	\hfill
	\begin{subfigure}{.24\textwidth}
			\setlength{\figurewidth}{.8\textwidth}
			\setlength{\figureheight}{\figurewidth}
\begin{tikzpicture}

\definecolor{color0}{rgb}{0.12156862745098,0.466666666666667,0.705882352941177}
\definecolor{color1}{rgb}{1,0.498039215686275,0.0549019607843137}

\begin{axis}[
height=\figureheight,
tick pos=left,
width=\figurewidth,
xmin=0, xmax=13,
ymin=0, ymax=1
]
\addplot [semithick, color0]
table {%
0 0.00154496590758788
1 0.00595218671054159
2 0.0251360947511553
3 0.0872974122530111
4 0.274388384485932
5 0.463780839331661
6 0.525411525923126
7 0.459495707460301
8 0.354241980526454
9 0.256245849258705
10 0.192481943919269
11 0.142524820769018
12 0.111173903471841
13 0.0803699813342398
};
\addplot [semithick, color0]
table {%
0 0.00155031158993128
1 0.00602566249776665
2 0.0254335331737066
3 0.0742688410660027
4 0.223060773642203
5 0.419086612358109
6 0.513295615486923
7 0.479359681630044
8 0.391196038468454
9 0.296828886278325
10 0.229010684661083
11 0.171173659311978
12 0.133997746982168
13 0.0968763403535691
};
\addplot [semithick, color0]
table {%
0 0.00156278473443071
1 0.00598631373789501
2 0.0251884573922203
3 0.0941644913918884
4 0.274424899648971
5 0.409553715489618
6 0.420613188528698
7 0.336678800229049
8 0.238571870231411
9 0.162151924476775
10 0.115377508276881
11 0.0822818308037783
12 0.0620797156975661
13 0.0440605533146553
};
\addplot [semithick, color0]
table {%
0 0.00151092609498786
1 0.0060714685971143
2 0.0261033778173858
3 0.0580398384461763
4 0.15855235698564
5 0.353496136092717
6 0.499723543644175
7 0.518246669353047
8 0.463643262807019
9 0.379316160719474
10 0.306040336378902
11 0.234781732037676
12 0.185021353540588
13 0.134505687841294
};
\addplot [semithick, color0]
table {%
0 0.00129426621883657
1 0.0060313398043502
2 0.0289654644923164
3 0.203388973345707
4 0.559903937603148
5 0.674500937161124
6 0.654645480438229
7 0.560273793661695
8 0.463250504070896
9 0.379741543012648
10 0.307543986893152
11 0.247266005074444
12 0.207251521804322
13 0.164494061271353
};
\addplot [semithick, color0]
table {%
0 0.00181687125752376
1 0.00605033437842151
2 0.0230902426199322
3 0.0514621727342445
4 0.148244656921166
5 0.297448074746156
6 0.366805196340366
7 0.309639778738464
8 0.21140907143523
9 0.133342613887164
10 0.092389244462535
11 0.0630669355985221
12 0.0463786361293306
13 0.0313830476423778
};
\addplot [semithick, color0]
table {%
0 0.00157089063675838
1 0.00600393343193515
2 0.0253112160170069
3 0.0975568819525387
4 0.262797034281458
5 0.352851489742273
6 0.32862858636265
7 0.242587422056723
8 0.161286987705183
9 0.105163109200404
10 0.0720440691637282
11 0.0499210983358927
12 0.036540034527243
13 0.0255376041112082
};
\addplot [semithick, color0]
table {%
0 0.00179158809650759
1 0.00603025206946787
2 0.0231724777949802
3 0.0438845999137289
4 0.120341264987133
5 0.26767977332309
6 0.362489613598894
7 0.328343849574914
8 0.23755919325172
9 0.155121658423331
10 0.109267776988187
11 0.0749736447799779
12 0.0548269288178251
13 0.0368457384161031
};
\addplot [semithick, color0]
table {%
0 0.00147894559938043
1 0.00604787012077172
2 0.0266733101317912
3 0.126403862680236
4 0.366713000325233
5 0.498025771039517
6 0.493675715911503
7 0.400045592343155
8 0.298617325870468
9 0.216844874920556
10 0.160653249384882
11 0.11871157184984
12 0.0925613443968708
13 0.0681454839098982
};
\addplot [semithick, color0]
table {%
0 0.0018287940189818
1 0.00601929530234662
2 0.0229585635077842
3 0.0485549671282038
4 0.134682929623774
5 0.268084011049692
6 0.330648812720112
7 0.27406449572162
8 0.183019805260485
9 0.112444823195352
10 0.0765624939459933
11 0.0515377801968014
12 0.0372030703767296
13 0.0248739119529986
};
\addplot [semithick, color0]
table {%
0 0.00155223961034575
1 0.00594705322822041
2 0.0251828830562424
3 0.102379675624475
4 0.250118090089805
5 0.283950176923185
6 0.222161951872553
7 0.14606890399511
8 0.0902362096895997
9 0.0562795799975371
10 0.0366854123059178
11 0.0247255497685888
12 0.0173730715662663
13 0.0119553763256049
};
\addplot [semithick, color0]
table {%
0 0.00153704397153317
1 0.00599277547572683
2 0.0255129887312383
3 0.094307665660822
4 0.296954181637543
5 0.487540674270926
6 0.544630509750279
7 0.475840000063538
8 0.369632517838397
9 0.271191599218321
10 0.206006363346795
11 0.153728108654177
12 0.121178955620239
13 0.0884396987173049
};
\addplot [semithick, color0]
table {%
0 0.001503813292061
1 0.00600411075494051
2 0.0258900700168309
3 0.0936247098659647
4 0.291692979300662
5 0.489379737436996
6 0.558972328703862
7 0.505687456291138
8 0.409949834820548
9 0.313293905926248
10 0.243481453515229
11 0.184990007407001
12 0.147319382483718
13 0.108764441794209
};
\addplot [semithick, color0]
table {%
0 0.00142657248035022
1 0.00605880224471437
2 0.0270749701509789
3 0.0855843076075788
4 0.253650006175661
5 0.469550566034988
6 0.583499536768578
7 0.57720580648765
8 0.517094247237017
9 0.435043661287281
10 0.35999045391359
11 0.286595215156752
12 0.234508881532478
13 0.178112004085128
};
\addplot [semithick, color0]
table {%
0 0.00138839213401906
1 0.00597172482486434
2 0.0271067099682192
3 0.134527734383228
4 0.420072276395858
5 0.607193887073774
6 0.64678433731676
7 0.580247960077847
8 0.484641181573545
9 0.389568760149002
10 0.314771221969106
11 0.25003690608588
12 0.207121098173294
13 0.159911777238239
};
\addplot [semithick, color0]
table {%
0 0.00149242280397682
1 0.00599012041734302
2 0.0260204997933823
3 0.112889632916845
4 0.330258171435118
5 0.466581921760364
6 0.468920115335401
7 0.37827462442203
8 0.276131314552078
9 0.195038096954061
10 0.142430127040127
11 0.103792298755494
12 0.079876358994267
13 0.0578056085556054
};
\addplot [semithick, color0]
table {%
0 0.00151463276741685
1 0.00601581491193179
2 0.0258338402691732
3 0.109504292782817
4 0.320363890763914
5 0.456933776826759
6 0.459139295080971
7 0.369609107899576
8 0.26720622233339
9 0.187457524710667
10 0.136368473557606
11 0.0988703672288314
12 0.0760468996513834
13 0.0548486567320312
};
\addplot [semithick, color0]
table {%
0 0.00179823115339309
1 0.00608496955478546
2 0.0234624285716395
3 0.0469179355105261
4 0.130229623130416
5 0.282779970869463
6 0.375685168853096
7 0.337394372331987
8 0.243935907644473
9 0.160514768216071
10 0.113965792334777
11 0.0784280387932688
12 0.0578605867825509
13 0.0391245325823791
};
\addplot [semithick, color0]
table {%
0 0.00145128996812974
1 0.00603716343456214
2 0.0266418970930314
3 0.0903801329358907
4 0.274512832730741
5 0.483968340392699
6 0.578804425411897
7 0.553880824613396
8 0.479690814859249
9 0.390700819940319
10 0.31592705197331
11 0.247647194078723
12 0.20091280033125
13 0.151309895385792
};
\addplot [semithick, color0]
table {%
0 0.00164914111332742
1 0.00599600973846397
2 0.0242975874749114
3 0.0725079810187069
4 0.222652916249611
5 0.398302586390748
6 0.460320450140975
7 0.393335529260461
8 0.286767742652405
9 0.195505605355008
10 0.14170931458431
11 0.101264648777551
12 0.0771101227044831
13 0.0542183968984997
};
\addplot [semithick, color1]
table {%
0 0.00131061598951507
1 0.0059829237754285
2 0.026483913969913
3 0.103256244652841
4 0.272221865156676
5 0.379294093820927
6 0.329489597129501
7 0.232584325536033
8 0.151286255021143
9 0.0949979059871336
10 0.0586498890115857
11 0.03588991299745
12 0.0218551149858534
13 0.0132713914781089
};
\end{axis}

\end{tikzpicture}\\[-1em]
			\caption{$\beta{=}2.08, \gamma{=}0.56$}
	\end{subfigure}
	\hfill
	 \begin{subfigure}{.24\textwidth}
		\setlength{\figurewidth}{.8\textwidth}
		\setlength{\figureheight}{\figurewidth}
\begin{tikzpicture}

\definecolor{color0}{rgb}{0.12156862745098,0.466666666666667,0.705882352941177}
\definecolor{color1}{rgb}{1,0.498039215686275,0.0549019607843137}

\begin{axis}[
height=\figureheight,
tick pos=left,
width=\figurewidth,
xmin=0, xmax=13,
ymin=0, ymax=1
]
\addplot [semithick, color0]
table {%
0 0.00157986648396205
1 0.0099215994745461
2 0.0585970049937675
3 0.125135758889327
4 0.133994855189133
5 0.106167185604469
6 0.0677231652477748
7 0.0388382354143459
8 0.0213788345978476
9 0.0119676763748808
10 0.00704135615452879
11 0.004208322826308
12 0.0027620570165634
13 0.00163117717476956
};
\addplot [semithick, color0]
table {%
0 0.00125430180654217
1 0.00990796956555614
2 0.0673706346557185
3 0.273911198685326
4 0.403521718606458
5 0.369095392951387
6 0.273889834316421
7 0.180501255234911
8 0.120705455482088
9 0.0828469357537132
10 0.0557002734159902
11 0.0376641227868976
12 0.0280343939458259
13 0.0184603473175276
};
\addplot [semithick, color0]
table {%
0 0.00166109448858019
1 0.00996951387676743
2 0.056875628944909
3 0.0951134053115965
4 0.120493029202377
5 0.151882930640852
6 0.149098097045604
7 0.105819231710774
8 0.0636392860553335
9 0.0364709389027249
10 0.0226648957011255
11 0.0139410683491354
12 0.00962918109749889
13 0.00573034546677634
};
\addplot [semithick, color0]
table {%
0 0.00181492244042278
1 0.00992223185287606
2 0.0534024885431986
3 0.0614616231977193
4 0.0708693192478051
5 0.10556133720073
6 0.120220229360741
7 0.0889641585530421
8 0.05161565120396
9 0.0276548110410431
10 0.0167852076636188
11 0.00998932350158987
12 0.00670791196074338
13 0.0038376505192687
};
\addplot [semithick, color0]
table {%
0 0.00148725936917112
1 0.00981266229254656
2 0.0598622268943996
3 0.14628041743921
4 0.169758149878938
5 0.137610602592561
6 0.0892910918989353
7 0.0520166435369535
8 0.0295733616105886
9 0.0170720327850944
10 0.0102279037159327
11 0.00625576501202674
12 0.00417952404856565
13 0.00251216044852546
};
\addplot [semithick, color0]
table {%
0 0.00115472113125556
1 0.00981989118153656
2 0.0706159767547573
3 0.365838852665696
4 0.481857888131167
5 0.328717180329959
6 0.169461729898446
7 0.0908125286356882
8 0.0570591912186873
9 0.0390388368653388
10 0.0250351248576373
11 0.016681819105213
12 0.0119441221829845
13 0.00790916104548785
};
\addplot [semithick, color0]
table {%
0 0.00141738186905453
1 0.00989703914648927
2 0.0625114595290188
3 0.192073535716872
4 0.225247782269668
5 0.156900180902936
6 0.0866641895906617
7 0.0472577114181316
8 0.026937509959735
9 0.016124524158989
10 0.00969998948945095
11 0.00600303074069283
12 0.0040456720351804
13 0.0024858076504123
};
\addplot [semithick, color0]
table {%
0 0.00139393217937873
1 0.00988570623267851
2 0.0632464832507782
3 0.212565651834363
4 0.234257438821124
5 0.135249886595378
6 0.0615028861127061
7 0.030686383749251
8 0.017010002961884
9 0.0101394380822389
10 0.0059585461092615
11 0.00365257652704411
12 0.00241051390125572
13 0.0014805250143111
};
\addplot [semithick, color0]
table {%
0 0.00124346029444825
1 0.00983664727819428
2 0.0675071286611396
3 0.29967984411432
4 0.365668670801528
5 0.21975786100111
6 0.100019569640227
7 0.0500915793646865
8 0.0292874414211471
9 0.018752564661124
10 0.0114951183011609
11 0.00734164866991575
12 0.0050380887069029
13 0.00321594600427889
};
\addplot [semithick, color0]
table {%
0 0.00133000859116686
1 0.00992245499606585
2 0.0649730174472419
3 0.212805776294698
4 0.338329946113733
5 0.381893147357064
6 0.350816494324876
7 0.261734347462975
8 0.183479918293481
9 0.126065490535602
10 0.087307920270426
11 0.0596694601682972
12 0.0453178988023797
13 0.0296622500378276
};
\addplot [semithick, color0]
table {%
0 0.0014872098088439
1 0.00990577805142306
2 0.0608093644361089
3 0.155039718365419
4 0.185818718971701
5 0.154885688400158
6 0.103595299529964
7 0.0616027820291929
8 0.0358327940735908
9 0.0211589831372903
10 0.0128610827802666
11 0.00794018314589777
12 0.00537944628082388
13 0.00327086801604247
};
\addplot [semithick, color0]
table {%
0 0.00128531471984572
1 0.00986689517827541
2 0.0663442937406267
3 0.254908112179716
4 0.346531596938191
5 0.279667126594279
6 0.180290913823547
7 0.107756455972847
8 0.0675933576067573
9 0.0440850985350863
10 0.0282988709277393
11 0.0184895394127829
12 0.013189729477062
13 0.00846923729801672
};
\addplot [semithick, color0]
table {%
0 0.00166701680134946
1 0.00991092134408013
2 0.0563256870733051
3 0.101806636604122
4 0.0989041701008577
5 0.0762255414812968
6 0.0473303665984428
7 0.0266981964758762
8 0.0141044623123445
9 0.00756515055029505
10 0.00432804816757035
11 0.00252268175198032
12 0.0016133926586947
13 0.000930749904911373
};
\addplot [semithick, color0]
table {%
0 0.00153283243884166
1 0.00994659525129984
2 0.0596476754476884
3 0.135914386204316
4 0.169843613208122
5 0.165152456564858
6 0.127402955909937
7 0.0811652173910512
8 0.0478295149925658
9 0.02812661416257
10 0.0173546423051207
11 0.0107519838931618
12 0.0074025580391922
13 0.00448462000312186
};
\addplot [semithick, color0]
table {%
0 0.00168769056522354
1 0.00994862697028282
2 0.0561509591863721
3 0.0913831205396715
4 0.101174943477002
5 0.106630249230059
6 0.0887330150519204
7 0.0573507977016657
8 0.0322025027823519
9 0.017691756443684
10 0.0105831970864429
11 0.00630473323673132
12 0.00419084318441882
13 0.00243877626640481
};
\addplot [semithick, color0]
table {%
0 0.00157717175849814
1 0.00984943098646261
2 0.0578647126875725
3 0.110772614222971
4 0.135007720625883
5 0.145100908640634
6 0.123286134190976
7 0.0814556491899659
8 0.0478079040419064
9 0.0273622386242141
10 0.016793798529403
11 0.0102808352004628
12 0.00698483561420546
13 0.00415163567006104
};
\addplot [semithick, color0]
table {%
0 0.0013718171509667
1 0.00981416658822132
2 0.0632295720917594
3 0.206610786761733
4 0.242245981979212
5 0.161557629473871
6 0.0853691852975185
7 0.0455213942667427
8 0.0260014469780358
9 0.0156541070084998
10 0.00940878748046137
11 0.00584294894958064
12 0.00392142422030809
13 0.00241600519766373
};
\addplot [semithick, color0]
table {%
0 0.00115212859672443
1 0.00980041707219006
2 0.0705778774608699
3 0.370251175717565
4 0.470042614007442
5 0.291878806505642
6 0.133563365271959
7 0.0670151960812827
8 0.0407190045141289
9 0.0273482157991783
10 0.0171863407222592
11 0.0113004100743757
12 0.00794749047613424
13 0.00521709192155327
};
\addplot [semithick, color0]
table {%
0 0.00107771659569822
1 0.0098303557894078
2 0.0737531886669971
3 0.461768955725129
4 0.568326920061662
5 0.322050975651667
6 0.12237136748806
7 0.0560591363259508
8 0.0337877299872281
9 0.0234933254481487
10 0.014752932461129
11 0.00980657954926463
12 0.00692913806807438
13 0.00464614808690045
};
\addplot [semithick, color0]
table {%
0 0.00130617369787169
1 0.00976281935300584
2 0.0648645550770958
3 0.24559274602512
4 0.281707624591631
5 0.163246512193
6 0.0735366550210286
7 0.0365675085953307
8 0.0207309744644466
9 0.0126950123625047
10 0.00758034533263196
11 0.00471841810533596
12 0.00313950160996136
13 0.00195047058496904
};
\addplot [semithick, color1]
table {%
0 0.00131061598951507
1 0.00987322795591926
2 0.0661504730608593
3 0.246506126612585
4 0.297980384344429
5 0.183881107262951
6 0.0915644195594875
7 0.0425331257103978
8 0.0192501755629394
9 0.00862022614203322
10 0.00384257897517436
11 0.00170947972588683
12 0.000759841929358014
13 0.000337606681178899
};
\end{axis}

\end{tikzpicture}\\[-1em]
		\caption{$\beta{=}3.01, \gamma{=}0.97$}
	\end{subfigure}
	\hfill
	 \begin{subfigure}{.24\textwidth}
		\setlength{\figurewidth}{.8\textwidth}
		\setlength{\figureheight}{\figurewidth}
\begin{tikzpicture}

\definecolor{color0}{rgb}{0.12156862745098,0.466666666666667,0.705882352941177}
\definecolor{color1}{rgb}{1,0.498039215686275,0.0549019607843137}

\begin{axis}[
height=\figureheight,
tick pos=left,
width=\figurewidth,
xmin=0, xmax=13,
ymin=0, ymax=1
]
\addplot [semithick, color0]
table {%
0 0.00108126909316748
1 0.0149251160177228
2 0.141959587028365
3 0.511509498987683
4 0.407344508594689
5 0.141529239677551
6 0.0395528007966909
7 0.01600047105833
8 0.00901512030409554
9 0.00581720694762298
10 0.00333963490949643
11 0.0020094156708649
12 0.00133420842163742
13 0.000784238973394483
};
\addplot [semithick, color0]
table {%
0 0.00126248969536979
1 0.0147867854241449
2 0.127830381021038
3 0.347476711694722
4 0.182901734685218
5 0.0399989894948474
6 0.00847389540503534
7 0.00306694432286372
8 0.00150556574818841
9 0.000845930596005068
10 0.000437218519114798
11 0.000240312672057393
12 0.000143124353023797
13 7.75583502625651e-05
};
\addplot [semithick, color0]
table {%
0 0.0012064298451296
1 0.0149433659918289
2 0.133108288990025
3 0.387515481615511
4 0.25963958035425
5 0.0849810167882981
6 0.0250709827587279
7 0.0103660021801396
8 0.00553260412960382
9 0.003311845290032
10 0.00184002014518629
11 0.00106082200928213
12 0.000678942762623067
13 0.000381466023823706
};
\addplot [semithick, color0]
table {%
0 0.00119579505477628
1 0.014750616709801
2 0.131667650516528
3 0.399240550027095
4 0.231088101804066
5 0.0509642524416727
6 0.0104393909431299
7 0.00371764674577769
8 0.00185803898894363
9 0.00107428979127814
10 0.000563863439449151
11 0.000315355744457984
12 0.000190787640438182
13 0.000105158152666235
};
\addplot [semithick, color0]
table {%
0 0.00117998193107246
1 0.0147741377672719
2 0.133159921943119
3 0.415241758231893
4 0.249987272349753
5 0.0569816131505307
6 0.011881808494375
7 0.00425727916041054
8 0.00215470654579655
9 0.00126130735145587
10 0.000667709602197343
11 0.000376410930676906
12 0.000229790338198472
13 0.000127636750300762
};
\addplot [semithick, color0]
table {%
0 0.00123443054320033
1 0.0148443274705844
2 0.1304514786383
3 0.362219104943423
4 0.215246558075707
5 0.0584206051584201
6 0.014890037773808
7 0.00577368040466775
8 0.0029541781178606
9 0.00170422101900405
10 0.000913391028234407
11 0.000513862590640849
12 0.000317005424856477
13 0.000174535773766335
};
\addplot [semithick, color0]
table {%
0 0.00115037091699166
1 0.0150611877242166
2 0.137885983030712
3 0.455814573325042
4 0.346344418210055
5 0.122309571871198
6 0.036352888738731
7 0.0151126921261069
8 0.00834166116585213
9 0.0052268002749374
10 0.00297249647946125
11 0.00176581593779185
12 0.00116989675819736
13 0.000678481557168706
};
\addplot [semithick, color0]
table {%
0 0.00145164212354037
1 0.0149277781670181
2 0.119078962783572
3 0.209547654117862
4 0.103773031205661
5 0.0376739498716168
6 0.0137245663297837
7 0.0061505280578719
8 0.00308487437075062
9 0.00164294393154511
10 0.000868395447772898
11 0.000471036047410973
12 0.000285355437855888
13 0.000149793027660348
};
\addplot [semithick, color0]
table {%
0 0.00159241797579656
1 0.0148610427261127
2 0.111420831203946
3 0.141776549156238
4 0.0615540933382267
5 0.0258305551873162
6 0.0110048833848472
7 0.00525887428942544
8 0.00256743760557693
9 0.00128964462567916
10 0.000666799760481933
11 0.000351997143987513
12 0.000208448475164158
13 0.000105768421089709
};
\addplot [semithick, color0]
table {%
0 0.00137186862673889
1 0.0147856778754142
2 0.121670442543318
3 0.25753078742803
4 0.12130863909461
5 0.0313566986191896
6 0.00820779215195476
7 0.00321256651688837
8 0.00155521346865689
9 0.000831176970967535
10 0.000426434041541611
11 0.000229607312341453
12 0.0001352337837203
13 7.11857322650547e-05
};
\addplot [semithick, color0]
table {%
0 0.00133977002140427
1 0.0150067127736863
2 0.125457033076867
3 0.293540391127339
4 0.161180572196863
5 0.0484784630477548
6 0.0140889752893902
7 0.00578396291543166
8 0.00291284552243474
9 0.00162781447199125
10 0.000867591151142002
11 0.000480125358806962
12 0.000295483099590889
13 0.000159766314758686
};
\addplot [semithick, color0]
table {%
0 0.00126480063479708
1 0.0149317949069321
2 0.128727369530599
3 0.309678764459676
4 0.230783925452521
5 0.121985831397916
6 0.0593032544769849
7 0.0305169060634153
8 0.0174727951742514
9 0.0105116981648082
10 0.00612690509536256
11 0.00361874057052423
12 0.00242660514544667
13 0.00136689069964078
};
\addplot [semithick, color0]
table {%
0 0.00105574069126015
1 0.0148668012753695
2 0.143474256503006
3 0.560681038657956
4 0.411810077345952
5 0.0961337376724954
6 0.0174393171910296
7 0.00591822689164709
8 0.00313744073326489
9 0.0019947687958695
10 0.00108899968505257
11 0.000639861170364428
12 0.000405459359323615
13 0.000236620408397019
};
\addplot [semithick, color0]
table {%
0 0.00109194780819941
1 0.0149545901791345
2 0.141534383266405
3 0.515471185101003
4 0.393568890811796
5 0.118552350478882
6 0.0287363156980601
7 0.0109549855147352
8 0.00600709217578252
9 0.00382236382054808
10 0.00214525675807767
11 0.00127642619568477
12 0.000833017309773362
13 0.000487284924537589
};
\addplot [semithick, color0]
table {%
0 0.00139250916123174
1 0.0150096594924448
2 0.121862374676831
3 0.220483470589931
4 0.155955718091379
5 0.103959603377937
6 0.0648488759332269
7 0.037116660227387
8 0.0211323327430228
9 0.0121761049030046
10 0.00710827358660524
11 0.00412889137197766
12 0.00276982667923097
13 0.00151987680666189
};
\addplot [semithick, color0]
table {%
0 0.00125299447524367
1 0.0149145749375558
2 0.129847233122722
3 0.354533301435774
4 0.208959972329662
5 0.057636053100102
6 0.0149190792466927
7 0.00585187303124279
8 0.00299274714267181
9 0.00172889431296512
10 0.000928663269527609
11 0.000521269096056137
12 0.000322248309124159
13 0.000177246650866194
};
\addplot [semithick, color0]
table {%
0 0.00138073133217249
1 0.0148976491920927
2 0.122553545020082
3 0.269848439877388
4 0.124071269787452
5 0.0288139751099363
6 0.00682398869293774
7 0.00256637960114432
8 0.00122786814182772
9 0.000657146773593923
10 0.000333669359997993
11 0.000178886188243886
12 0.000104719307512421
13 5.52691190076564e-05
};
\addplot [semithick, color0]
table {%
0 0.00147161883723901
1 0.0150529845871306
2 0.119662077856951
3 0.214289647146872
4 0.0994238815044595
5 0.0314139832161
6 0.0101210758166081
7 0.00428213179407467
8 0.00208557562445831
9 0.00109722504454367
10 0.000570948636013964
11 0.000305636520974677
12 0.000182638387443074
13 9.5317069602547e-05
};
\addplot [semithick, color0]
table {%
0 0.00125850479013833
1 0.0148093018653299
2 0.127724704759221
3 0.325620329016234
4 0.205162544426883
5 0.071688719713932
6 0.0232896363870356
7 0.0100751665004638
8 0.00532633588224641
9 0.00308899201268882
10 0.00169540280602888
11 0.000966302537474781
12 0.000611074892427664
13 0.000337201500073815
};
\addplot [semithick, color0]
table {%
0 0.00135289365938041
1 0.0149771853409322
2 0.124295692080623
3 0.271227839376846
4 0.15700956715454
5 0.0577072635803982
6 0.0203624851010461
7 0.00907618582462041
8 0.00470840852041498
9 0.00264528493173378
10 0.00143953603713358
11 0.000805008153221643
12 0.000504586275229982
13 0.000273354462711322
};
\addplot [semithick, color1]
table {%
0 0.00131061598951507
1 0.0148683590491655
2 0.124720771985736
3 0.277277529822087
4 0.16454741686527
5 0.065149160670049
6 0.0232828815680417
7 0.00807187055414517
8 0.00277099505587131
9 0.000948114718039603
10 0.000324046791606558
11 0.000110710297137938
12 3.78157874348204e-05
13 1.29208271774742e-05
};
\end{axis}

\end{tikzpicture}\\[-1em]
		\caption{$\beta{=}3.81, \gamma{=}1.35$}
	\end{subfigure}
	\caption{\textbf{SIR experiment:} Prior samples \mybox{pythonblue} from the trained PriorCVAE model for different values of $\beta$ and $\gamma$ along with the ODE solution \mybox{pythonorange}. The ODE solution is always enveloped by the PriorCVAE samples.}
	\label{fig:sir_priors}
\end{figure*}
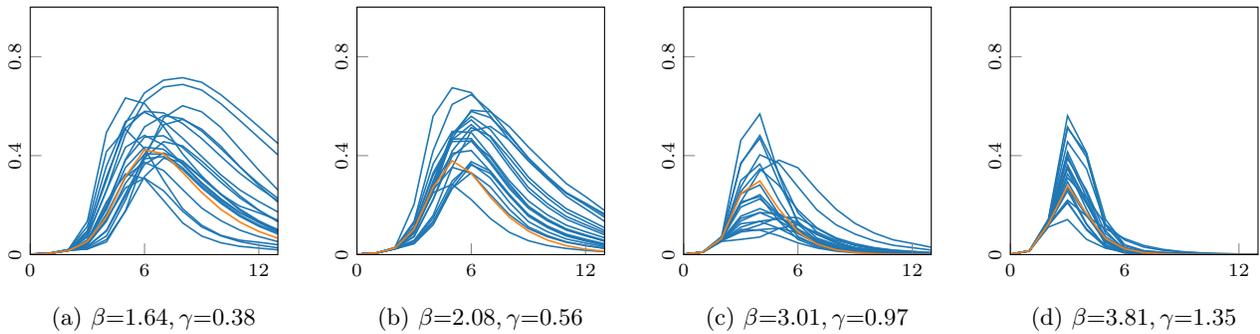

\begin{figure*}[h!]
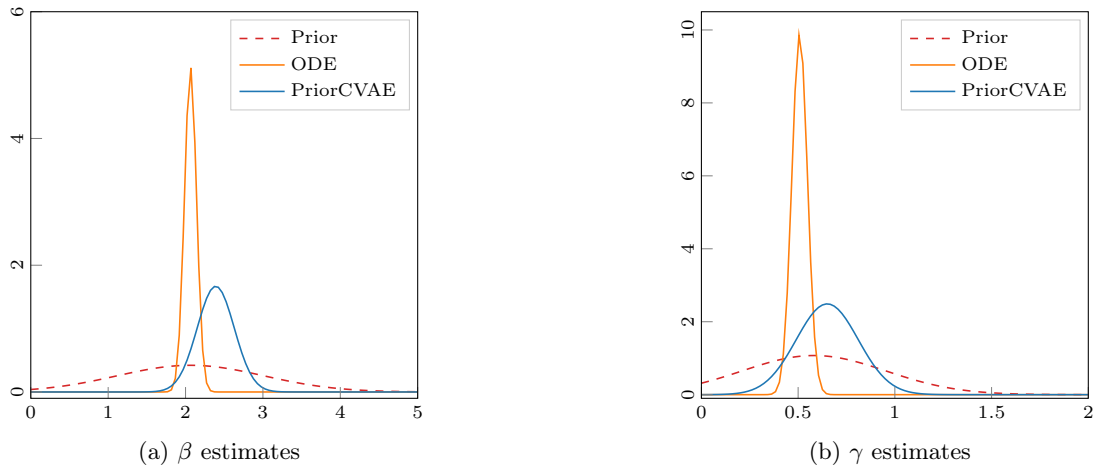

	\scriptsize
	\setlength{\figurewidth}{.3\textwidth}
	\setlength{\figureheight}{\figurewidth}
	\pgfplotsset{axis on top,scale only axis,width=\figurewidth,height=\figureheight, ylabel near ticks,y tick label style={rotate=90}, tick label style={font=\scriptsize}}
	\begin{subfigure}{0.48\textwidth}
		\centering
		\input{figures/SIR/SIR_beta_posterior}
		\caption{$\beta$ estimates}
	\end{subfigure}
	\hfill
	\begin{subfigure}{0.48\textwidth}
		\centering
		\input{figures/SIR/SIR_gamma_posterior}
		\caption{$\gamma$ estimates}
	\end{subfigure}
	\caption{\textbf{SIR experiment:} MCMC parameter estimates of the ODE and the PriorCVAE model.}
	\label{fig:sir_params}
\end{figure*}

\end{document}